\documentclass[twocolumn]{IEEEtran}
\pdfoutput=1
\usepackage[pdftex]{graphicx}
\usepackage{amsmath}
\usepackage{mathrsfs}
\usepackage{algorithm, algorithmic}
\usepackage{array}
\usepackage[caption=false,font=footnotesize]{subfig}
\ifCLASSOPTIONcaptionsoff
  \usepackage[nomarkers]{endfloat}
  \let\MYoriglatexcaption\caption
  \renewcommand{\caption}[2][\relax]{\MYoriglatexcaption[#2]{#2}}
\fi
\usepackage{url}

\usepackage{bm}        % for \bm macro
\usepackage{hyperref}
\usepackage{cleveref}  % for \cref
\usepackage{amsfonts}  % for \mathbb
\usepackage{amssymb}   % for \backsim
\usepackage{booktabs}  % for three-line table, \toprule, \midrule, \bottomrule
\usepackage{tabularx}  % for \begin{tabularx} ... \end{tabularx}
\usepackage[flushleft]{threeparttable} % for table note
\usepackage{wasysym}
\usepackage{rotating}
\usepackage[english]{babel}
\usepackage{float}
\crefformat{table}{Tab.~#2#1#3}
\crefformat{section}{Section~#2#1#3}
\crefformat{figure}{Fig.~#2#1#3}
\crefformat{equation}{Eq.~(#2#1#3)}
\crefformat{algorithm}{Algorithm~#2#1#3}
\newcommand{\specialcell}[2][c]{%
  \begin{tabular}[#1]{@{}l@{}}#2\end{tabular}}

\newcommand{\myfourstar}{~~\llap{\raisebox{-0.3ex}{\rotatebox{45}{\wasylozenge}}}~~}

\begin{document}
\title{Reweighted Infrared Patch-Tensor Model With Both Non-Local and Local Priors for Single-Frame Small Target Detection}

\author{Yimian~Dai, Yiquan~Wu
	\thanks{Y. Dai is with the College of Electronic and Information Engineering, Nanjing University of Aeronautics and Astronautics, Nanjing 211106, China (e-mail: yimian.dai@gmail.com).}  
	\thanks{Y. Wu is with the College of Electronic and Information Engineering, Nanjing University of Aeronautics and Astronautics, Nanjing 211106, China, and also with Key Laboratory of Spectral Imaging Technology CAS, Xi'an Institute of Optics and Precision Mechanics of CAS, Xi'an 710000, China (e-mail: nuaaimagestrong@163.com).}
	\thanks{Manuscript received December 19, 2016; revised March 29, 2017.}
}

\markboth{IEEE Journal of Selected Topics in Applied Earth Observations and Remote Sensing,~Vol.~xx, No.~x,
December~2016}%
{Dai \MakeLowercase{\textit{et al.}}: Reweighted infrared patch-tensor model with both non-local and local priors for single-frame small target detection}

\maketitle
\begin{abstract}	
	Many state-of-the-art methods have been proposed for infrared small target detection.
	They work well on the images with homogeneous backgrounds and high-contrast targets.
	However, when facing highly heterogeneous backgrounds, they would not perform very well, mainly due to: 1) the existence of strong edges and other interfering components, 2) not utilizing the priors fully. 
	Inspired by this, we propose a novel method to exploit both local and non-local priors simultaneously. 
	Firstly, we employ a new infrared patch-tensor (IPT) model to represent the image and preserve its spatial correlations.    
	Exploiting the target sparse prior and background non-local self-correlation prior, the target-background separation is modeled as a robust low-rank tensor recovery problem. 
	Moreover, with the help of the structure tensor and reweighted idea, we design an entry-wise local-structure-adaptive and sparsity enhancing weight to replace the globally constant weighting parameter. 
	The decomposition could be achieved via the element-wise reweighted higher-order robust principal component analysis with an additional convergence condition according to the practical situation of target detection.  
	Extensive experiments demonstrate that our model outperforms the other state-of-the-arts, in particular for the images with very dim targets and heavy clutters. 
\end{abstract}


\IEEEpeerreviewmaketitle

% \tableofcontents

\section{Introduction}\label{sec:introduction}
\IEEEPARstart{I}{nfrared} small target detection is a key technique for many applications, including early-warning system, precision guided weapon, missile tracking system, and maritime surveillance system \cite{Kim2012PR451,Liu2014PR479,Bai2016IToC4612}. 
Traditional sequential detection methods, such as 3D matched filter \cite{Reed1988IToAaES244}, improved 3D filter \cite{Li2005OE4410}, and multiscan adaptive matched filter \cite{Melendez1995PoS2561}, are workable in the case of static background, exploiting the target spatial-temporal information. 
Nevertheless, with the recent fast development of high-speed aircrafts \cite{Zheng2016IJoSTiAEOaRS91} like anti-ship missiles, the imaging backgrounds generally change quickly due to rapid relative motion between the imaging sensor and the target. 
The performance of the spatial-temporal detection method degrades rapidly. 
Therefore, the research of single-frame infrared small target detection is of great importance and has attracted a lot of attention in recent years.

Different from general object or saliency detection tasks, the main challenge of infrared small target detection is lacking enough information.
Due to the long imaging distance, the target is always small without any other texture or shape features. 
As the target type, imaging distance, and neighboring environment differ a lot in real scenes, the target brightness could vary from extremely dim to very bright (see \cref{fig:Original} for example). 
In the absence of spatial-temporal information and the target features like shape and size, the characteristics of the background \cite{Guo2016IJoSTiAEOaRS94} and the relation between the background and target are very important priors for single-frame infrared small target detection. 
Thus how to design a model to incorporate and exploit these priors is vital for infrared small target detection in a single image.

\subsection{Prior work on single-frame infrared small target detection}
The previously proposed single-frame infrared small target detection methods could be roughly classified into two categories. 
In the first type, a local background consistent prior is exploited, assuming the background is slowly transitional and nearby pixels are highly correlated. 
As a result, the target is viewed as the one that breaks this local correlation. 
Under this assumption, the classical methods, such as two-dimensional least mean square (TDLMS) filter \cite{Hadhoud1988IToCaS355} and Max-Median filter \cite{Deshpande1999SISoOSEaI}, enhance the small target by subtracting the predicted background from the original image. 
Unfortunately, besides the targets, they enhance the edges of the sky-sea surface or heavy cloud clutter as well, since these structures also break the background consistency as the target does. 
To differentiate the real target and high-frequency change, some edge analysis approaches \cite{Bae2012IPT551, Cao2008IJoIaMW292} have been proposed to extend these methods to estimate the edge direction in advance and preserve the edges. 
Bai et al.\ \cite{Bai2010PR436} designed a new Top-Hat transformation using two different but correlated structuring elements. 
Another class of local prior based methods exploits the local contrast, which is computed by comparing a pixel or a region only with its neighbors. 
The seminal work of Laplacian of Gaussian (LoG) filter based method \cite{Kim2009JoIMaTW309} has motivated a broad range of studies on the Human Visual System (HVS), and has led to a series of HVS based methods, e.g., Difference of Gaussians (DoG) \cite{Wang2012IPT556}, second-order directional derivative (SODD) filter \cite{Qi2013IGaRSL103}, local contrast measure (LCM) \cite{Chen2014IToGaRS521}, improved local contrast measure (ILCM) \cite{Han2014IGaRSL1112}, multiscale patch-based contrast measure (MPCM) \cite{Wei2016PR58}, multiscale gray difference weighted image entropy \cite{Deng2016IToAaES521}, improved difference of Gabors (IDoGb) \cite{Han2016IGaRSL133}, local saliency map (LSM) \cite{Chen2016IGaRSL137}, weighted local difference measure (WLDM) \cite{Deng2016IToGaRS547}, local difference measure (LDM) \cite{Deng2017PR61}, etc.

% Existing priors (or single-frame detection methods) are roughly classified into two categories — local prior based on the intrinsic consistency of backgrounds and non-local prior based on the self-similarity of them.
% As a result they are not particularly robust to variations in target appearance and fail on challenging tracking problems.
% IPI 的工作，以及其缺点，向量化破坏空间结构，对强边缘没办法。
% WIPI 的工作，极其缺点
%% LRSR 的工作，以及其缺点，用字典去刻画没有结构，大小，方位，亮度未知的小目标很难。
% IPI 的缺点
% 1. 破坏空间结构
% 2. 抑制强边缘不好

The second type of single-frame infrared small target detection methods which has not been explored extensively, exploits the non-local self-correlation property of background patches, assuming that all background patches come from a single subspace or a mixture of low-rank subspace clusters. 
Then, target-background separation can be realized with the low-rank matrix recovery \cite{Fornasier2011SJoO214}. 
Essentially, this type of methods attempts to model the infrared small target as an outlier in the input data. 
To this end, Gao et al. \cite{Gao2013IToIP2212} generalized the traditional infrared image model to a new infrared patch-image model via local patch construction. 
Then the target-background separation problem is reformulated as a robust principal component analysis (RPCA) \cite{Candes2011JotA583} problem of recovering low-rank and sparse matrices, achieving a state-of-the-art background suppressing performance. 
To correctly detect the small target located in a highly heterogeneous background, He et al. \cite{He2015IPT68} proposed a low-rank and sparse representation model under the multi-subspace-cluster assumption. 

\subsection{Motivation}

% So is the target size, which is possible from $1 \times 1$ to $10 \times 10$ pixels or even more. 
Existing methods detect the infrared small target by either utilizing the local pixel correlation or exploiting the non-local patch self-correlation. 
From our observation, the unsatisfying performance of local prior methods \cite{Deng2016IToAaES521,Deng2016IToGaRS547} in detecting the dim target under complicated background mainly lies in their imperfect grayscale based center-difference measures.
The saliency of a dim but true target would be easily overwhelmed by the measured saliency of some rare structures.
We call this phenomenon the \textit{rare structure effect}.
In contrast, the non-local prior methods \cite{Gao2013IToIP2212,Dai2016IPT77,Dai2017IP&T81} suffer from the salient edge residuals. Its intrinsic reason is because the strong edge is also a sparse component as the same as the target due to lack of sufficient similar samples. 
Since the target might be dimmer than the edge, they would simultaneously be wiped out by simply increasing the threshold. 
% 
% Similar to the target, the edge not only breaks the local pixel correlation but also can be viewed as an outlier \cite{Wan2015ISJPP99} of the low-rank patch-image.
% Thus it is easy to mistake the edges as the targets and preserve them in the target image. 

% In fact, these two priors are not totally equivalent. 
% The method based on either of them is not robust enough when facing extremely complex background (see \cref{fig:Original} for example). 
% We would give a detailed explanation as follows.
 
% Similar to the target, the edge not only breaks the local pixel correlation but also can be viewed as an outlier \cite{Wan2015ISJPP99} of the low-rank patch-image.
% Thus it is easy to mistake the edges as the targets and preserve them in the target image. 
% Since the target might be dimmer than the edge, they would simultaneously be wiped out by simply improving the threshold no matter in the local prior or non-local prior based methods.

\begin{figure}[htbp]
  \centering  
  \includegraphics[width=0.29\textwidth]{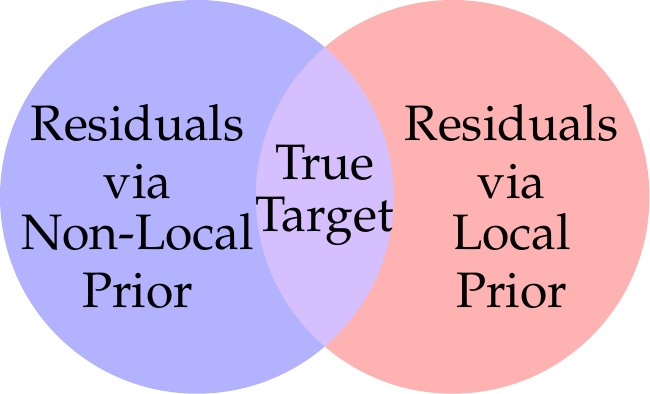}  
  \caption{Illustration of our motivation behind exploiting both non-local and local priors.}
  \label{fig:Venn}
\end{figure}

Our key observation is that the non-local prior and local prior are not equivalent, and in fact they are complementary for the problem of infrared small target detection, as illustrated in \cref{fig:Venn}. 
The often appearing false alarm components in local (non-local) prior methods could be well suppressed by the non-local (local) prior methods.
% the residuals or salient components by the methods based on local prior or non-local prior are not intersecting except the true target, as illustrated in \cref{fig:Venn}. 
For example, the stubborn strong edges in the non-local prior based methods, can be easily identified by the local edge analysis approaches.
Naturally, an intuitive way to solve above-mentioned dilemma is to extract the local structure information and merge it into the non-local prior based detection framework. 
Therefore, how to simultaneously and appropriately utilize both the local and non-local priors has been an important issue for improving the detection performance under very complex backgrounds. 
To the best of our knowledge, very few single-frame infrared small target detection methods concern this problem.

% In other words, the non-local prior and local prior are not equivalent, and in fact they are complementary for the problem of infrared small target detection.
% Therefore, by integrating the local prior into a non-local framework, the detection performance is expected to be improved.

To address this issue, we propose a single-frame small target detection framework with reweighted infrared patch-tensor model (RIPT). Our main contributions consist of the following three folds:

% 本文的贡献：
% 1. IPT, robust tensor recovery
% 2. Local and Non-local priors, 权重，local structure weight 确保了抑制强边缘
% 3. sparsity enhancement weight 根据小目标检测的实际，减少计算时间
\begin{enumerate}
	\item To dig out more information from the non-local self-correlationship in patch space, we generalize the patch-image model to a novel infrared patch-tensor model (IPT) and formulate the target-background separation task as an optimization problem of recovering low-rank and sparse tensors. 
  %The state-of-the-art IPI model can be viewed as a special case of IPT model exploiting only mode-$3$ unfolding matrix with a global constant weight.
	\item To incorporate the local structure prior into the IPT model, an element-wise weight is designed based on structure tensor, which helps to suppress the remaining edges and preserve the dim target simultaneously. 
	\item To reduce the computing time, we adopt a reweighted scheme to enhance the sparsity of the target patch-tensor. 
  Considering the particularity of infrared small target detection, an additional stopping criterion is used to avoid excessive computation.
\end{enumerate}

The proposed RIPT model is validated on different real infrared image datasets. Compared with the state-of-the-art methods, our proposed model achieves a better background suppressing and target detection performance.

The remaining of this paper is organized as follows. We propose the non-local correlation based IPT model in Section \ref{sec:IPT}. The details of the local structure weight construction are described in Section \ref{sec:WLS}. In Section \ref{sec:RIPT}, we further propose the reweighted IPT model and its detailed optimization scheme is also provided. Section \ref{sec:Experiment} presents detailed experimental results and discussion. Finally we conclude this paper in Section \ref{sec:Conclusion}.

\section{Non-local correlation driven infrared patch-tensor model} \label{sec:IPT}
To dig out more spatial correlationships, we develop a novel target-background separation framework named infrared patch-tensor model in this section. 
Before describing the details, several mathematical notations are defined first in \cref{tab:Notations}.
\begin{table*}[htb!]
	\centering
	\caption{Mathematical Notations}
	\begin{tabularx}{\textwidth}{l >{\raggedright\arraybackslash}X}		
		\toprule
		Notation                                          & Explanation                                                                                                             \\
		\midrule
		% 1 - TDLMS
		$\bm{\mathcal{X}},~\bm{X},~\bm{x},~x$             & tensor, matrix, vector, scalar.                                                                                         \\
		$\bm{X}_{(n)}$                                    & mode-$n$ matricization of tensor                                                                                        
		$\bm{\mathcal{X}} \in \mathbb{R}^{I_1\times I_2\times \cdots \times I_N}$, obtained by arranging the mode-$n$ fibers as the columns of the resulting matrix of size $\mathbb{R}^{I_n\times\prod_{k\ne n}I_k}$. \\
		$\mathrm{vec}(\bm{\mathcal{X}})$                  & vectorization of tensor $\bm{\mathcal{X}}$.                                                                             \\
		$\langle\bm{\mathcal{X}},\bm{\mathcal{Y}}\rangle$ & inner product of tensor $\bm{\mathcal{X}}$ and $\bm{\mathcal{Y}}$, which is defined as                                  
		$\langle\bm{\mathcal{X}},\bm{\mathcal{Y}}\rangle :=
		\mathrm{vec}(\bm{\mathcal{X}})^{\top}\mathrm{vec}(\bm{\mathcal{Y}})$. \\
		$\Vert\bm{\mathcal{X}}\Vert_{0}$                  & $\ell_0$ norm of tensor $\bm{\mathcal{X}}$ which counts the number of non-zero elements.                                \\    
		$\Vert\bm{\mathcal{X}}\Vert_{1}$                  & $\ell_1$ norm of tensor     $\bm{\mathcal{X}}$.                                                                         \\
		$\Vert\bm{\mathcal{X}}\Vert_{\mathrm{F}}$         & Frobenius norm of tensor                                                                                                
		$\bm{\mathcal{X}}$, which is defined by  $\Vert\bm{\mathcal{X}}\Vert_{\mathrm{F}} := \sqrt{\langle\bm{\mathcal{X}},\bm{\mathcal{X}}\rangle}$. \\
		$\mathrm{fold}_i(\bm{X})$                         & returns tensor $\bm{\mathcal{Z}}$ that                                                                              
		$\bm{Z}_{(i)} = \bm{X}$. \\
		$\Vert\bm{X}\Vert_*$                              &                                                                                                                         
		nuclear norm of matrix $\bm{X}$, which is defined by
		$\Vert\bm{X}\Vert_* = \sum_i \bm{\sigma}_i$, where the SVD of
		$\bm{X} = \bm{U}\mathrm{diag}(\bm{\sigma})\bm{V}^{\top}$. \\
		$\mathcal{S}_{\mu}(\bm{x})$                       & element-wise shrinkage operator is defined as $\mathcal{S}_{\mu}(\bm{x}) := \mathrm{sign}(\bm{x})\max(|\bm{x}|-\mu,0)$. 
		$\mathcal{S}_{\mu}(\bm{x})$ is the closed-form solution of the problem:
		$\hat{\bm{y}} = \arg\min_{\bm{y}}\Vert\bm{x}-\bm{y}\Vert^2_{\mathrm{F}} +
		\lambda\Vert\bm{y}\Vert_1$ \cite{Beck2009SJoIS21}. \\
		$\mathcal{D}_{\mu}(\bm{X})$                       & matrix singular value thresholding operator:                                                                            
		$\mathcal{D}_{\mu}(\bm{X}) := \bm{U}\mathrm{diag}(\bar{\sigma})\bm{V}^{\top}$,
		where $\bm{X} = \bm{U}\mathrm{diag}(\sigma)\bm{V}^{\top}$ is the SVD of
		$\bm{X}$ and $\bar{\sigma} := \max(\sigma-\mu,0)$. $\mathcal{D}_{\mu}(\bm{X})$
		is the closed-form solution of the problem:
		$\hat{\bm{Y}} = \arg\min_{\bm{Y}}\Vert\bm{X}-\bm{Y}\Vert^2_{\mathrm{F}} +
		\frac{\mu}{2}\Vert\bm{Y}\Vert_*$ \cite{Cai2010SJoO204}. \\
		$\mathcal{T}_{i,\mu}(\bm{\mathcal{X}})$           &                                                                                                                         
		$\mathcal{T}_{i,\mu}(\bm{\mathcal{X}}) := \mathrm{fold}_{i}(\mathcal{D}_{\mu}(\bm{X}_{(i)}))$.  \\
		\bottomrule
	\end{tabularx}%
	\label{tab:Notations}
\end{table*}

Given an infrared image, it could be modeled as a linear superposition of target image,  background image and noise image:
\begin{equation}
	f_{\textup{F}} = f_{\textup{B}} + f_{\textup{T}} + f_{\textup{N}},
	\label{eq:Image}
\end{equation}
where $f_{\textup{F}}$, $f_{\textup{B}}$, $f_{\textup{T}}$, and $f_{\textup{N}}$ represent the input image, background image, target image, and noise image, respectively. 
Via a window sliding from the top left to the bottom right over an image, we stack the patches into a 3D cube (see the construction step in \cref{fig:Overview}). 
Then \cref{eq:Image} is transferred to the patch space with spatial structure preserved:
\begin{equation}
	\bm{\mathcal{F}} = \bm{\mathcal{B}} + \bm{\mathcal{T}} + \bm{\mathcal{N}},
	\label{eq:Patch_Tensor}
\end{equation}
where $\bm{\mathcal{F}}, \bm{\mathcal{B}}, \bm{\mathcal{T}}, \bm{\mathcal{N}} \in \mathbb{R}^{I\times J\times P}$ are the input patch-tensor, background patch-tensor, and target patch-tensor, respectively. 
$I$ and $J$ are the patch height and width, $P$ is the patch number.

\begin{figure}[htb!]
	\centering
	\includegraphics[width=0.5\textwidth]{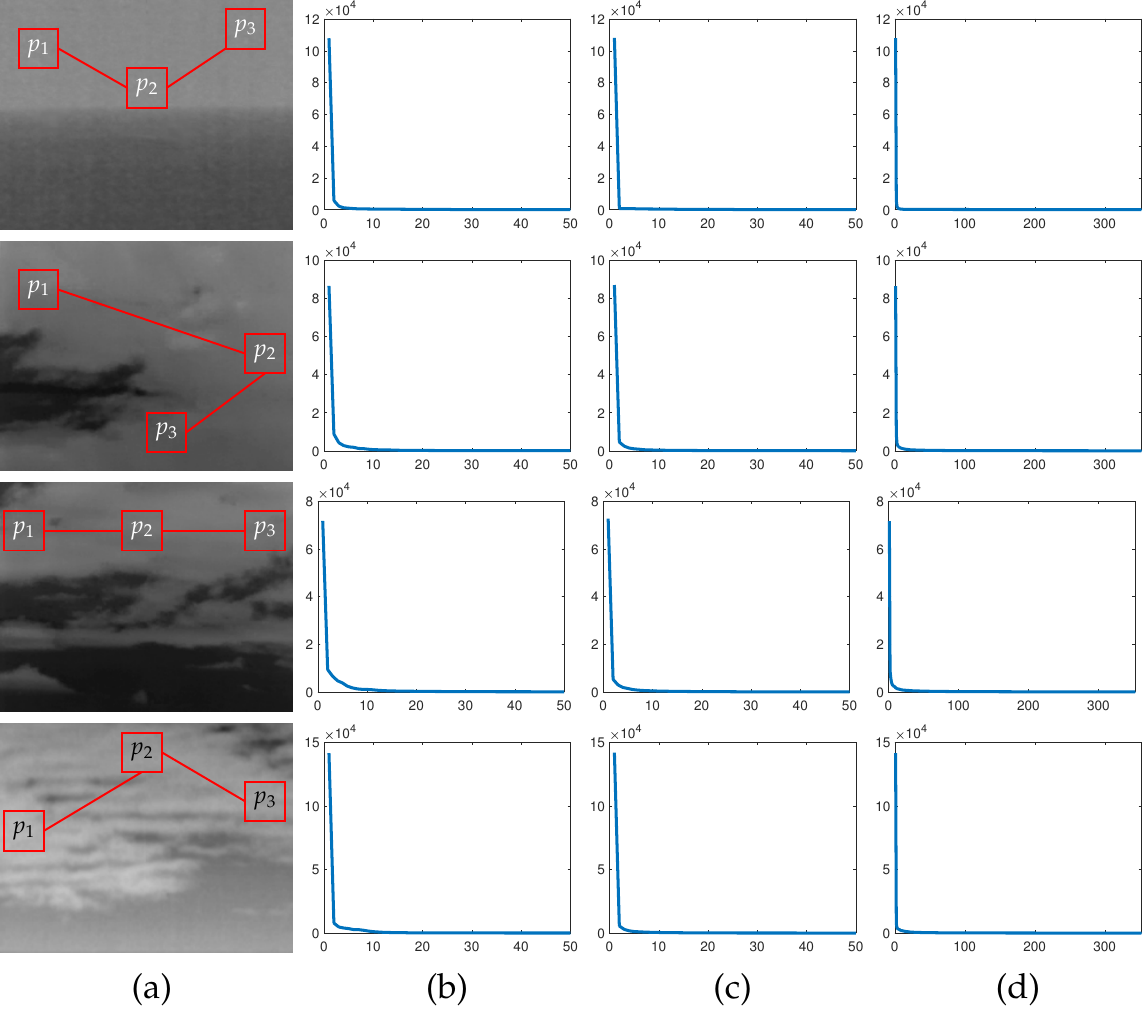}
	\caption{The illustration of the non-local similarity and the low-rank property of the mode-$i$ $(i = 1, 2, 3)$ unfolding of the patch-tensor. (a) Four representative background images. (b) -- (d) Singular values of the mode-$1$, mode-$2$, and mode-$3$ unfolding matrices of the corresponding background patch-tensors.}
	\label{fig:Patch_Tensor_Low_Rank}
\end{figure}

\textbf{Background patch-tensor $\bm{\mathcal{B}}$}. 
Generally, the background is considered as slowly transitional, which implies high correlations among both the local and non-local patches in the image, as illustrated in \cref{fig:Patch_Tensor_Low_Rank}(a).
Although patches $p_1,p_2,p_3$ locate in the different region of the image, they are equivalent.
Based on this non-local correlationship, the state-of-the-art IPI model imposed the low-rank constraint to background patch-image. 
As a patch-image is the mode-$3$ unfolding matrix of a patch-tensor, the patch-image model could be viewed as a special case of the proposed patch-tensor model essentially.
Since the main difficulty of detecting the infrared small target in a single image is lacking enough information, only considering the low-rank structure in one unfolding is insufficient to deal with highly difficult scenes. 
Naturally, it motivates us to think whether we can utilize the other two unfolding modes.
Actually, the mode-$1$ and mode-$2$ unfolding matrices of the infrared patch-tensor are also low-rank. 
In \cref{fig:Patch_Tensor_Low_Rank}(b) -- (d), the singular values of all the unfolding matrices rapidly decrease to zero, which demonstrates that every unfolding mode of the background patch-tensor is intrinsically low-rank.
Therefore, we can consider the background patch-tensor $\bm{\mathcal{B}}$ as a low-rank tensor, and their unfolding matrices are also all low-rank defined as: 
\begin{equation}
	\textup{rank}(\bm{B}_{(1)}) \le r_1,~ \textup{rank}(\bm{B}_{(2)}) \le r_2,~ \textup{rank}(\bm{B}_{(3)}) \le r_3,
	\label{eq:Background_Unfolding}
\end{equation}
where $r_1$, $r_2$, and $r_3$ all are constants, representing the complexity of the background image. 
The larger their values are, the more complex the background is.

\textbf{Target patch-tensor $\bm{\mathcal{T}}$}. 
Since the small target merely occupies several pixels in the whole image, the target patch-tensor is an extremely sparse tensor in fact. That is:
\begin{equation}
	\Vert \bm{\mathcal{T}} \Vert_{0} \le k,
	\label{eq:Target_Patch_Tensor}
\end{equation}
where $k$ is a small integer determined by the number and size of the small target. 
% This property reminds us one thing that the sparsity of the target patch-tensor does matter in the separation process. However, both Ref.\ \cite{Gao2013IToIP2212} and Ref.\ \cite{Dai2016IPT77} neglect to pay attention to the sparsity of the target patch-image during the iteration. Luckily, in the IPI model, the separated target image is clean enough when the iteration ends. Unfortunately, by simply applying the HoRPCA algorithm (singleton model) proposed in Ref.\ \cite{Goldfarb2014SJoMAaA351} to our infrared patch-tensor model, we observe that the target image would be contaminated with more and more background details as the algorithm converges. It motivates us to adopt a reweighted scheme to enhance the sparsity in our proposed RIPT model which will be discussed in the following subsection.

\textbf{Noise patch-tensor $\bm{\mathcal{N}}$}. 
In this paper, we just assume that the noise is additive white Gaussian noise and $\Vert \bm{\mathcal{N}} \Vert_\textup{F} \le \delta$ for some $\delta > 0$. Thus we have 
\begin{equation}
\Vert \bm{\mathcal{F-B-T}} \Vert_{\textup{F}} \le \delta.
\end{equation}
It should be noted that although the values of parameters $k, r, \delta$ are different depending on the images, we would not directly use them in the following sections.

Ideally, we would like to obtain a low-rank and sparse decomposition and solve the following problem:
\begin{equation}
	\min_{\bm{\mathcal{B}}, \bm{\mathcal{T}}} \  \textup{rank}(\bm{B}) + \lambda\Vert\bm{\mathcal{T}}\Vert_0, ~ \mathrm{s.t.} ~ \bm{\mathcal{B}} + \bm{\mathcal{T}} = \bm{\mathcal{F}}.
	\label{eq:NP_Hard}
\end{equation}
Unfortunately, the rank computation of a given tensor is a NP-hard problem in general \cite{Haastad1990JoA114}. 
In Ref.\ \cite{Goldfarb2014SJoMAaA351}, Goldfarb and Qin proposed the Higher-order RPCA (robust tensor recovery) through replacing the rank by a convex surrogate Tucker-rank $\textup{CTrank}(\bm{\mathcal{B}})$, and $\Vert\bm{\mathcal{T}}\Vert_0$ by $\Vert\bm{\mathcal{T}}\Vert_1$ to make the above problem tractable. 
In the singleton model, the tensor rank regularization term is defined as the sum of all the nuclear norms of the mode-$i$ unfoldings, i.e., $\textup{CTrank}(\bm{\mathcal{B}}) = \sum_i \Vert \bm{B}_{(i)}\Vert_*,\ i = 1,2,3$. 
With this relaxation, our proposed IPT model with random noise assumption can be solved by minimizing the following convex problem:
\begin{equation}
	\min_{\bm{\mathcal{B}}, \bm{\mathcal{T}}} \sum_{i=1}^{3} \Vert \bm{B}_{(i)} \Vert_* + \lambda\Vert\bm{\mathcal{T}}\Vert_1, ~ \mathrm{s.t.} ~ \Vert \bm{\mathcal{F}} - \bm{\mathcal{B}} - \bm{\mathcal{T}} \Vert_{\textrm{F}} \le \delta.
	\label{eq:IPT_model}
\end{equation}
$\lambda$ is a weighting parameter that controls the global trade-off between the background patch-tensor and the target patch-tensor. 
Larger $\lambda$ can shrink those non-target but sparse components to zeros in the target patch-tensor. 
Nevertheless, it will also over-shrink the dim target which should be preserved. 
On the contrary, a smaller $\lambda$ does retain the dim target, but it keeps the strong cloud edges as well. 
Therefore, adopting a global constant weighting parameter $\lambda$ is not a reasonable scheme for detecting the infrared small target in a complex scene. 
Naturally, it motivates us to design an entry-wise weighting scheme.

\section{Incorporating local structure prior}\label{sec:WLS}
\begin{figure}[htb!]
	\centering
	\includegraphics[width=0.5\textwidth]{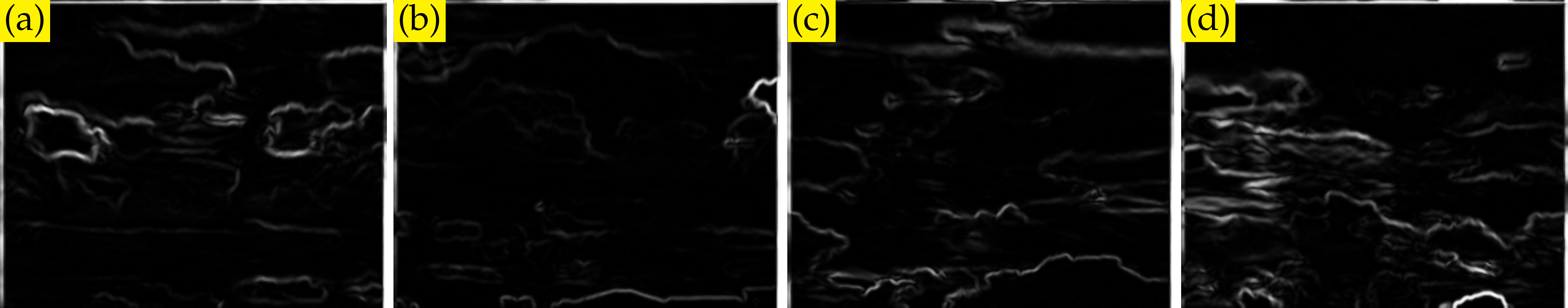}
	\caption{Illustrations of the structure tensor based local structure weight map for \cref{fig:Original}(a) -- (d).}
	\label{fig:Structure}
\end{figure}

In this section, we focus our emphasis on combining the local structure prior and non-local correlation prior together into our model. 
we construct a local structure weight and interpret it as an edge salience measure. 
For the sake of simplicity, the local structure weight is designed on the basis of the image structure tensor. 
Structure tensor is widely used in many partial differential equation (PDE) based methods \cite{Weickert1999IJoCV312-3,Wu2017SP131} to estimate the local structure information in the image, including edge orientation. 
To integrate the local information, the structure tensor is constructed based on a local regularization of a tensorial product, which is defined as
\begin{equation}
	\bm{J}_{\alpha}(\nabla u_{\sigma}) = G_{\alpha} * (\nabla \bm{u}_{\sigma} \otimes \nabla \bm{u}_{\sigma}) =
	\begin{pmatrix}
		J_{11} & J_{12} \\
		J_{21} & J_{22} \\
	\end{pmatrix},
	\label{eq:Structure_Tensor}
\end{equation}
where $\bm{u}_{\sigma}$ is a Gaussian-smoothed version of a given image $\bm{u}$. 
$\sigma > 0$ is the standard deviation of the Gaussian kernel which denotes the noise scale, making the edge detector ignorant of small details. 
$\bm{J}_{\alpha}$ is a symmetric and positive semi-definite matrix, which has two orthonormal eigenvectors denoted as $\bm{w}$ and $\bm{v} = \bm{w}^{\bot}$, where $\bm{w} = \left( 2 J_{12}, J_{22} - J_{11} + \sqrt{ \left( J_{22} - J_{11} \right)^2 + 4 J_{12}^2 } \right)^{\top}, \bm{w} = \bm{w} / \vert \bm{w} \vert$. 
$\bm{w}$ points to the maximum contrast direction of the geometry structure while $\bm{v}$ points to the minimum direction \cite{Shao2011TTHPES397}. 
Their corresponding eigenvalues $\lambda_1$ and $\lambda_2$ can be calculated via
\begin{equation}
	\lambda_1, \lambda_2 = \left( J_{11} + J_{22} \right) \pm \sqrt{ \left( J_{22} - J_{11} \right)^2 + 4 J_{12}^2 }.
	\label{eq:Structure_Eigenvalues}
\end{equation}
These two values can be used as two feature descriptors of the local geometry structure, where at the flat region, $\lambda_1 \approx \lambda_2 \approx 0$; at the edge region, $\lambda_1 \gg \lambda_2 \approx 0$; at the corner region, $\lambda_1 \ge \lambda_2 \gg 0$. 
For the sake of low computational cost, we take $\lambda_1 - \lambda_2$ as the edge awareness feature since its value of the edge pixel is much larger than that belongs to the flat region and corner.
By applying \cref{eq:Structure_Tensor} and \cref{eq:Structure_Eigenvalues} to every pixel in the input image $f_{\text{F}}$, two matrices $\bm{L}_1$ and $\bm{L}_2$ can be obtained, which consists of the large and small structure tensor eigenvalues of all the pixels, respectively. 
Then we can transform $\bm{L}_1$ and $\bm{L}_2$ to their corresponding patch-tensors $\bm{\mathcal{L}}_1$ and $\bm{\mathcal{L}}_2$. 
Finally, we define the local structure weight patch-tensor as follows
\begin{equation}
	\bm{\mathcal{W}}_{\mathrm{LS}} = \text{exp} \left( h \cdot \frac{ \left( \bm{\mathcal{L}}_1 - \bm{\mathcal{L}}_2 \right) - d_{\min} }{d_{\max} - d_{\min}} \right),
	\label{eq:W_LS}
\end{equation}
where $h$ is a weight stretching parameter, $d_{\max}$ and $d_{\min}$ are the maximum and minimum of $\bm{\mathcal{L}}_1 - \bm{\mathcal{L}}_2$, respectively. 
\cref{fig:Structure} displays the edge awareness maps of \cref{fig:Original}, which demonstrates that the structure tensor based local structure weight has a good performance in identifying the edges. 
It should be noticed that for the sake of displaying effect, \cref{fig:Structure} is created via a normalized version of $-\text{exp} \left( -h \cdot \frac{ \left( \bm{\mathcal{L}}_1 - \bm{\mathcal{L}}_2 \right) - d_{\min} }{d_{\max} - d_{\min}} \right)$. 
In the proposed algorithm, we still calculate the local structure weight via \cref{eq:W_LS}.

With the help of $\bm{\mathcal{W}}_{\mathrm{LS}}$, we can rewrite \cref{eq:IPT_model} into a weighted IPT model:
\begin{equation}
	\min_{\bm{\mathcal{B}}, \bm{\mathcal{T}}} \sum_{i=1}^{3} \Vert \bm{B}_{(i)} \Vert_* + \lambda\Vert\bm{\mathcal{W}}_{\mathrm{LS}} \odot \bm{\mathcal{T}}\Vert_1, ~ \mathrm{s.t.} ~ \Vert \bm{\mathcal{F}} - \bm{\mathcal{B}} - \bm{\mathcal{T}} \Vert_{\textrm{F}} \le \delta,
	\label{eq:WIPT_model}
\end{equation}
where $\odot$ is the notation of Hadamard product. 
Thanks to the weighted IPT model (\ref{eq:WIPT_model}), the strong edges could be well suppressed in the target image. 
\section{Reweighted infrared patch-tensor model and its solution}\label{sec:RIPT}
% In this section, we further incorporate the local information into our proposed model to improve its performance. 
% Then the optimization problem will be solved via alternatively updating the background patch-tensor and target patch-tensor. 
% Finally, the whole procedure of detecting the infrared small target in a single image is presented.

\subsection{Reweighted infrared patch-tensor model}
% 通常 RPCA 的收敛条件是满足误差重构小于多少
% WIPT 仍然需要很多次迭代才行
% 有趣的现象是在进入这个之前，E 的非零数字往往已经不动了。
% 考虑到小目标占主要成分，既然这些都不改变小目标，那么直接以非零数不再变化即可。
% 事实上，在目标图像中，true target 占据了最亮的那几个，在后面的迭代过程中，他们的数值变化很小。
% 因此直接以非零数不再变化即可。
% 由此，target image 的 sparsity degree 就比较重要了。
% 最理想的情况当然是，只有小目标是非零的，也就是说需要促进稀疏性
The computing time is also a major concern in infrared small target detection.
Generally, the stopping criterion of a RPCA algorithm is that the reconstruction error is less than a certain small value.
To meet this criterion, WIPT needs dozens of iterations, which is still time-consuming.
An interesting phenomenon we find is that the non-zero entry number in the target patch-tensor has ceased to grow before the algorithm converges.
In fact, in the target image, true target merely occupies the brightest of the few. 
In the second half of the iteration, their values barely change.
Therefore, considering the particularity of infrared small target, it is reasonable to replace the reconstruction error with the target patch-tensor sparsity in our proposed model. 
The algorithm stops iteration once the non-zero entry ceases to grow.
Then the sparsity of the target patch-tensor becomes critical in reducing the computing time.
We hope the non-zero entries keep decreasing as the iteration goes on, leaving the final target image the sparsest.
Unfortunately, the real situation is against our expectation in IPT and WIPT, where the target image deteriorates as the algorithm converges. 
Naturally, it motivates us to take a sparsity enhancing approach to solve this problem.

In Ref.\ \cite{Candes2008JoFAaA145}, Cand\`es proposed a reweighted  $\ell_1$ minimization for enhancing sparsity. 
By minimizing a sequence of weighted $\ell_1$ norm, a significant performance improvement is obtained on sparse recovery. 
Inspired by it, many improved RPCA models have been proposed \cite{Peng2014IToC4412,Xie2016IToIP2510,Lu2016IToIP252,Gu2016IJoCV}. 
Motivated by these state-of-the-art models, we adopt a similar reweighted scheme for the values in the target patch-tensor. 
The large weights discourage non-zero entries, and the small weights preserve non-zero elements. 
The sparsity enhancement weight is defined as follows 
\begin{equation}
	\bm{\mathcal{W}}_{\mathrm{SE}}^{k+1}(i,j,p) = \frac{1}{\vert \bm{\mathcal{T}}^{k}(i,j,p) \vert + \epsilon},
	\label{eq:Init_W_SE}
\end{equation}
where $\epsilon > 0$ is a preset constant to avoid division by zero. 
Then besides the relative error $\Vert \bm{\mathcal{F}} - \bm{\mathcal{B}}^{k+1} - \bm{\mathcal{T}}^{k+1} \Vert_{\text{F}} / \Vert \bm{\mathcal{F}} \Vert_{\text{F}} < \varepsilon$, we could add a new end condition that counts the non-zero entry element, namely $\Vert \bm{\mathcal{T}}^{k+1} \Vert_0 = \Vert \bm{\mathcal{T}}^{k} \Vert_0$.
With the help of this empirical observation, the computing time could be largely decreased, as illustrated in \cref{fig:Sparsity} and \cref{tab:Time}.

Another intrinsic characteristic that both Ref.\ \cite{Gao2013IToIP2212} and Ref.\ \cite{Dai2016IPT77} neglect is the fact that the small target is always brighter than its neighborhood environment in infrared images due to the physical imaging mechanism \cite{Gao2016N212}. 
Therefore, besides the sparsity constraint \cite{Chen2016IGaRSL1310,Chen2017IJoSTiAEOaRS101} of the target patch-tensor, it is reasonable to assume that all the entries in $\bm{\mathcal{T}}$ are non-negative. 
To this end, we incorporate this target non-negative prior into the reweighted IPT model via rewriting \cref{eq:Init_W_SE} as follows
\begin{align}
	\bm{\mathcal{W}}_{\mathrm{SE}}^{k+1}(i,j,p)       & = \delta \left( \bm{\mathcal{T}}^k (i,j,p) \right) \notag \\
	                                                  & =                                                         
	\begin{cases}
	\frac{1}{\bm{\mathcal{T}}^{k}(i,j,p) + \epsilon}, & \text{if } \bm{\mathcal{T}}^{k}(i,j,p) > 0;               \\
	\infty,                                           & \text{if } \bm{\mathcal{T}}^{k}(i,j,p) \le 0,              
	\end{cases}
	\label{eq:W_SE}
\end{align}
where $\delta\left(\cdot\right)$ is an indicator function. 
We combine the local structure weight $\bm{\mathcal{W}}_{\mathrm{LS}}$ and sparsity enhancing weight $\bm{\mathcal{W}}_{\mathrm{SE}}^{k}$ to get the adaptive weight as follows
\begin{equation}
	\bm{\mathcal{W}}^{k} = \bm{\mathcal{W}}_{\mathrm{LS}} \odot \bm{\mathcal{W}}_{\mathrm{SE}}^{k}.
	\label{eq:W}
\end{equation}
Finally, we generalize the proposed IPT model and WIPT model to a novel reweighted infrared patch-tensor model (RIPT) as follows
\begin{equation}
	\min_{\bm{\mathcal{B}}, \bm{\mathcal{T}}} \sum_{i=1}^{3} \Vert \bm{B}_{(i)} \Vert_* + \lambda\Vert\bm{\mathcal{W}} \odot \bm{\mathcal{T}}\Vert_1, ~ \mathrm{s.t.} ~ \bm{\mathcal{B}} + \bm{\mathcal{T}} = \bm{\mathcal{F}}.
	\label{eq:Convex_RIPT}
\end{equation}

\subsection{Solution of RIPT model} \label{subsec:solution}
In this subsection, we show how to solve the proposed RIPT model as a reweighted robust tensor recovery problem via an Alternating Direction Method of Multipliers (ADMM) \cite{Boyd2011FaTiML31}. 
The augmented Lagrangian function of \cref{eq:Convex_RIPT} is defined as 
\begin{align}
	\mathcal{L} = & \sum_{i=1}^{N} \Vert \bm{B}_{i,(i)} \Vert_* +                   
	\lambda\Vert \bm{\mathcal{W}} \odot \bm{\mathcal{T}} \Vert_{1} + \notag \\ 
	              & \sum_{i=1}^{N} \frac{1}{2\mu} \Vert \bm{\mathcal{B}}_i + 
	\bm{\mathcal{T}} - \bm{\mathcal{F}}\Vert^2 - \left\langle\bm{\mathcal{Y}}_{i},
	\bm{\mathcal{B}}_i + \bm{\mathcal{T}} - \bm{\mathcal{F}}
	\right\rangle ,
	\label{eq:Lagrange}
\end{align}
where $\bm{\mathcal{Y}}_{i} \in \mathbb{R}^{I\times J\times P}, \ i = 1, 2, 3$ are the Lagrange multiplier tensors, and $\mu$ is a positive penalty scalar. 
ADMM decomposes the minimization of $\mathcal{L}$ into two subproblems that minimize $\bm{\mathcal{B}}_i$ and $\bm{\mathcal{T}}$, respectively. More specifically, the iterations of ADMM go as follows:

Updating $\bm{\mathcal{B}}_i$ with the other terms fixed.
\begin{align}
	\bm{\mathcal{B}}_i^{k+1} = \arg\min_{\bm{\mathcal{B}}_i}                                                                                                                                  
	\left\Vert \bm{B}_{i,(i)} \right\Vert_* + \frac{1}{2\mu} \left\Vert \bm{\mathcal{B}}_i - (\bm{\mathcal{F}} + \mu\bm{\mathcal{Y}}_{i}^{k} - \bm{\mathcal{T}}^{k}) \right\Vert^2_{\text{F}} 
	\label{eq:B_subproblem}
\end{align}

Updating $\bm{\mathcal{T}}$ with the other terms fixed.
\begin{align}
	\bm{\mathcal{T}}^{k+1} = & \arg\min_{\bm{\mathcal{T}}} \lambda \left\Vert                        
	\bm{\mathcal{W}}^{k} \odot \bm{\mathcal{T}} \right\Vert_{1} +   \notag \\
	                         & \sum_{i=1}^{N} \frac{1}{2 \mu} \left\Vert \bm{\mathcal{T}} - \left( 
	\bm{\mathcal{F}} + \mu \bm{\mathcal{Y}}_{i}^{k} - \bm{\mathcal{B}}_i^{k+1}
	\right) \right\Vert^2_{\text{F}}
	\label{eq:T_subproblem}
\end{align}

Updating the multiplier $\bm{\mathcal{Y}}_i^{k+1}$ with the other terms fixed.
\begin{equation}
	\bm{\mathcal{Y}}_i^{k+1} = \bm{\mathcal{Y}}_i^{k} + \frac{1}{\mu^k} \left(
	\bm{\mathcal{F}} - \bm{\mathcal{B}}_i^{k+1} - \bm{\mathcal{T}}^{k+1} \right)
	, ~ i = 1, \cdots, N.
	\label{eq:Update_Y}
\end{equation}

The subproblems (\ref{eq:B_subproblem}) and (\ref{eq:T_subproblem}) can be solved via the following two operators, respectively. 
\begin{align}
	\bm{\mathcal{B}}_i^{k+1} & = \mathcal{T}_{i, \mu} \left( \bm{\mathcal{F}} + \mu \bm{\mathcal{Y}}_{i}^{k} - \bm{\mathcal{E}}^{k} \right) \label{eq:B_solver} \\  
	\bm{\mathcal{T}}^{k+1}   & = {\mathcal{S}}_{ \frac{\mu \lambda}{N}  \bm{\mathcal{W}}^{k} } \left[ \frac{1}{N} \sum_{i=1}^{N} \left(                         
	\bm{\mathcal{F}} + \mu \bm{\mathcal{Y}}_{i}^{k} - \bm{\mathcal{B}}_i^{k+1}
	\right) \right]
	\label{eq:T_solver}
\end{align}
From \cref{eq:T_solver}, it could be observed that the weighting parameter determines the soft-threshold, controlling the trade-off between the target patch-tensor and background patch-tensor. 
Therefore, our element-wise adaptive weight tensor could simultaneously preserve the small target and suppress the strong edges. Finally, the solution of the proposed model is given in \cref{alg:alg1}. 
\begin{algorithm}[htbp]
	\caption{Target-Background Separation via RIPT model}
	\label{alg:alg1}
	\begin{algorithmic}
		% Input
		\renewcommand{\algorithmicrequire}{\textbf{Input:}}
		\REQUIRE $\bm{\mathcal{F}}, \ \bm{\mathcal{W}}_{\text{LS}}, \ \lambda, \ \varepsilon, \ N$
		% Initialize
		\renewcommand{\algorithmicrequire}{\textbf{Initialize:}}
		\REQUIRE $\bm{\mathcal{T}}^0 := \bm{0}$;
		$\bm{\mathcal{B}}_i^0 := \bm{\mathcal{F}}, \bm{\mathcal{Y}}_{i}^0 := \bm{0}, i := 1,\cdots,N$;
		$\bm{\mathcal{W}}_{\text{SE}}^0 := \bm{1}, \bm{\mathcal{W}}^0 := \bm{\mathcal{W}}_{\text{LS}} \odot \bm{\mathcal{W}}_{\text{SE}}^0$;
		$\mu := 5 \cdot \text{std}(\text{vec}(\bm{\mathcal{F}}))$, $k := 0$
		% while loop
		\WHILE{not converged}
		% Step 1: Update B
		\STATE $\vartriangleright$ \textbf{Step 1:} Fix the others and update $\bm{\mathcal{B}}_i$ by
		\FOR{$i=1$ to $N$}
		\STATE $\bm{\mathcal{B}}_i^{k+1} := \mathcal{T}_{i, \mu} \left( \bm{\mathcal{F}} + \mu \bm{\mathcal{Y}}_{i}^{k} - \bm{\mathcal{E}}^{k} \right)$;
		\ENDFOR
		% Step 2: Update T
		\STATE $\vartriangleright$ \textbf{Step 2:} Fix the others and update $\bm{\mathcal{T}}$ by
		\STATE $\bm{\mathcal{T}}^{k+1} := {\mathcal{S}}_{ \frac{\mu \lambda}{N} \bm{\mathcal{W}}^{k} } \left[ \frac{1}{N} \sum_{i=1}^{N} \left(
			\bm{\mathcal{F}} + \mu \bm{\mathcal{Y}}_{i}^{k} - \bm{\mathcal{B}}_i^{k+1}
		\right) \right]$;
		% Step 3: Update Y
		\STATE $\vartriangleright$ \textbf{Step 3:} Fix the others and update $\bm{\mathcal{Y}}_i$ by
		\FOR{$i=1$ to $N$}
		\STATE $\bm{\mathcal{Y}}_i^{k+1} := \bm{\mathcal{Y}}_i^{k} + \frac{1}{\mu^k} \left( \bm{\mathcal{F}} - \bm{\mathcal{B}}_i^{k+1} - \bm{\mathcal{T}}^{k+1} \right)$;
		\ENDFOR
		% Step 4: Update W
		\STATE $\vartriangleright$ \textbf{Step 4:} Fix the others and update $\bm{\mathcal{W}}$ by
		\FOR{$(i,j,p) \in [1,\dots,I] \times [1,\dots,J] \times [1,\dots,P]$}
		\STATE $\bm{\mathcal{W}}_{\mathrm{SE}}^{(k+1)}(i,j,p) := \delta \left( \bm{\mathcal{T}}^k (i,j,p) \right)$;
		\ENDFOR
		\STATE $\bm{\mathcal{W}}^{k+1} := \bm{\mathcal{W}}_{\mathrm{LS}} \odot \bm{\mathcal{W}}_{\mathrm{SE}}^{k+1}$;
		% Step 5: Update Y
		\STATE $\vartriangleright$ \textbf{Step 5:} Update $\mu$ by
		\STATE $\mu^{k+1} := \mu^k / \rho$;
		% Step 6: Check convergence
		\STATE $\vartriangleright$ \textbf{Step 6:} Check the convergence conditions
		\STATE $\frac{ \left\Vert \bm{\mathcal{F}} - \bm{\mathcal{B}}^{k+1} - \bm{\mathcal{T}}^{k+1} \right\Vert_{\text{F}} }{ \left\Vert \bm{\mathcal{F}} \right\Vert_{\text{F}} } < \varepsilon \quad \textrm{or} \quad \left\Vert \bm{\mathcal{T}}^{k+1} \right\Vert_0 = \left\Vert \bm{\mathcal{T}}^{k} \right\Vert_0$;
		% Step 7: Update k
		\STATE $\vartriangleright$ \textbf{Step 7:} Update $k$
		\STATE $k := k + 1$;
		\ENDWHILE
		\renewcommand{\algorithmicensure}{\textbf{Output:}}
		\ENSURE $\frac{1}{N} \left( \sum_{i=1}^{N} \bm{\mathcal{B}}_i^{k} \right), ~ \bm{\mathcal{T}}^{k}$
	\end{algorithmic}
\end{algorithm}

\subsection{Detection Procedure}
\begin{figure*}[htbp]
  \centering
  \includegraphics[width=0.95\textwidth]{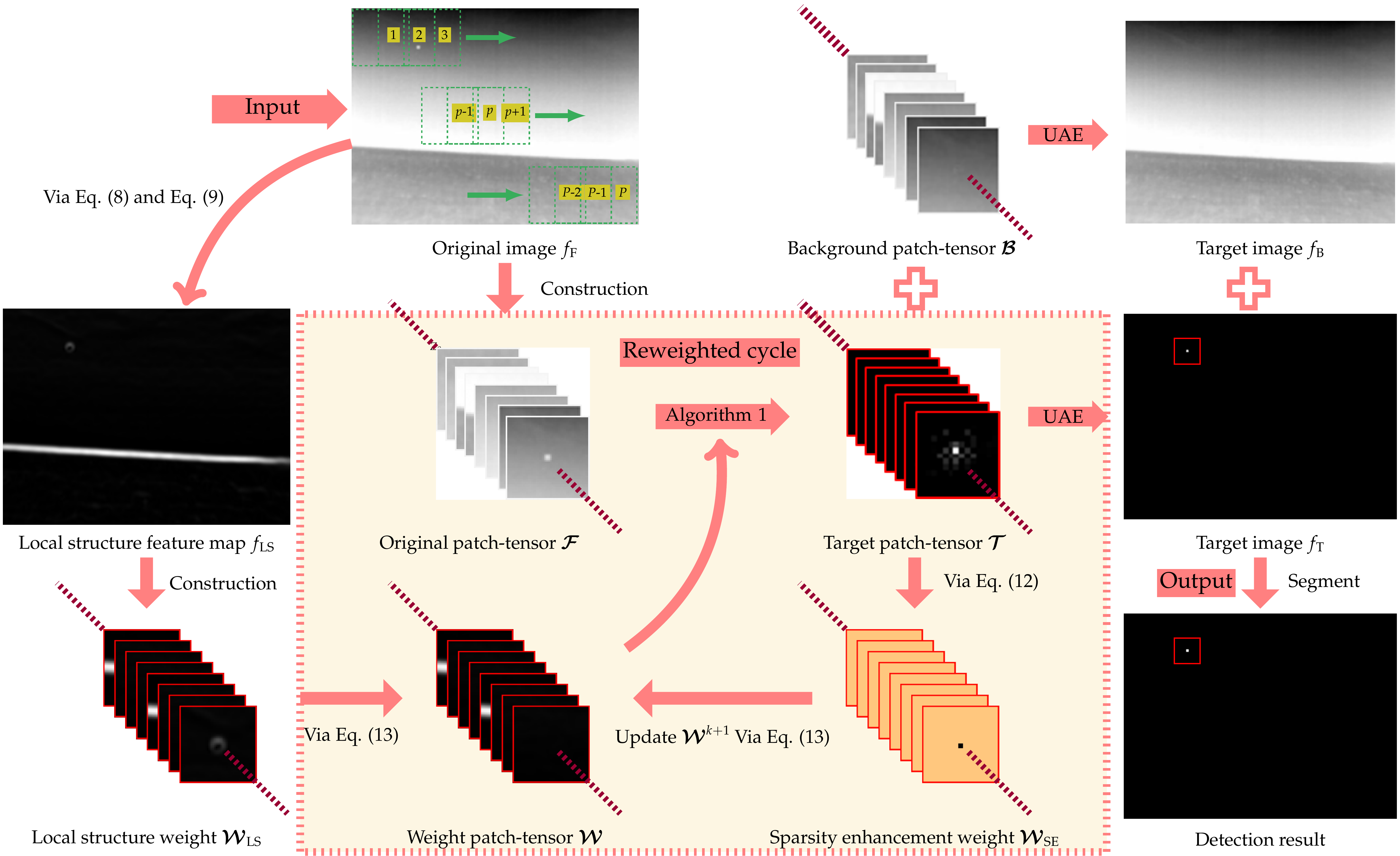}
  \caption{The overview of the proposed reweighted infrared patch-tensor model in this paper.}
  \label{fig:Overview} 
\end{figure*}

In \cref{fig:Overview}, we present the whole procedure of detecting the infrared small target via the model proposed in this paper. The detailed steps are as follows:
\begin{enumerate}
  \item Given an infrared image $f_{\textup{F}}$, its local structure feature map $f_{\textup{LS}}$ is calculated via \cref{eq:Structure_Eigenvalues}. 
  \item The original patch-tensor $\bm{\mathcal{F}}$ and local structure weight patch-tensor $\bm{\mathcal{W}}_{\mathrm{LS}}$ are constructed from $f_{\textup{F}}$ and $f_{\textup{LS}}$. 
  \item \cref{alg:alg1} is performed to decompose the patch-tensor $\bm{\mathcal{F}}$ into the background patch-tensor $\bm{\mathcal{B}}$ and target patch-tensor $\bm{\mathcal{T}}$. 
  \item The background image $f_{\textup{B}}$ and target image $f_{\textup{T}}$ are reconstructed from the background patch-tensor $\bm{\mathcal{B}}$ and target patch-tensor $\bm{\mathcal{T}}$, respectively. For the sake of implementation convenience, we adopt the uniform average of estimators (UAE) reprojection scheme \cite{Salmon2012SP922}. 
  \item The target is segmented out as the same as Ref.\ \cite{Gao2013IToIP2212}. The adaptive threshold is determined by: 
  \begin{equation}  
t_{\textup{up}} = \max(v_{\min}, \bar{f}_{\textup{T}} + k\sigma),
\end{equation}  
where $\bar{f}_{\textup{T}}$ and $\sigma$ are the average and standard deviation of the target image $f_{\textup{T}}$.
$k$ and $v_{\min}$ are constants determined empirically.
\end{enumerate}
% First, given an infrared image $f_{\textup{F}}$, its local structure feature map $f_{\textup{LS}}$ is calculated via \cref{eq:Structure_Eigenvalues}. 
% Second, the original patch-tensor $\bm{\mathcal{F}}$ and local structure weight patch-tensor $\bm{\mathcal{W}}_{\mathrm{LS}}$ are constructed from $f_{\textup{F}}$ and $f_{\textup{LS}}$. 
% Third, \cref{alg:alg1} is performed to decompose the patch-tensor $\bm{\mathcal{F}}$ into the background patch-tensor $\bm{\mathcal{B}}$ and target patch-tensor $\bm{\mathcal{T}}$. 
% Fourth, the background image $f_{\textup{B}}$ and target image $f_{\textup{T}}$ are reconstructed from the background patch-tensor $\bm{\mathcal{B}}$ and target patch-tensor $\bm{\mathcal{T}}$, respectively. For the sake of implementation convenience, we adopt the uniform average of estimators (UAE) reprojection scheme \cite{Salmon2012SP922} to undertake the reconstruction task. 
% The last step is the segmentation. On the basis of previous successful separation, it would be very easy. We can either take the thresholding approach given in \cite{Gao2013IToIP2212} or simply take the brightest connected domains as the candidate targets. 

\section{Experimental validation}\label{sec:Experiment}
To fully evaluate the proposed algorithm, we conduct a series of experiments using images of various scenarios and include ten state-of-the-art methods for comparison.
% In this section, the proposed RIPT model would be evaluated with extensive experiments. 
% Firstly, we provide some preparatory information about the experiment such as the evaluation metrics, the comparing methods, and the details of the tested images. 
% Second, the convergence of our solving algorithms is inspected. 
% Third, we discuss the impact of several important parameters, including the patch size, step length, and the regulation parameters $\lambda$, $\mu$. 
% Then we present a lot of the target images separated via our proposed RIPT model to verify its robustness. 
% Finally, we compare the proposed model with other state-of-the-art methods for both background suppression and detection on several real infrared sequences to demonstrate its superiority as well as the necessity of our improvement. 

\subsection{Experimental setup}

\begin{figure*}[htbp]
  \centering
  \includegraphics[width=0.95\textwidth]{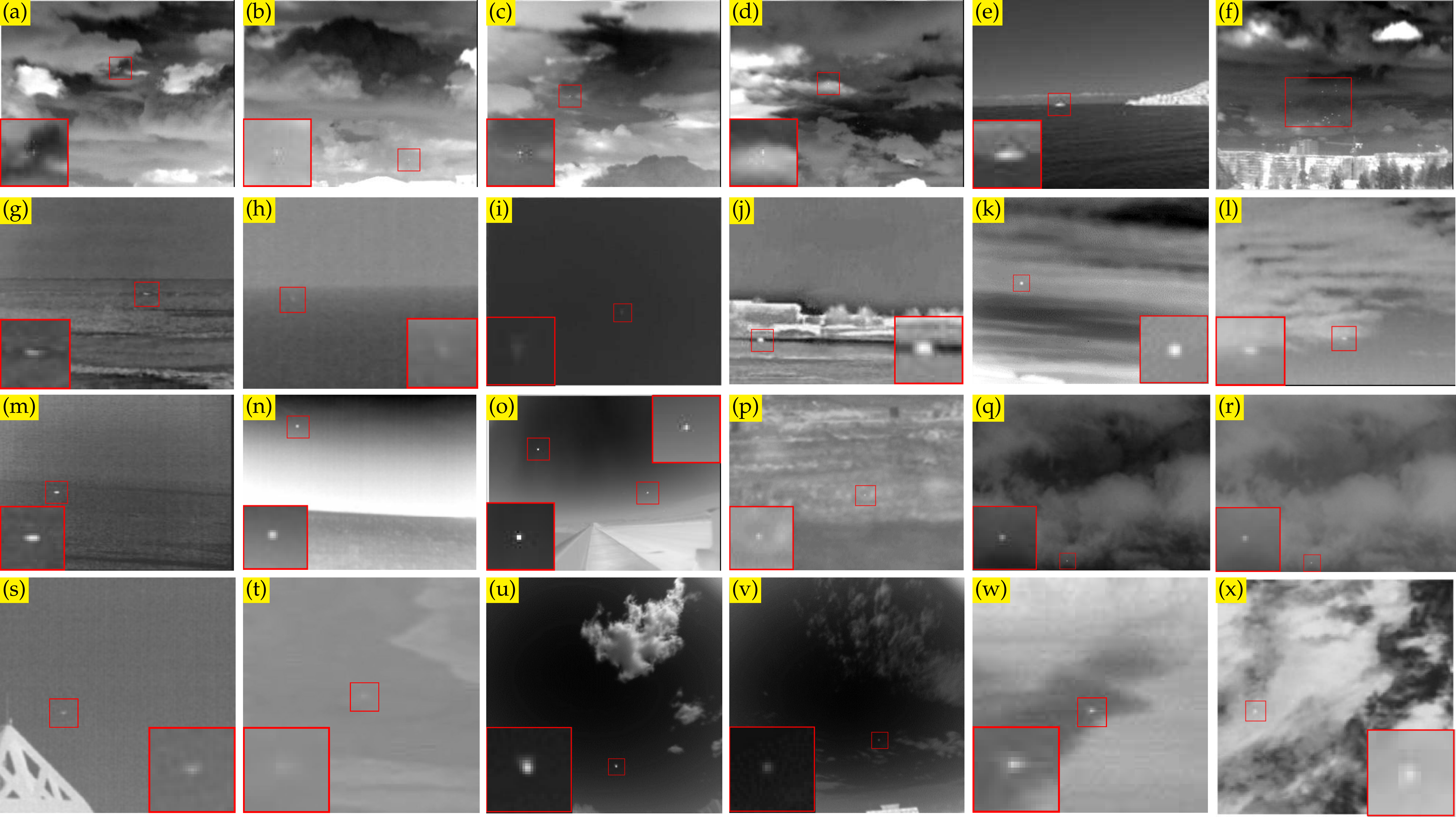}
  \caption{The representative infrared images for experiments. For better visualization, the demarcated area is enlarged, which is better to be seen by zooming on a computer screen. }
  \label{fig:Original}
\end{figure*}

\textbf{Datasets}.
We test the proposed model on extensive real infrared images to cover different scenarios, as illustrated in \cref{fig:Original}, varying from the flat backgrounds with salient targets to complex backgrounds with heavy clutters and extremely dim targets.
All targets are labeled with red boxes. 
Since some targets are so dim that could hardly be observed by human eyes directly, we enlarge the demarcated area.
Taking into account that the biggest difficulty of current infrared small target detection is how to detect those very dim targets with strong clutters, a good detection performance on those extremely complex images is more convincing than the satisfying result on relatively simple images. 
Therefore, in the following experiments, our main focus is put on the datasets with complex scenes that \cref{fig:Original}(a) -- (d) and (l) belong to. 
The detailed characteristics of these five sequences are presented in \cref{tab:Characteristic}.
% 重点说一下包含各种各样的场景，以回复审稿人提的问题
% 包含了各种各样的场景，简单一点的，比如 Set 10 23，复杂的，肉眼不可见的 Set 2 XXX，
% 但本文重点研究 目标非常微弱，以及 有强干扰的场景，比如 Set 2 XXX 这样的
% 后面的 Sequence 也是 Set 2 这些
% It is fair to say that the backgrounds in the images of last four rows are much more complicated than those in the first row. 

\begin{table*}[htbp]  
  \centering
  \caption{Target and background characteristics of real infrared image sequences}
  \begin{tabularx}{\textwidth}{l l l l >{\raggedright}X X}
    \toprule
               & \# Frame & Image Resolution & Target Shape & Target characteristics & Background characteristics \\
    \midrule
    Sequence 1-4 & 400      & $255\times 320$  & Gaussian & \specialcell[t]{\myfourstar Tiny, very dim with low contrast. \\ \myfourstar Moves along the cloud edges or \\ \quad \ buried in the cloud.} & \specialcell[t]{\myfourstar  Strong undulant cloud backgrounds. \\   \myfourstar Approximately noise-free.}  \\    
    % \midrule
    Sequence 5 & 30      & $200\times 256$  & Rectangular & \specialcell[t]{\myfourstar Small due to long imaging distance.  \\ \myfourstar Brightness, contrast, and size vary a lot.} & \specialcell[t]{\myfourstar  Heavy cloudy-sky background clutters\\ \quad \ in uniform backgrounds with noise.}  \\ 
    \bottomrule
  \end{tabularx}
  \label{tab:Characteristic}
\end{table*}

\textbf{Baselines and Parameter settings}. 
The proposed algorithm is compared with ten state-of-the-art solutions, including three filtering based methods (Max-Median \cite{Deshpande1999SISoOSEaI}, Top-Hat \cite{Rivest1996OE357}, TDLMS \cite{Hadhoud1988IToCaS355}), three HVS based methods (PFT \cite{Guo20082ICoCVaPR}, MPCM \cite{Wei2016PR58}, WLDM \cite{Deng2016IToGaRS547}), and four recently developed low-rank methods (IPI \cite{Gao2013IToIP2212}, PRPCA \cite{Wang2015IPT69}, WIPI \cite{Dai2016IPT77}, NIPPS \cite{Dai2017IP&T81}). 
\cref{tab:Parameters} summarizes all the methods involved in the experiments and their detailed parameter settings. 
For all the low-rank methods, i.e.\ IPI, PRPCA, WIPI, NIPPS, IPT, and RPIT, they are all solved via ADMM. 
All the algorithms are implemented in MATLAB 2016b on a PC of 3.4 GHz and 4GB RAM.
The source code of our method is publicly available at \url{https://github.com/YimianDai/DENTIST}.

% Parameter setting table
\begin{table*}[htbp]  
  \setlength{\tabcolsep}{4pt}
  \centering
  \caption{Detailed parameter settings for twelve tested methods}
  \begin{tabularx}{\textwidth}{l >{\raggedright\hsize=.3\textwidth}X >{\hsize=.09\textwidth}X X }
    \toprule
    No. & Methods & Acronyms   & Parameter settings \\
    \midrule
    % 1 - Max-Median
    1   & Max-Median filter \cite{Deshpande1999SISoOSEaI} & Max-Median & Support size: $5\times 5$ \\
    % 2 - Top-Hat
    2   & Top-Hat method \cite{Rivest1996OE357} & Top-Hat    & Structure shape: square, structure size: $3\times 3$ \\
    % 3 - PFT
    3   & Phase spectrum of Fourier Transform \cite{Guo20082ICoCVaPR} & PFT & Disk radius: $3$ \\    
    % 4 - MPCM
    4   & Multiscale Patch-based Contrast Measure \cite{Wei2016PR58} & MPCM & $N = 1,3,\ldots,9$ \\
    % 6 - WLDM
    5   & Weighted Local Difference Measure \cite{Deng2016IToGaRS547} & WLDM & $L = 4, m = 2, n = 2$ \\
    % 5 - TDLMS
    6   & Two-Dimensional Least Mean Square filter \cite{Hadhoud1988IToCaS355} & TDLMS & Support size: $5\times 5$, step size: $\mu = 5\times 10^{-8}$ \\
    % 7 - IPI
    7   & Infrared Patch-Image Model \cite{Gao2013IToIP2212} & IPI & Patch size: $50\times 50$, sliding step: $10$, $\lambda = \frac{L}{\sqrt{\min{(IJ,P)}}}$, $L = 3$, $\varepsilon = 10^{-7}$ \\
    % 8 - PRPCA
    8   & Patch-level RPCA method \cite{Wang2015IPT69} & PRPCA & Patch size: $50\times 50$, sliding step: $10$, $\lambda = \frac{L}{\sqrt{\min{(IJ,P)}}}$, $L = 3$, $\varepsilon = 10^{-7}$                                                   \\
    % 9 - WIPI
    9   & Weighted Infrared Patch-Tensor Model \cite{Dai2016IPT77} & WIPI & Patch size: $51\times 51$, sliding step: $10$, smoothing parameter $h = 15$, $\varepsilon = 10^{-7}$ \\
    10 & Non-negative IPI model via Partial Sum minimization of singular values \cite{Dai2017IP&T81} &
    NIPPS & Patch size: $50\times 50$, sliding step: $10$, $\lambda = \frac{L}{\sqrt{\min{(IJ,P)}}}$, $L = 2$, $r = 5\times 10^{-3}$ \\
    % 11 - IPT
    11   & Infrared Patch-Tensor Model & IPT & Patch size: $50\times 50$, sliding step: $10$, $\lambda = \frac{L}{\sqrt{\max{(I,J,P)}}}$, $L = 3$\\ %, maximum iterations $k_{\max} = 1$ 
    % 12 - RIPT
    12  & Reweighted Infrared Patch-Tensor Model & RIPT       & Patch size: $50\times 50$, sliding step: $10$, $\lambda = \frac{L}{\sqrt{\min{(I,J,P)}}}$, $L = 1$, $h = 10$, $\epsilon = 0.01$, $\varepsilon = 10^{-7}$                    \\
    \bottomrule
  \end{tabularx}
  \label{tab:Parameters}
\end{table*}

\begin{figure}[htbp]
	\centering
	\includegraphics[scale = 0.4]{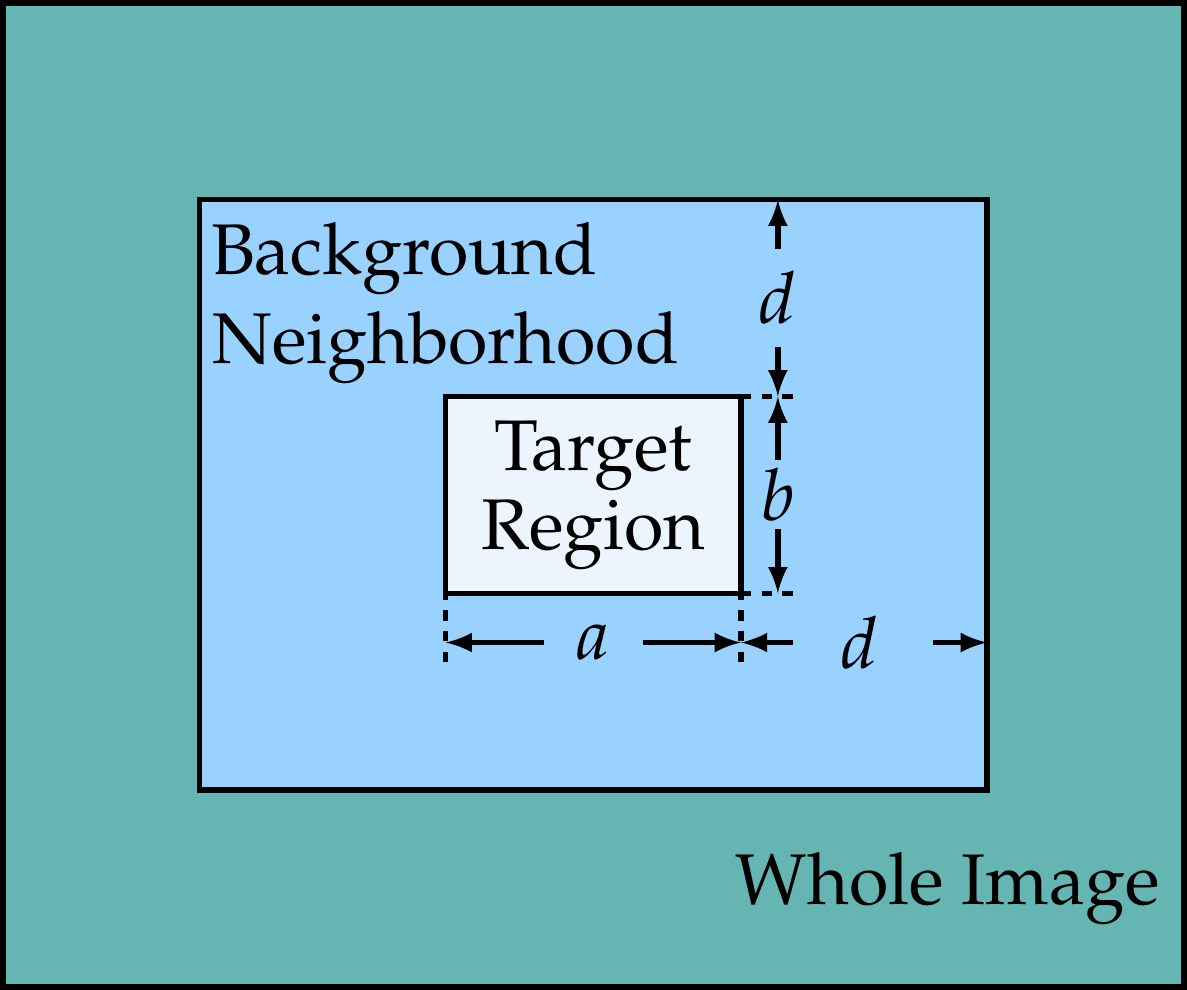}
	\caption{The target and background neighborhood regions of a small target.}
	\label{fig:Neighborhood} 
\end{figure}

\textbf{Evaluation Metrics}.
% For our infrared small target detection task, the biggest difficulty is the interference of complex backgrounds. These undesirable background clutters raise the false alarm rates of detection methods, and might even overwhelm the dim targets. Therefore, the ability of successfully suppressing the background clutters is the major concern in evaluating an infrared small target detection method. 
For a comprehensive evaluation, four metrics including the local signal to noise ratio gain (LSNRG), background suppression factor (BSF), signal to clutter ratio gain (SCRG), and receiver operating characteristic (ROC) curve are adopted in comparison of background suppressing performance. 
LSNRG measures the local signal to noise ratio (LSNR) gain, which is defined as
\begin{equation}
	\textup{LSNRG} = \frac{\textup{LSNR}_{\textup{out}}}{\textup{LSNR}_{\textup{in}}},
	\label{eq:LSNRG}
\end{equation}
where $\textup{LSNR}_{\textup{in}}$ and $\textup{LSNR}_{\textup{out}}$ are the LSNR values before and after background suppression. 
LSNR is given as $\textup{LSNR} = P_{\textup{T}} / P_{\textup{B}}$. $P_{\textup{T}}$ and $P_{\textup{B}}$ are the maximum grayscales of the target and neighborhood, respectively. 
BSF measures the background suppression quality using the standard deviation of the neighborhood region. It is defined as:
\begin{equation}
	\text{BSF} = \frac{\sigma_\text{in}}{\sigma_\text{out}},
	\label{eq:BSF}
\end{equation} 
where $\sigma_\text{in}$ and $\sigma_\text{out}$ are the standard variances of background neighborhood before and after background suppression. 
The most widely used SCRG is defined as the ratio of signal-to-clutter ratio (SCR) before and after processing:
\begin{equation}
	\textup{SCRG} = \frac{\text{SCR}_\text{out}}{\text{SCR}_\text{in}},
	\label{eq:SCRG}
\end{equation}
where SCR represents the difficulty of detecting the infrared small target, and it is defined by $\text{SCR} = \vert\mu_{\text{t}} - \mu_{\text{b}}\vert / \sigma_\text{b}$. 
$\mu_\text{t}$ is the average target grayscale. $\mu_\text{b}$ and $\sigma_\text{b}$ are the average grayscale and standard deviation of the neighborhood region. 
For all these three metrics, the higher their values are, the better background suppression performance the detection method has. 
All three metrics are computed in a local region, as illustrated in \cref{fig:Neighborhood}. 
The target size is $a \times b$, and $d$ is the neighborhood width.
we set $d = 20$ in this paper.
Among all the existing metrics, the detection probability $P_{\text{d}}$ and false-alarm rate $F_{\text{a}}$ are the key performance indicators, which are defined as follows
\begin{align}
	P_\text{d} & = \frac{\text{number of true detections}}{\text{number of actual targets}}, \\
	F_\text{a} & = \frac{\text{number of false detections}}{\text{number of images}}.        
\end{align}
The ROC curve shows the trade-off between the true detections and false detections. 

\subsection{Validation of the proposed method}
In this subsection, we take a closer look at the proposed method by validating their robustness against various scenes and noisy cases. 
At last, the roles of the patch-tensor, local structure weight, and sparsity enhancement weight are examined in depth to evaluate each prior individually.
\subsubsection{Robustness to various scenes}
\begin{figure*}[htbp]
  \centering
  \includegraphics[width=0.98\textwidth]{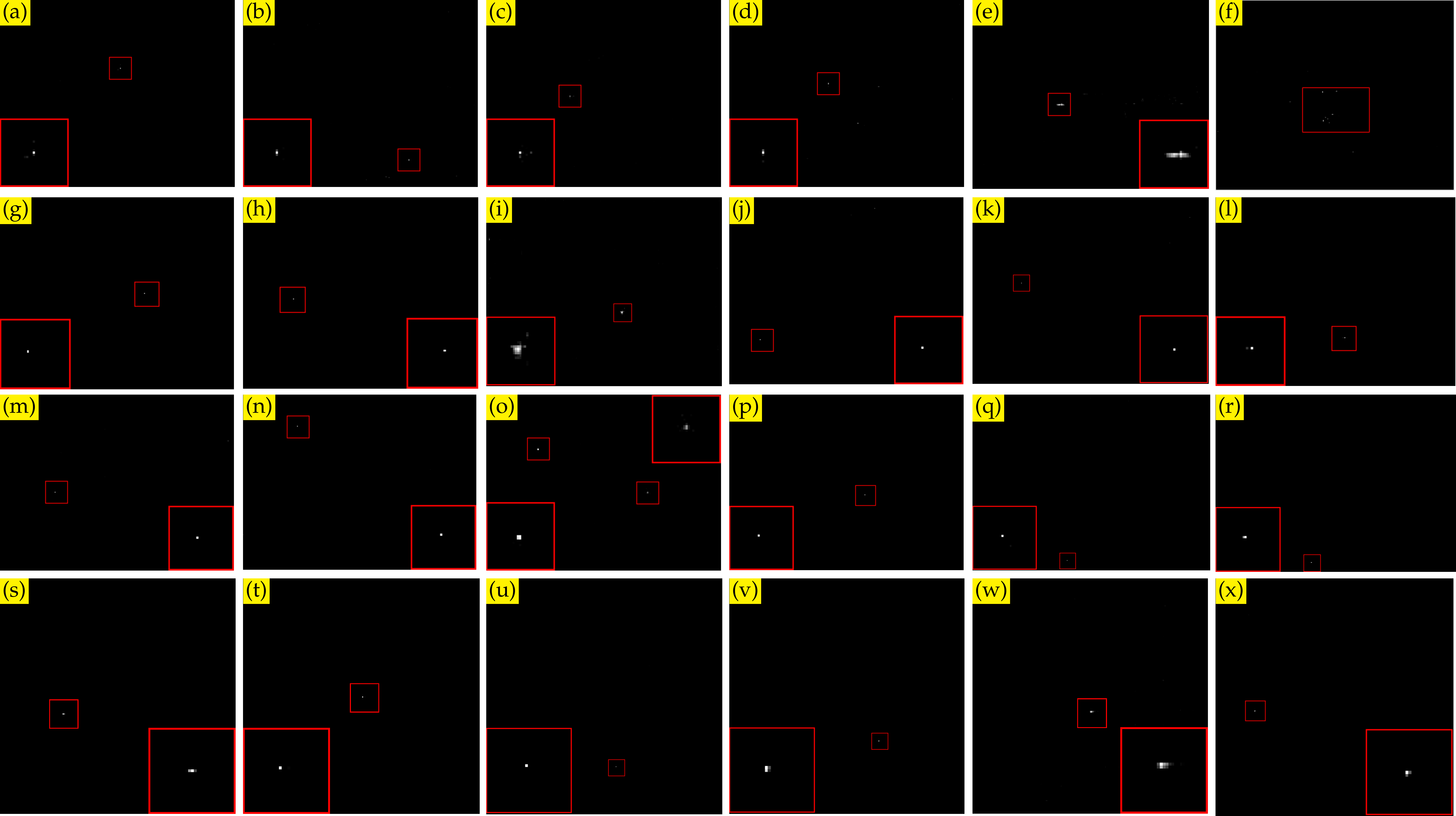}
  \caption{The separated target images by the proposed RIPT model for \cref{fig:Original}. 
  For better visualization, the demarcated area is enlarged, which is better to be seen by zooming on a computer screen.}
  \label{fig:Evaluation}
\end{figure*}

In \cref{fig:Evaluation}, we show the separated target images for the images of \cref{fig:Original}. 
Observing \cref{fig:Evaluation}, it can be clearly seen that the background clutters are completely wiped out, leaving the target the only sole component in the target image. 
Since \cref{fig:Original} contains a lot of different scenarios, it is fair to say that the proposed method is quite robust to various scenes.

\subsubsection{Robustness to noise}
\begin{figure*}[htbp]
  \centering
  \includegraphics[width=0.98\textwidth]{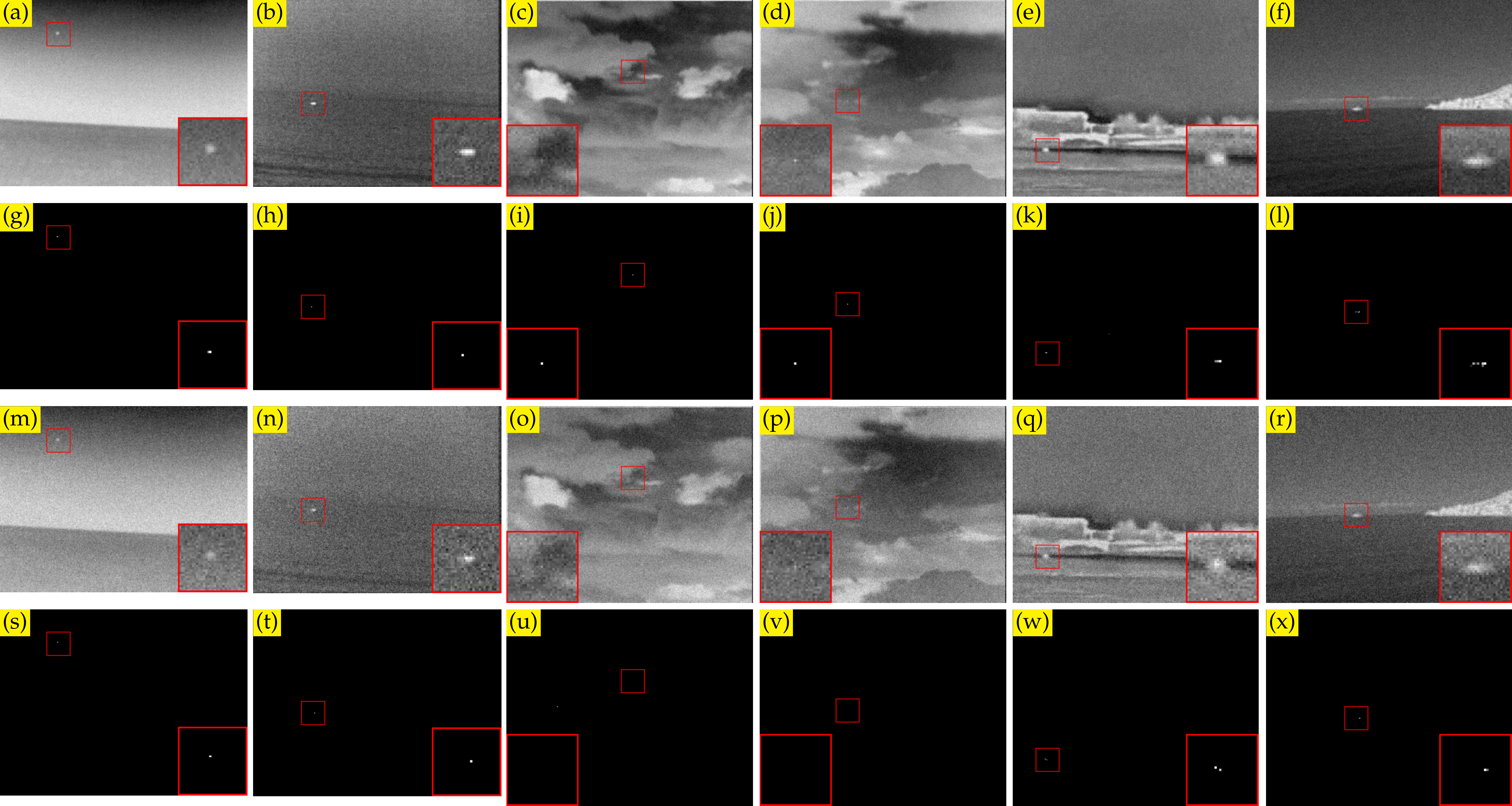}
  \caption{The first and third rows are images contaminated by additive white Gaussian noise with standard deviation of $10$ and $20$, and the second and fourth rows are the corresponding detection results by the proposed method.}
  \label{fig:Noise}
\end{figure*}

Noise is another key influence factor. 
In \cref{fig:Noise}, we evaluate the proposed method's performance in the case of noise with different levels. 
When the noise standard deviation is $10$, the proposed method could well enhance the targets and suppress the clutters and noise. 
As the noise standard deviation increases to $20$, RIPT still detects the target in \cref{fig:Noise}(m) -- (n) and (q) -- (r), but fails in \cref{fig:Noise}(o) -- (p). 
Nevertheless, this failure is acceptable, since the target is totally overwhelmed by the noise in \cref{fig:Noise}(o) -- (p) (see the enlarged box).
Therefore, the noise influence depends not only on the intensity of the noise itself but also on the original contrast of the target.
As long as the polluted target can maintain a weak contrast like \cref{fig:Noise}(c) -- (d), the proposed method is still able to detect it.

\subsubsection{Roles of components in the proposed model}
\begin{figure*}[htbp]
	\centering
	\subfloat[]{
		\includegraphics[width=0.23\textwidth]{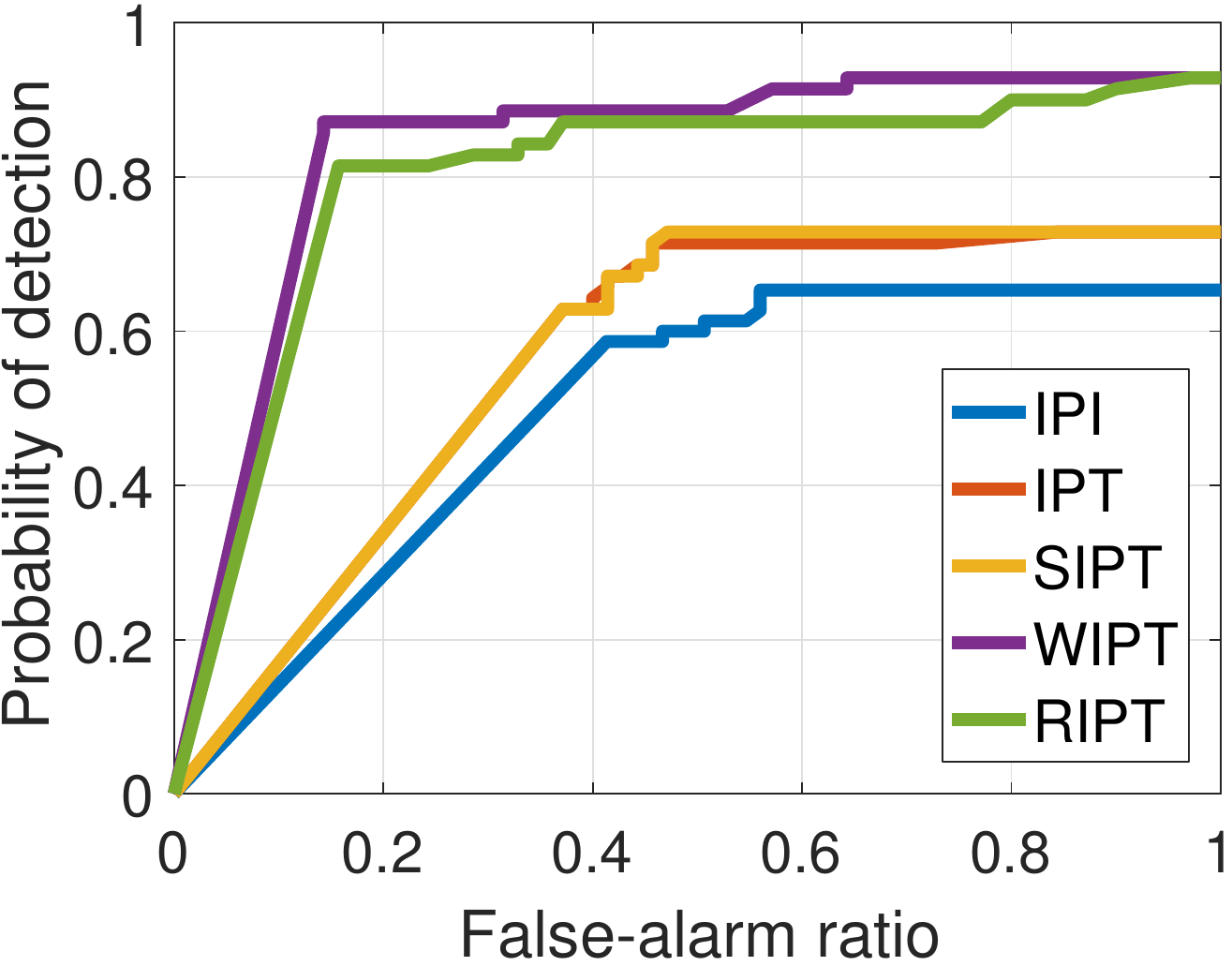}
	}
	\subfloat[]{
		\includegraphics[width=0.23\textwidth]{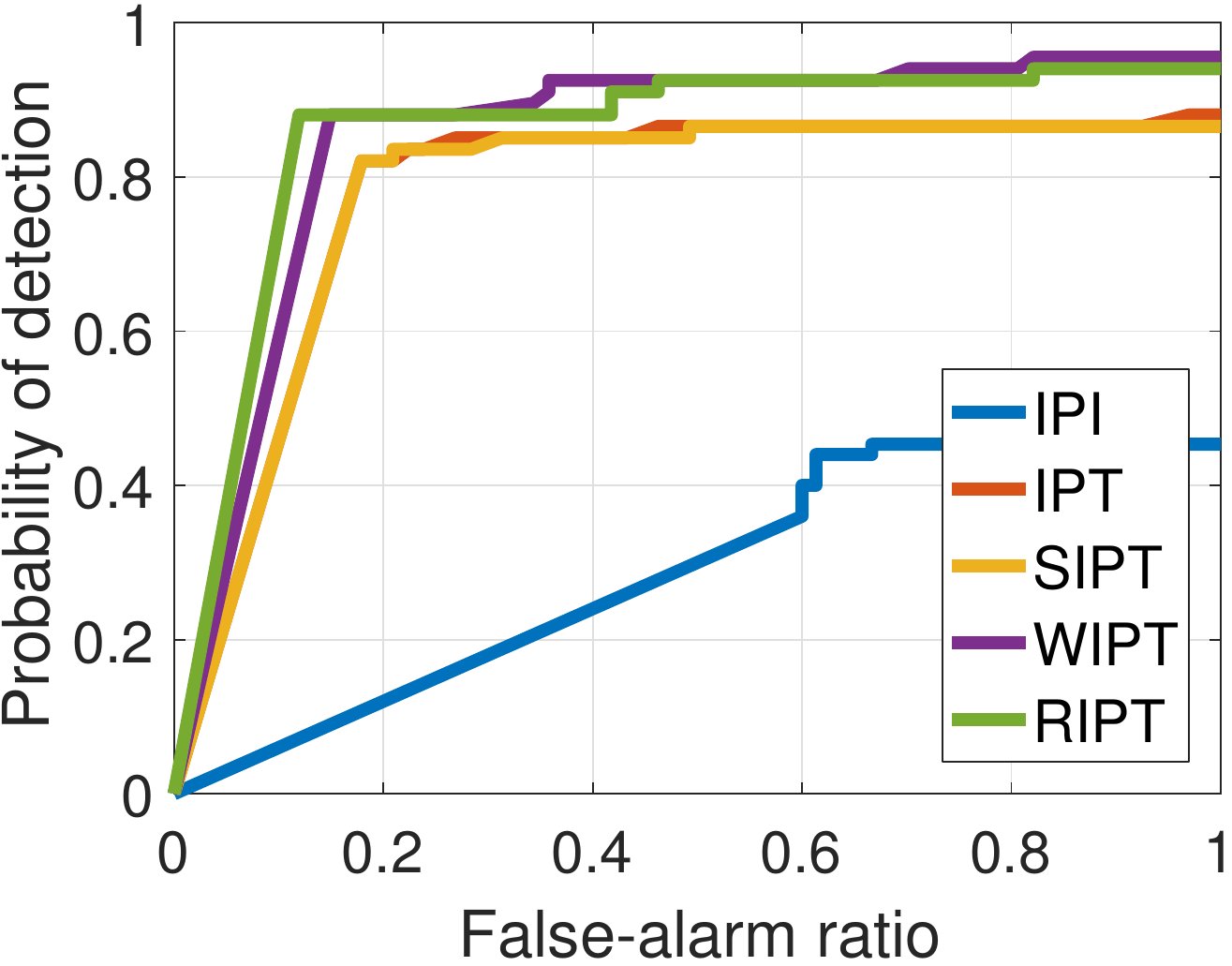}
	}                                           
	\subfloat[]{
		\includegraphics[width=0.23\textwidth]{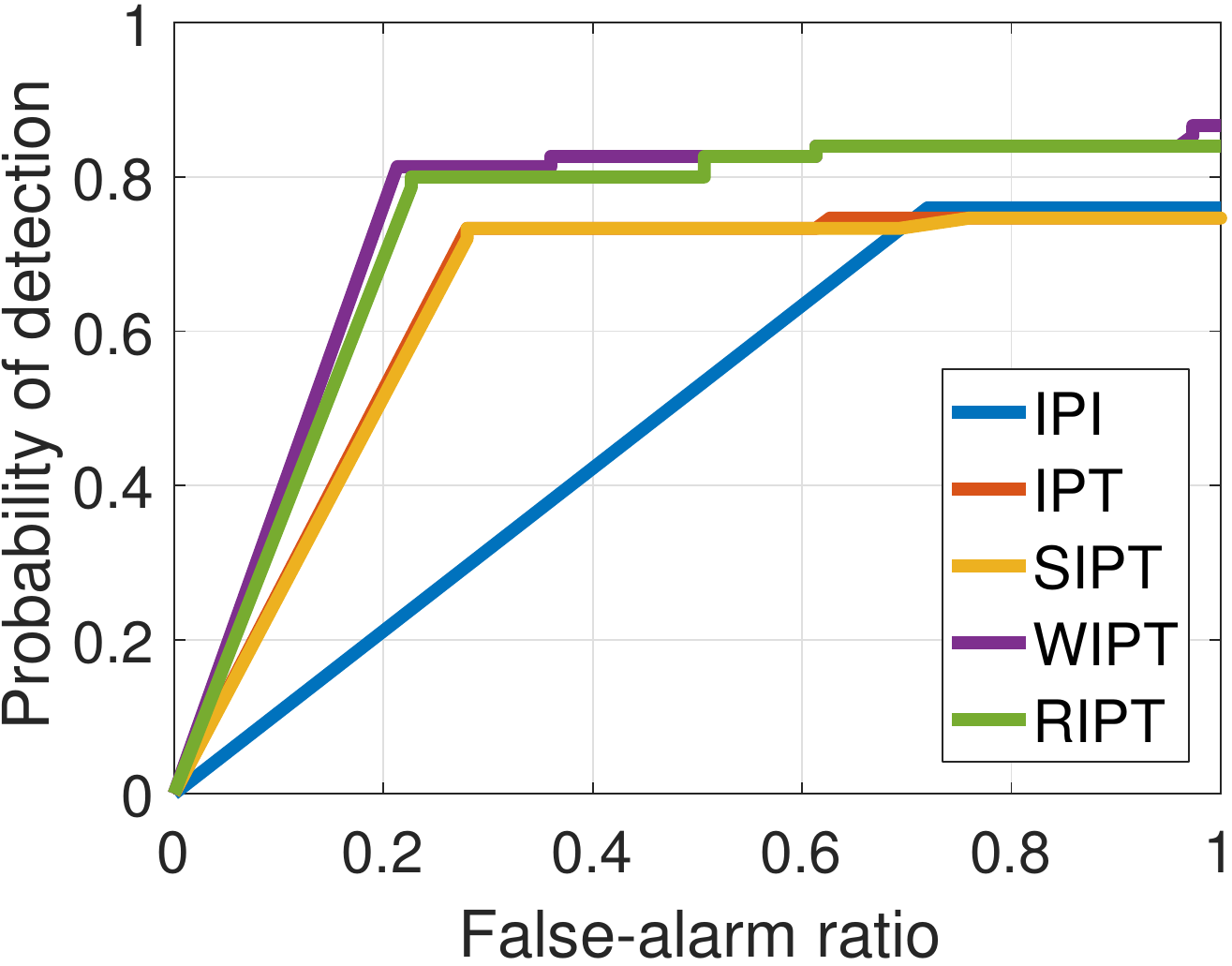}
	} 
	\subfloat[]{
		\includegraphics[width=0.23\textwidth]{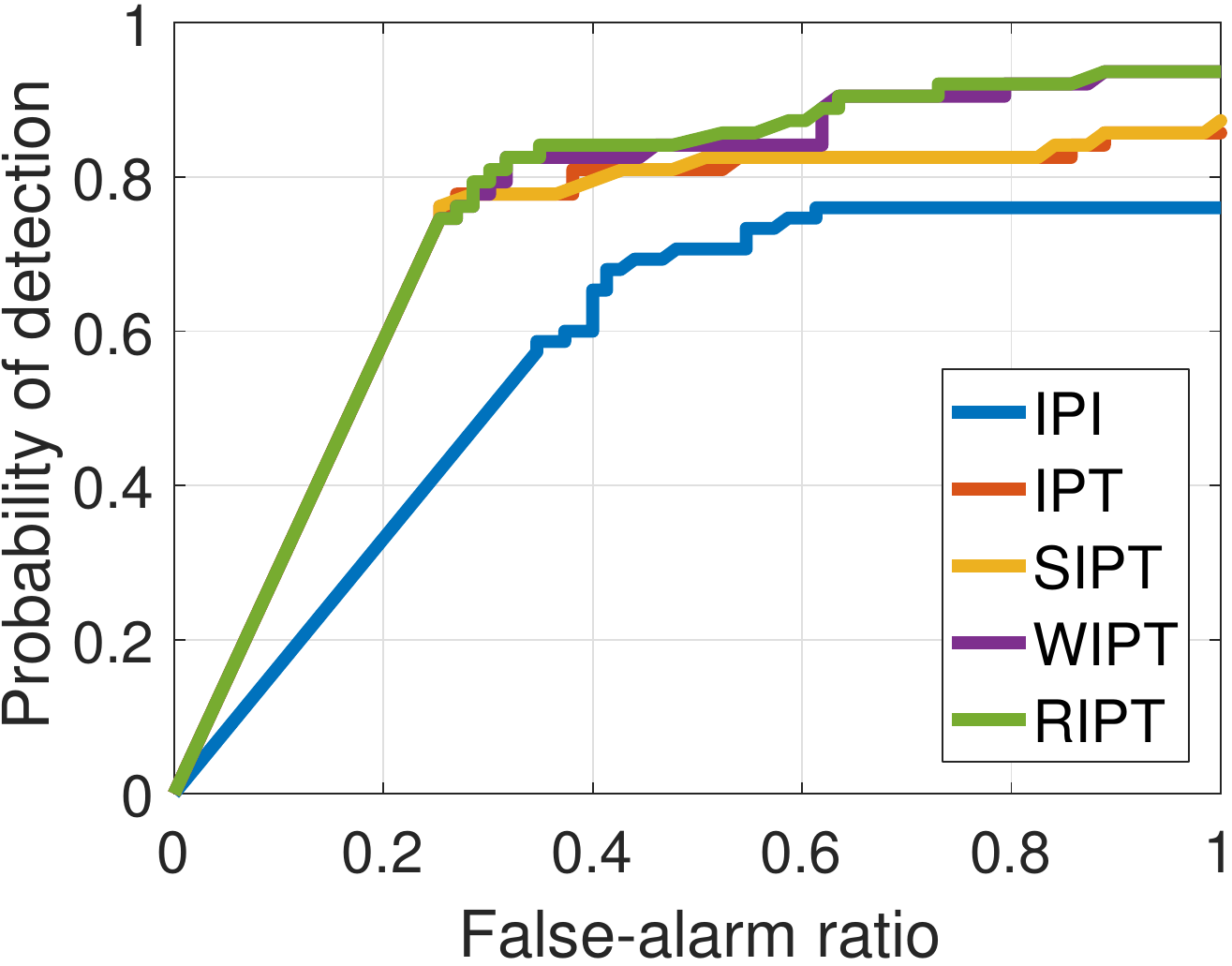}
	} 
	\caption{Illustration for the effects of the introduced priors. (a) -- (d) are the ROC curves on Sequence 1 -- 4.}
	\label{fig:Pior}
\end{figure*}

\begin{figure*}[htb!]
	\centering
	\subfloat[]{
		\includegraphics[width=0.23\textwidth]{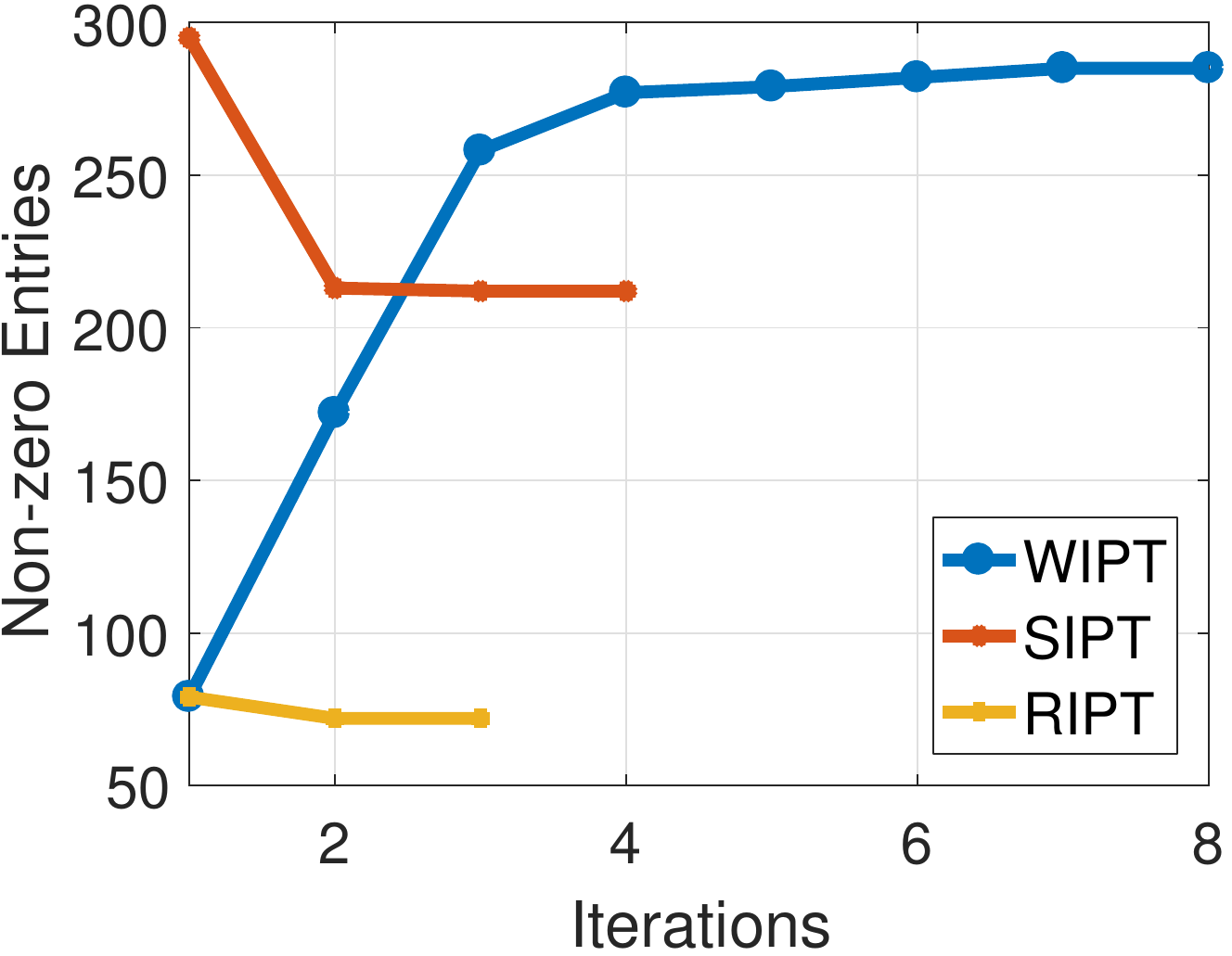}
	}
	\subfloat[]{
		\includegraphics[width=0.23\textwidth]{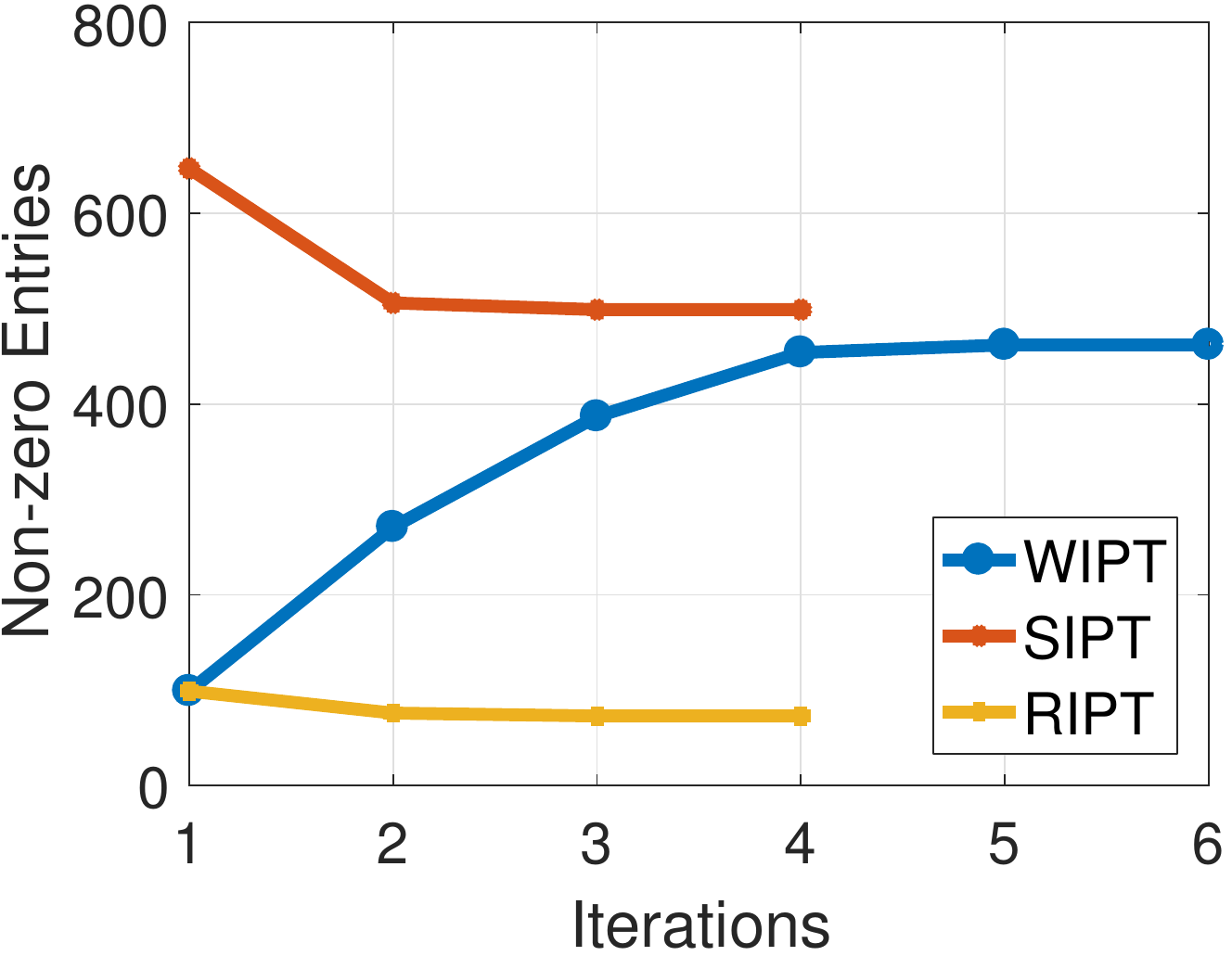}
	}                                           
	\subfloat[]{
		\includegraphics[width=0.23\textwidth]{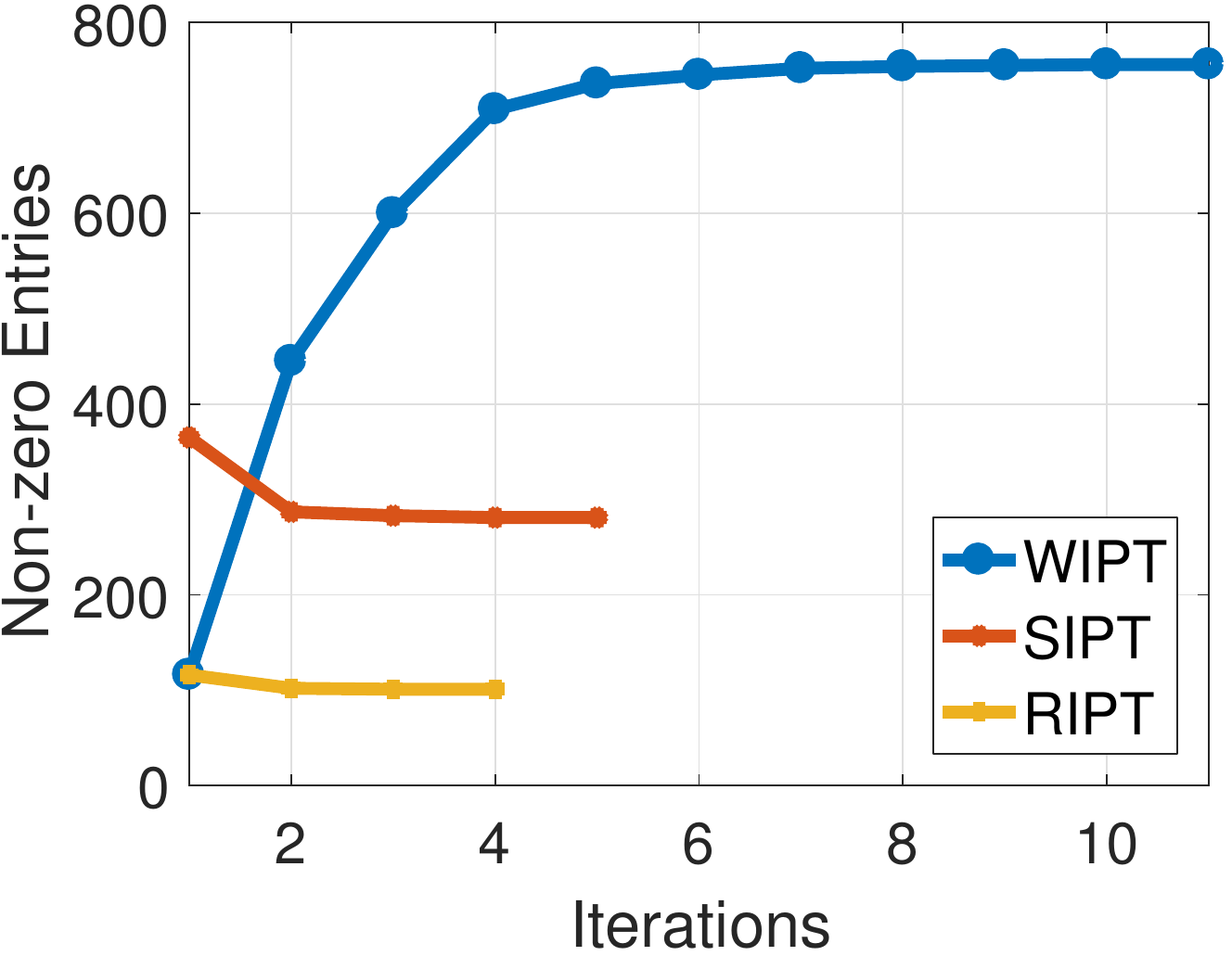}
	} 
	\subfloat[]{
		\includegraphics[width=0.23\textwidth]{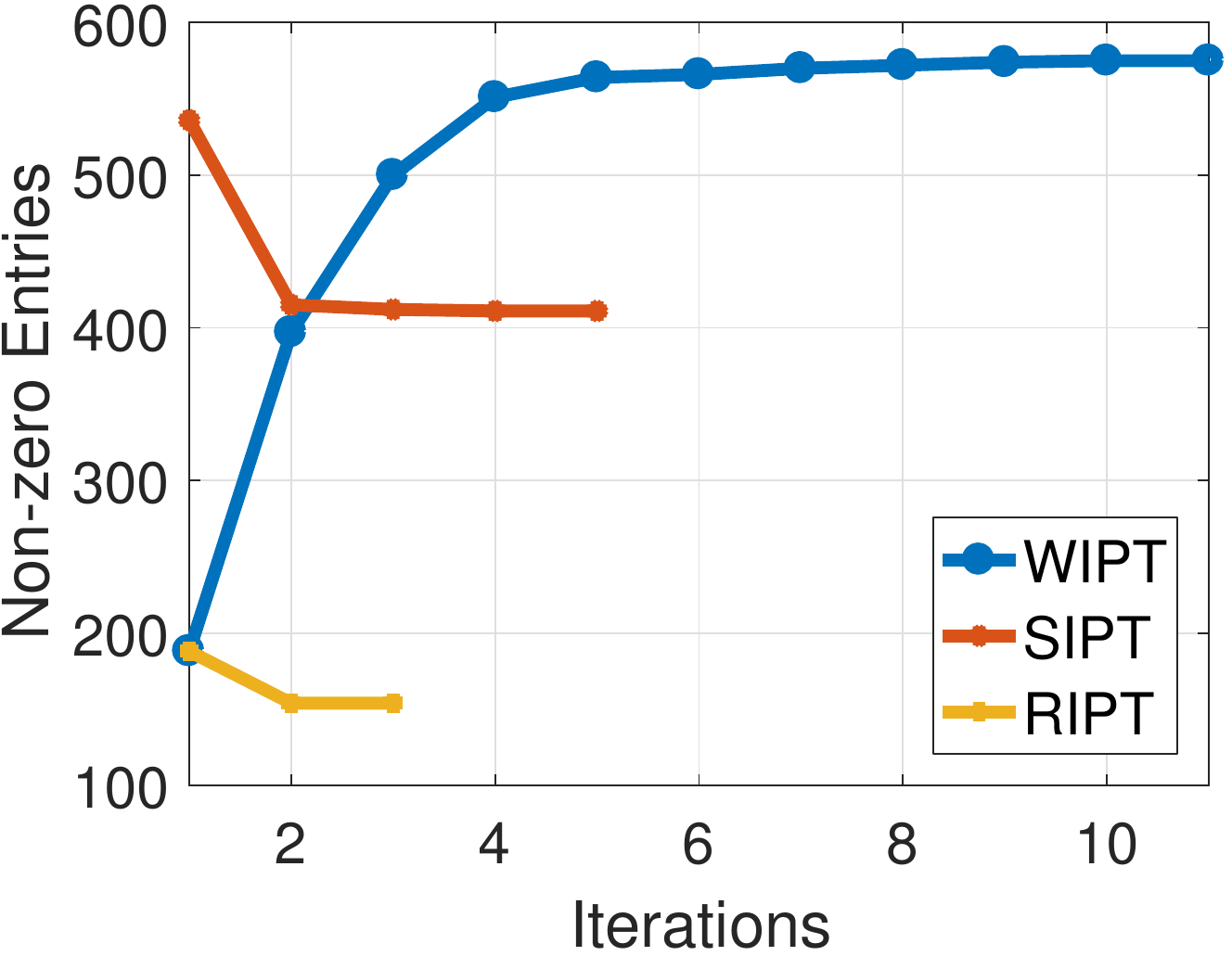}
	} 
	\caption{Illustration for the effect of the sparsity enhancement weight: (a) Sequence 1, (b) Sequence 2, (c) Sequence 3, and (d) Sequence 4.}
	\label{fig:Sparsity}
\end{figure*}

% 这一段的目的，就是下面这个
To further understand the effects of the components in the proposed RIPT model, we evaluate each prior individually with experiment to investigate how these priors influence the final detection performance.
The ROC curves of IPI, IPT, IPT with sparsity enhancement weight (SIPT), WIPT, and RIPT for Sequence 1 -- 4 are given in \cref{fig:Pior}, leading to the following observations.
(1) The four patch-tensor based methods outperform the state-of-the-art IPI method, which demonstrates that the patch-tensor model, involving mode-$1$ and mode-$2$ unfolding matrices, does contribute to the final detection performance. 
(2) By comparing WIPT and RIPT with IPT and SIPT, we see that incorporating local structure prior significantly improves the detection probability. 
(3) Although the sparsity enhancement weight does not contribute to the final detection performance, it significantly reduces the iteration number, as illustrated in \cref{fig:Sparsity}.
These observations indicate that the introduced priors are effective, and, when combined together, lead to excellent performance as reported in the next subsections.

\subsection{Algorithm Complexity and Computational Time }
The proposed model is solved via ADMM, which has been proved a $\mathcal{O}(1/k)$ convergence \cite{He2012SJoNA502,Wen2016IToCPP99}. 
Therefore, our solving algorithm is ensured to converge. 
The algorithm complexity and computational time for \cref{fig:Original}(a) with various methods are given in \cref{tab:Time}. 
The image size is $M\times N$, and $m,n$ are the rows and columns of the patch-image or mode-$3$ unfolding.
Although the computational complexities of these methods seem the same, their computing time differs a lot.
For the filtering and HVS based methods, the difference in computing time lies in whether the code could be vectorized.
For the low-rank methods, the dominant factor is the iteration number.
It can be seen from the data in \cref{tab:Time} that the low-rank methods are generally slower than the filtering and HVS based methods. 
Nevertheless, considering low-rank method could handle more difficult scenes, this trade-off is acceptable. 
Among the low-rank methods, the RIPT costs the least time. 
The underlying reason is that both the local structure weight and sparsity enhancement weight help to reduce the iteration number. 
In addition, unlike the weight in WIPI, the time for constructing the weight is neglectable in RIPT.
% 从他们的计算复杂度上看不出太大的差别。对于 HVS，主要在于是否能够向量化，而对于 低秩，主要在于 SVD 的迭代次数。
% 本文方法，是低秩方法中耗时最少的。这主要来源于两方面，一个是 strucutre tensor 计算快，而 WIPI 的计算太慢。另外一个是 sparsity enhancement         
\begin{table*}[htbp]
	\centering
	\setlength{\tabcolsep}{3pt}
	\caption{Algorithm Complexity and Computational Time Comparisons of Different Methods}
	\begin{tabular}{ccccccccccc}
		\toprule
		           & TDLMS                & PFT                       & MPCM                 & WLDM                 & IPI                 & WIPI                & NIPPS               & IPT                 & WIPT                & RIPT                \\
		\midrule
		Complexity & $\mathcal{O}(L^2MN)$ & $\mathcal{O}(MN\log{MN})$ & $\mathcal{O}(L^3MN)$ & $\mathcal{O}(L^3MN)$ & $\mathcal{O}(mn^2)$ & $\mathcal{O}(mn^2)$ & $\mathcal{O}(mn^2)$ & $\mathcal{O}(mn^2)$ & $\mathcal{O}(mn^2)$ & $\mathcal{O}(mn^2)$ \\ [0.5ex]		
		Time (s)   & 0.162                & 0.025                     & 0.083                & 6.059                & 16.998              & 52.995              & 15.515              & 8.598               & 6.932               & 3.169               \\
		\bottomrule
	\end{tabular}%
	\label{tab:Time}%
\end{table*}%

\subsection{Parameters analysis}
For the proposed model, the related parameters, such as the patch size, sliding step, weight stretching parameter $h$, weighting parameter $\lambda$, and penalty factor $\mu$, are all important factors, which usually affects the model fitness on the real databases. 
Therefore, a better performance can be obtained by finely tuning these parameters. 
Nevertheless, the optimal values are always related to the infrared image content.
In \cref{fig:Parameter_Analysis}, we give the ROC curves for different model parameters on Sequence 1 -- 4 to evaluate their influence. 
These parameters are validated to obtain a local optimal value with other parameters fixed. 
The stepped shape of our ROC curves might seem a bit odd. It is because we have adopted a much larger weighting parameter $\lambda$ than normal RPCA-based foreground-background separation tasks in order to better fit the actual situation of single-frame infrared small target detection.

\subsubsection{Patch size $I\times J$}
\begin{figure*}[h!tb]
	\centering
	\subfloat{
		\includegraphics[width=0.215\textwidth]{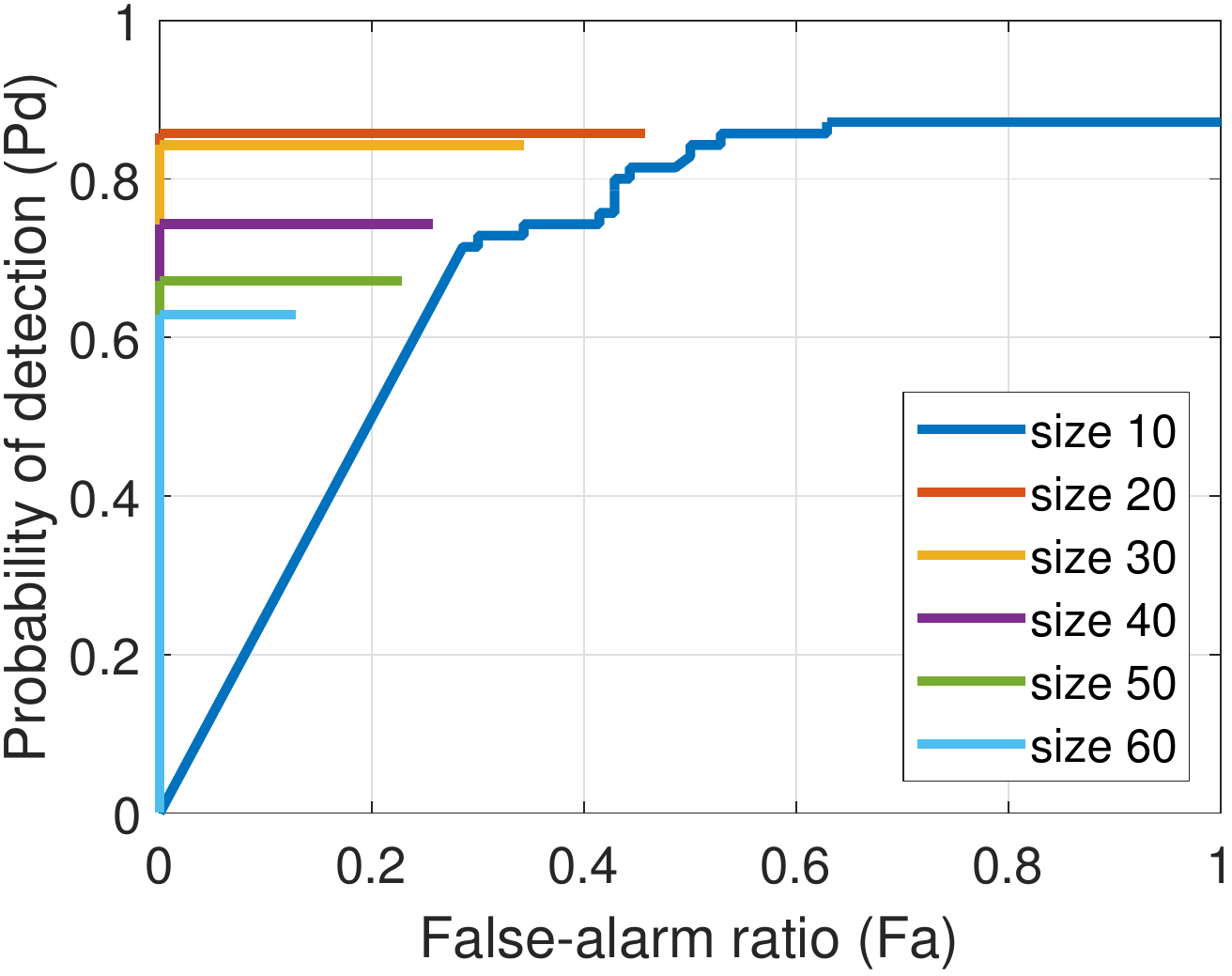}
	}
	\subfloat{
		\includegraphics[width=0.215\textwidth]{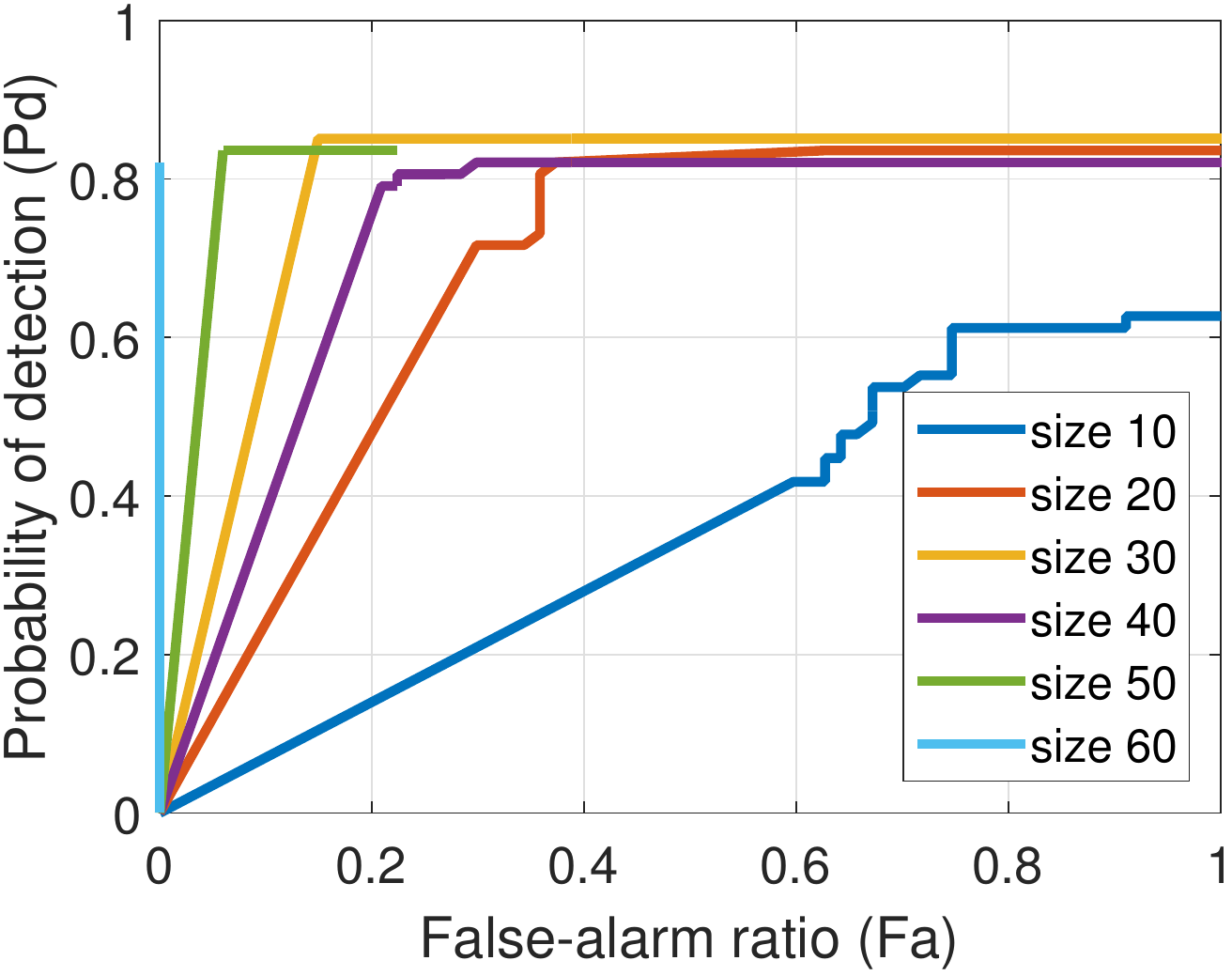}
	}
	\subfloat{
		\includegraphics[width=0.215\textwidth]{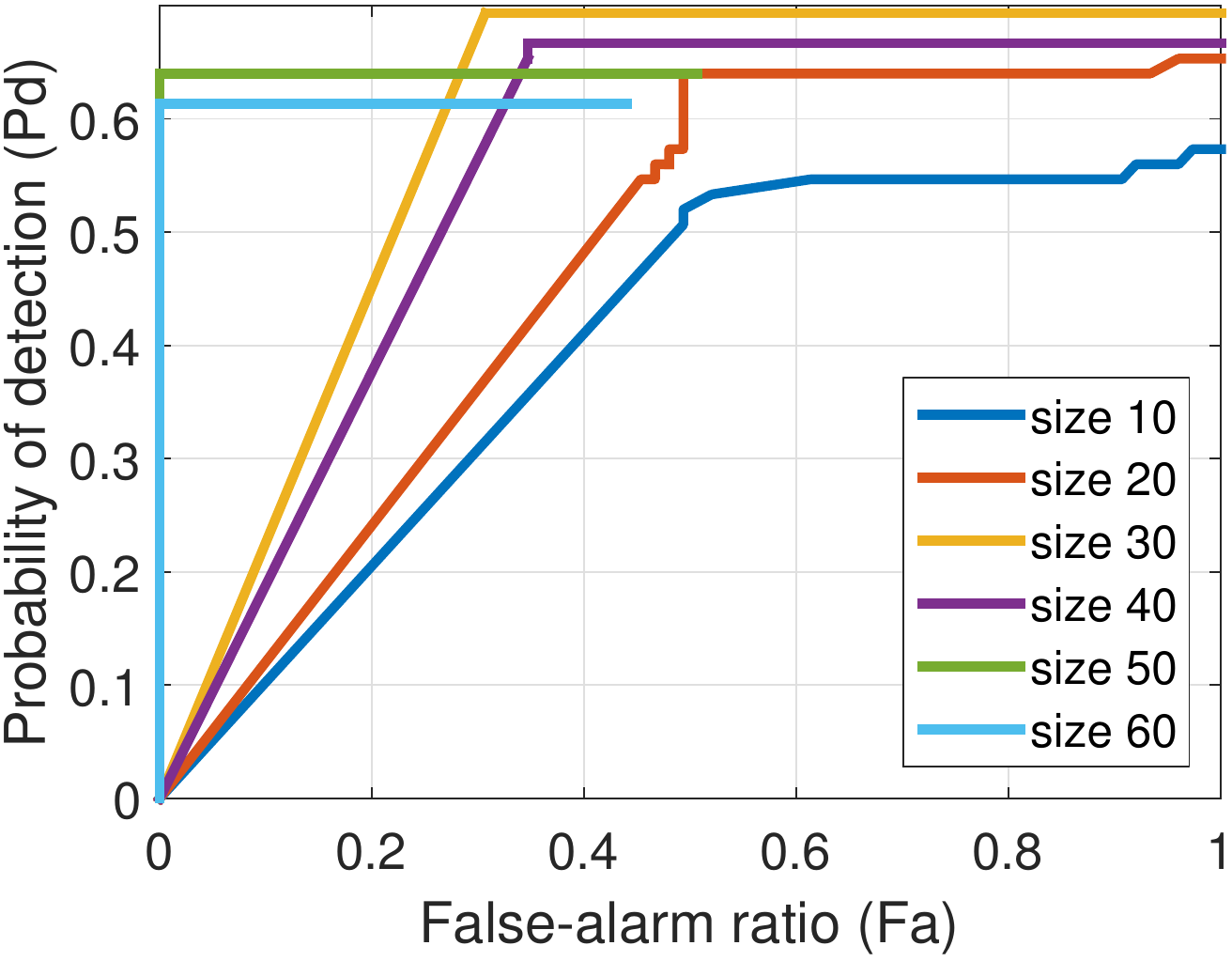}
	} 
	\subfloat{
		\includegraphics[width=0.215\textwidth]{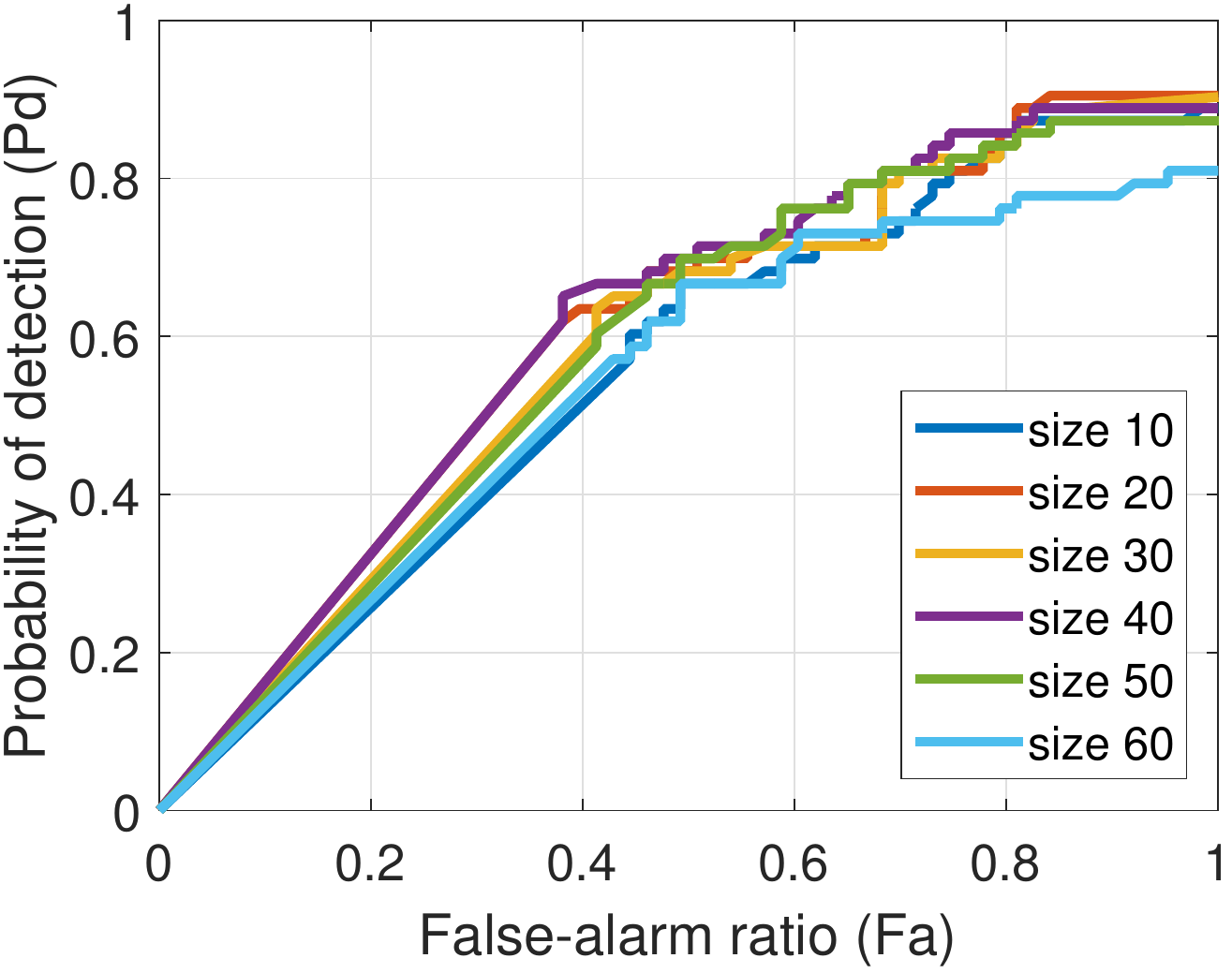}
	}   
																																															                                            
	\subfloat{
		\includegraphics[width=0.215\textwidth]{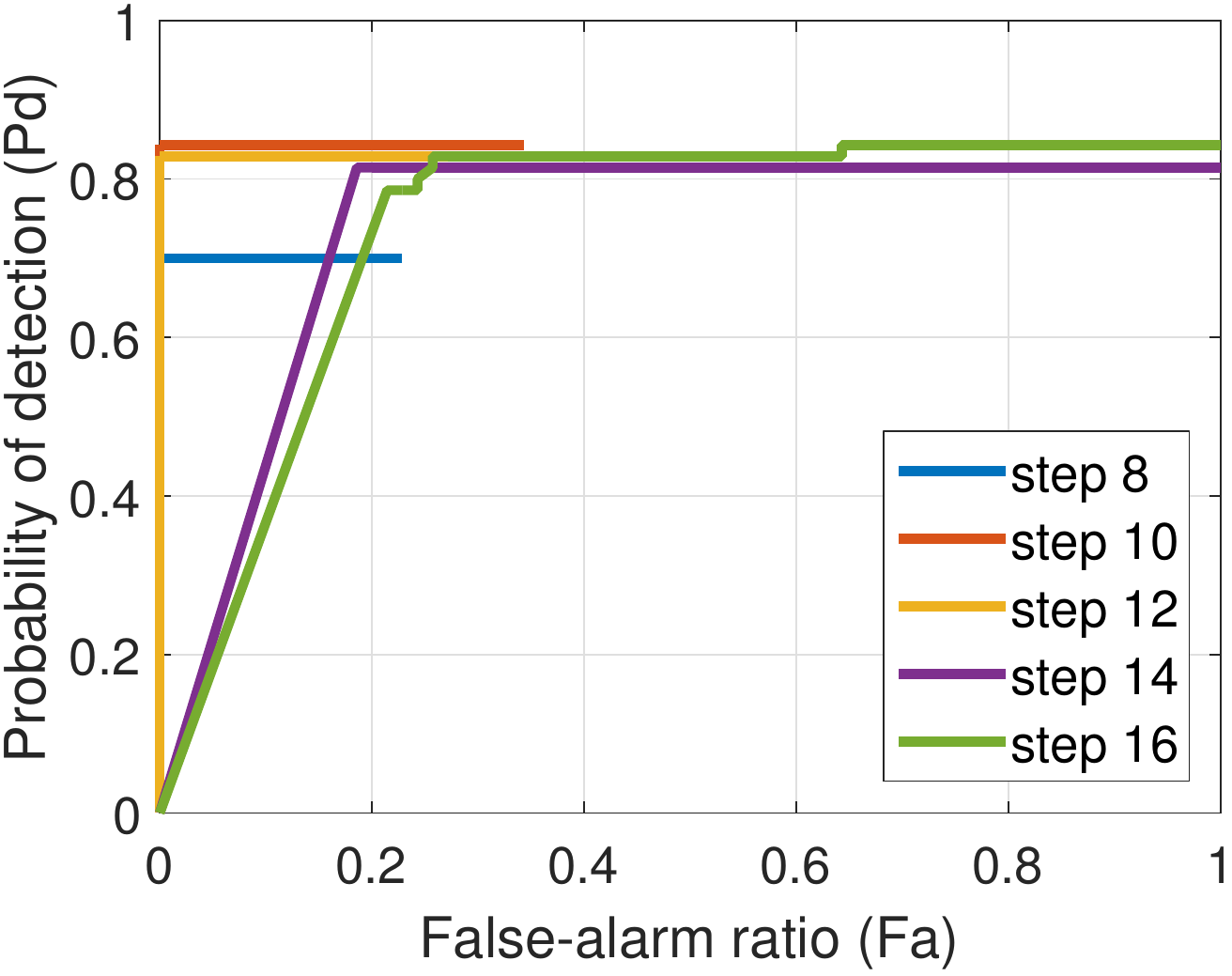}
	}
	\subfloat{
		\includegraphics[width=0.215\textwidth]{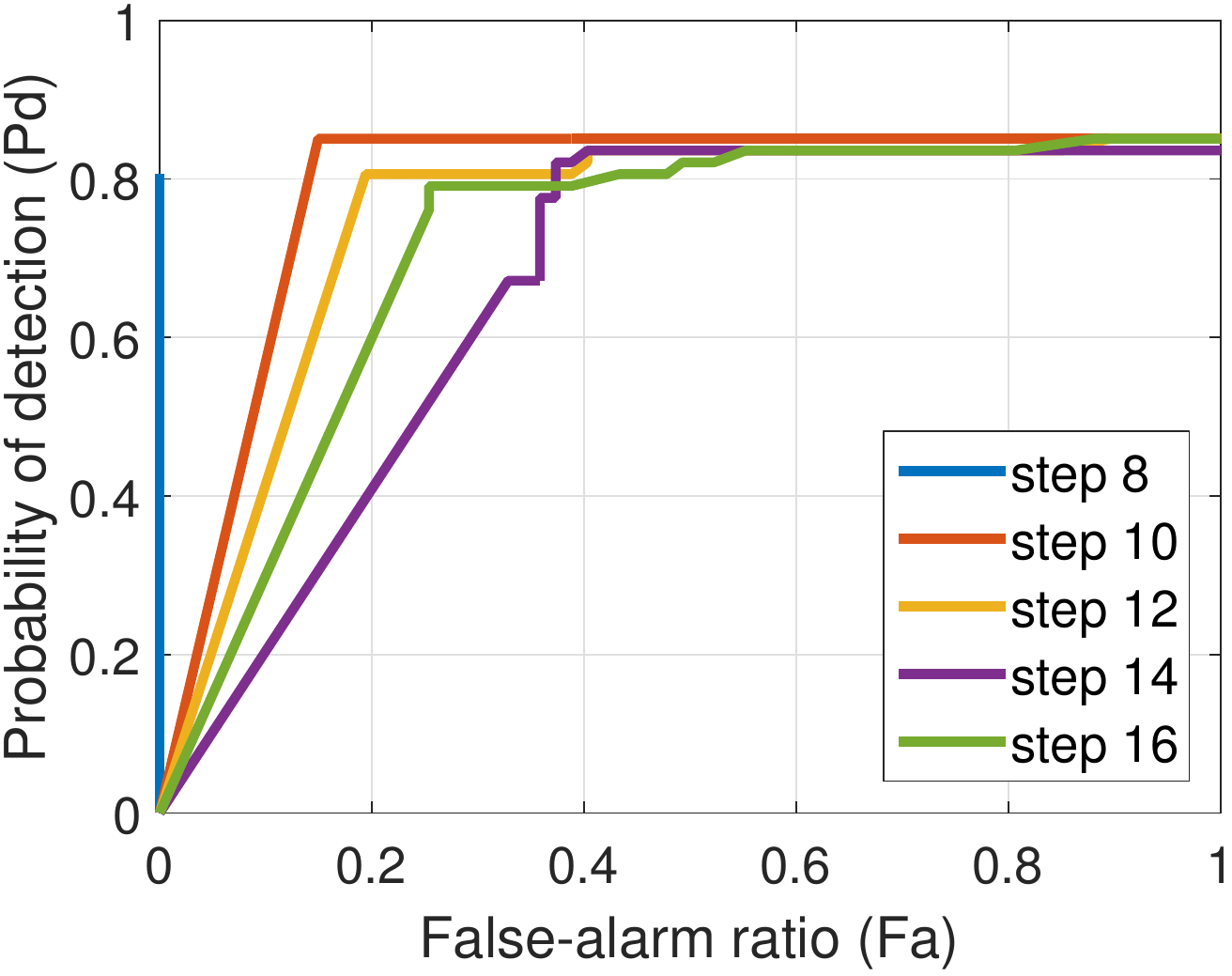}
	}
	\subfloat{
		\includegraphics[width=0.215\textwidth]{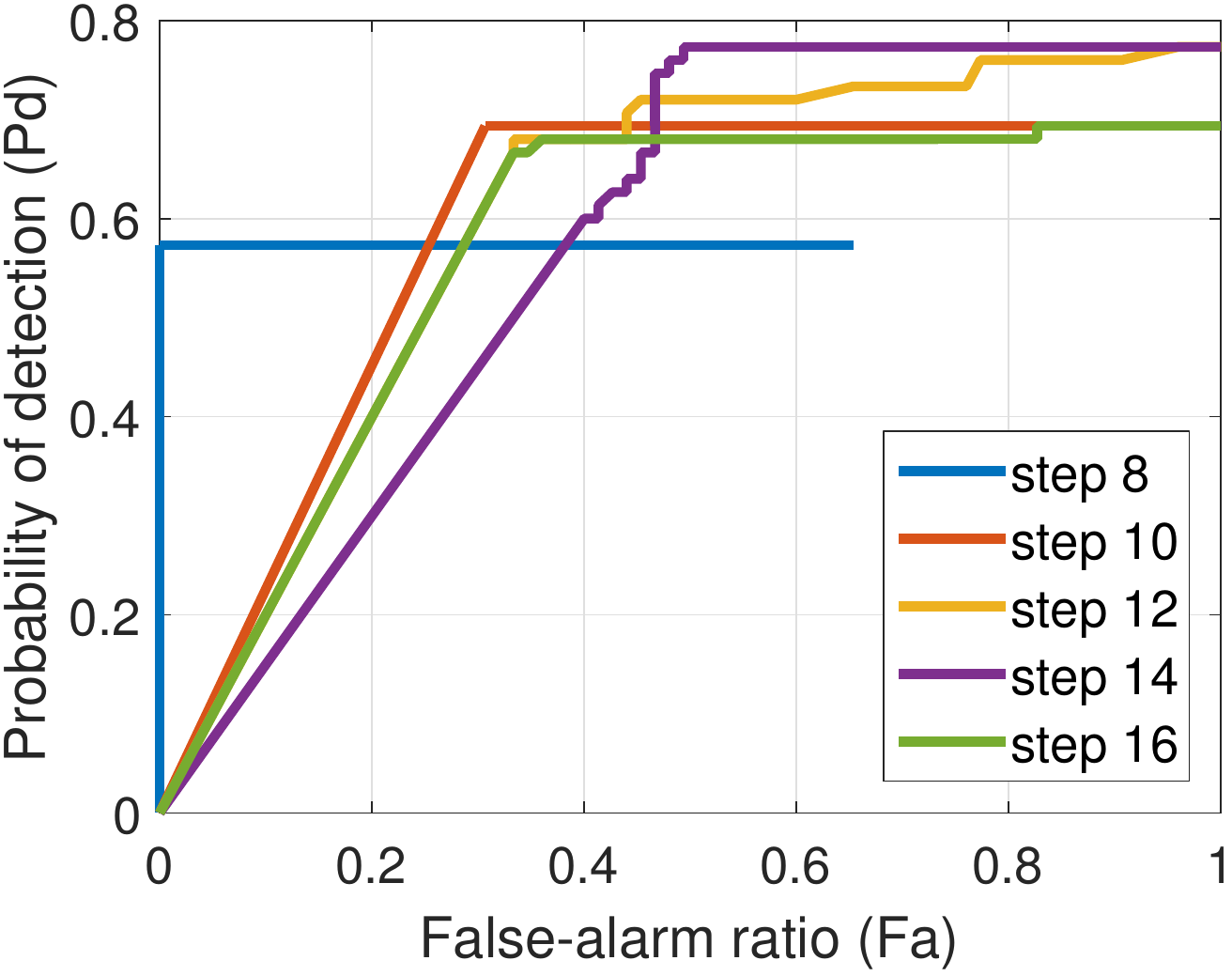}
	} 
	\subfloat{
		\includegraphics[width=0.215\textwidth]{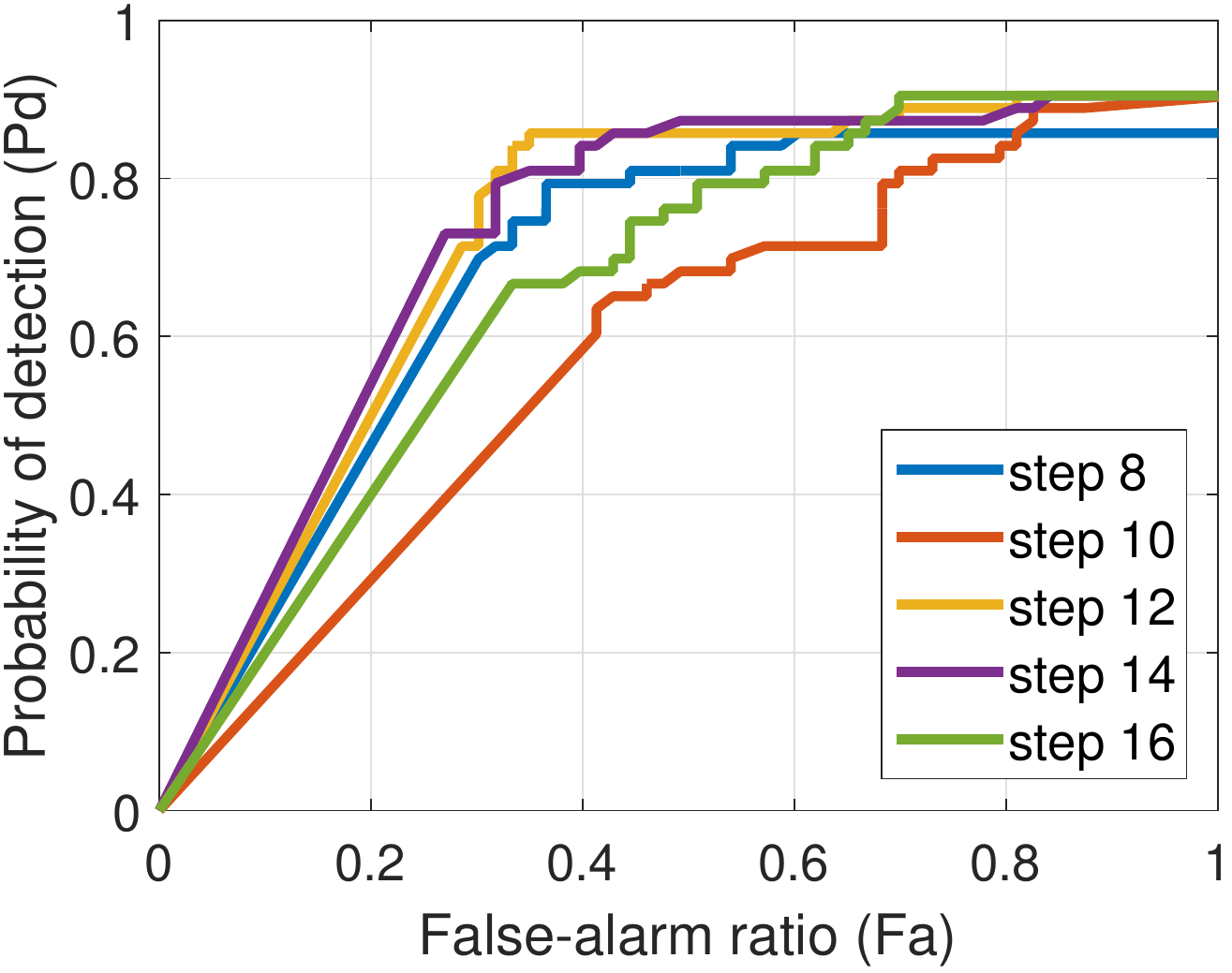}
	}
																																															                                            
	\subfloat{
		\includegraphics[width=0.215\textwidth]{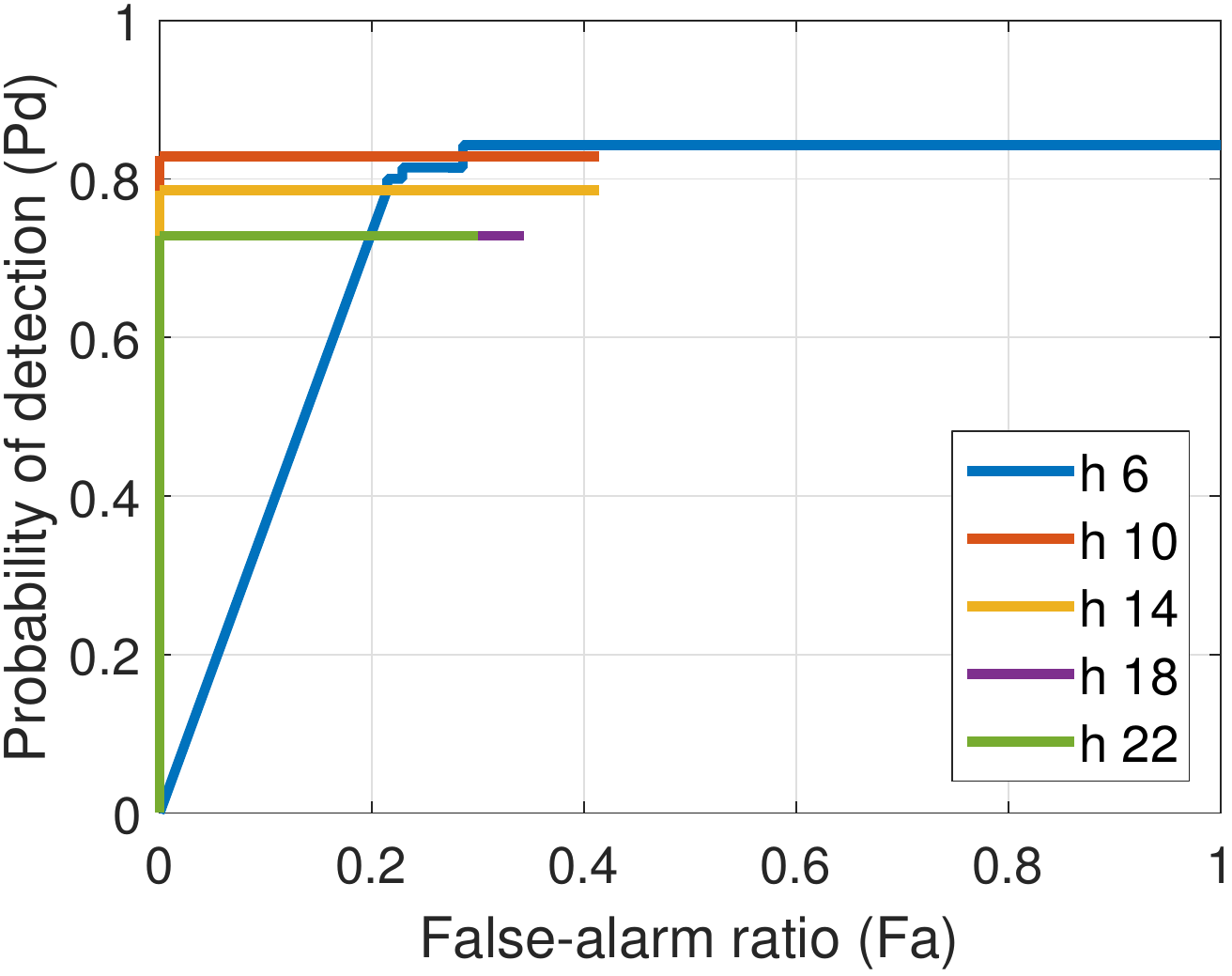}
	}
	\subfloat{
		\includegraphics[width=0.215\textwidth]{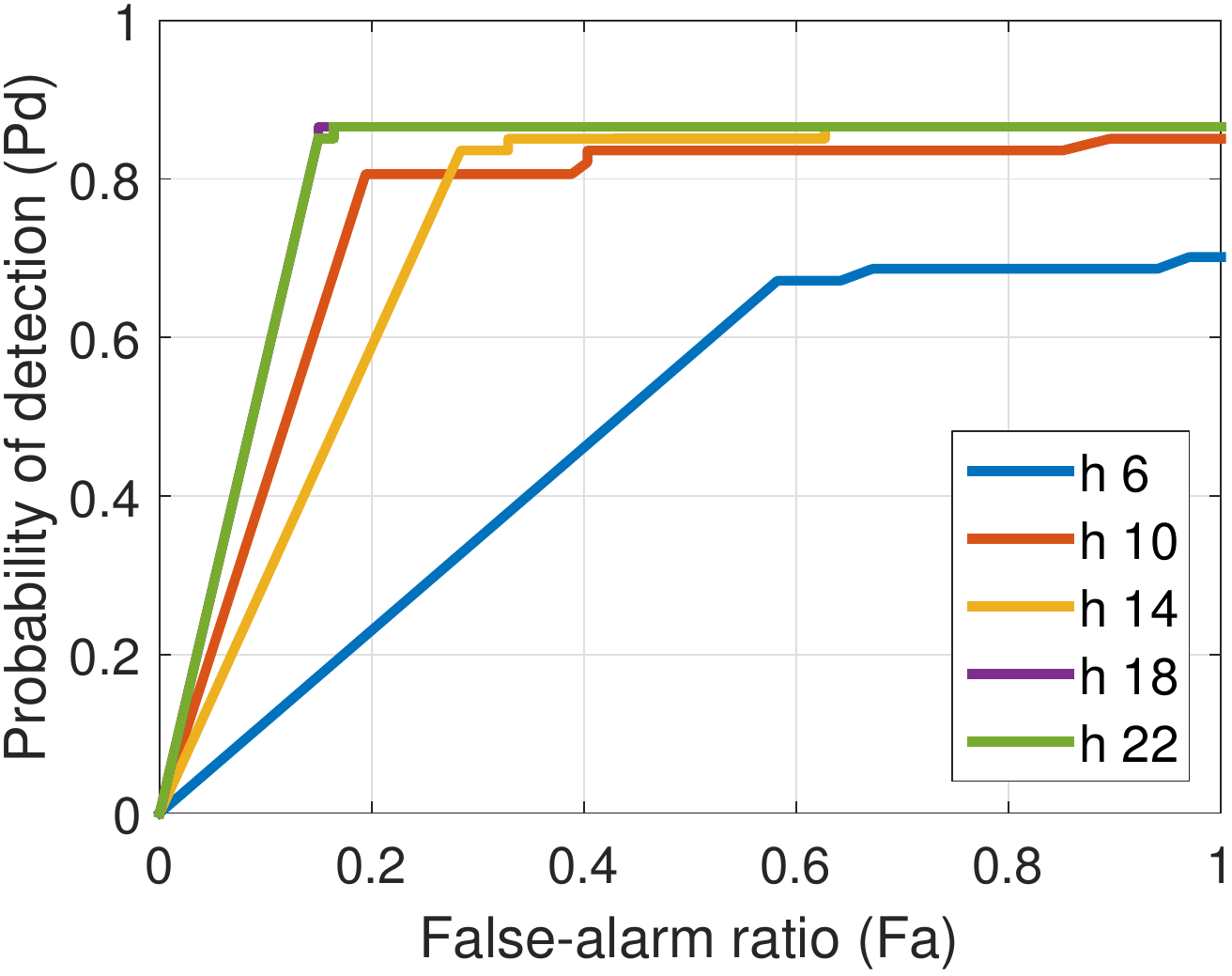}
	}
	\subfloat{
		\includegraphics[width=0.215\textwidth]{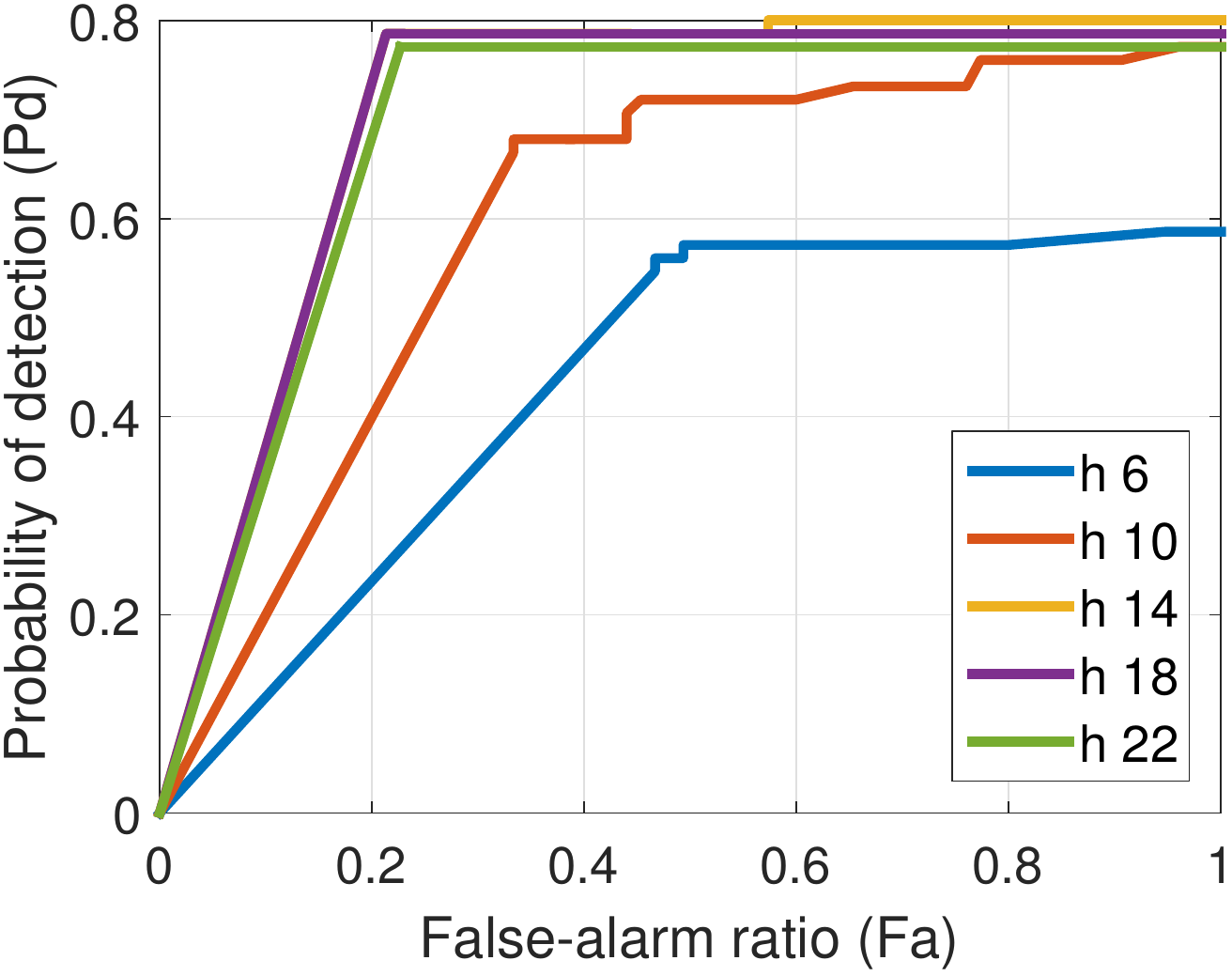}
	} 
	\subfloat{
		\includegraphics[width=0.215\textwidth]{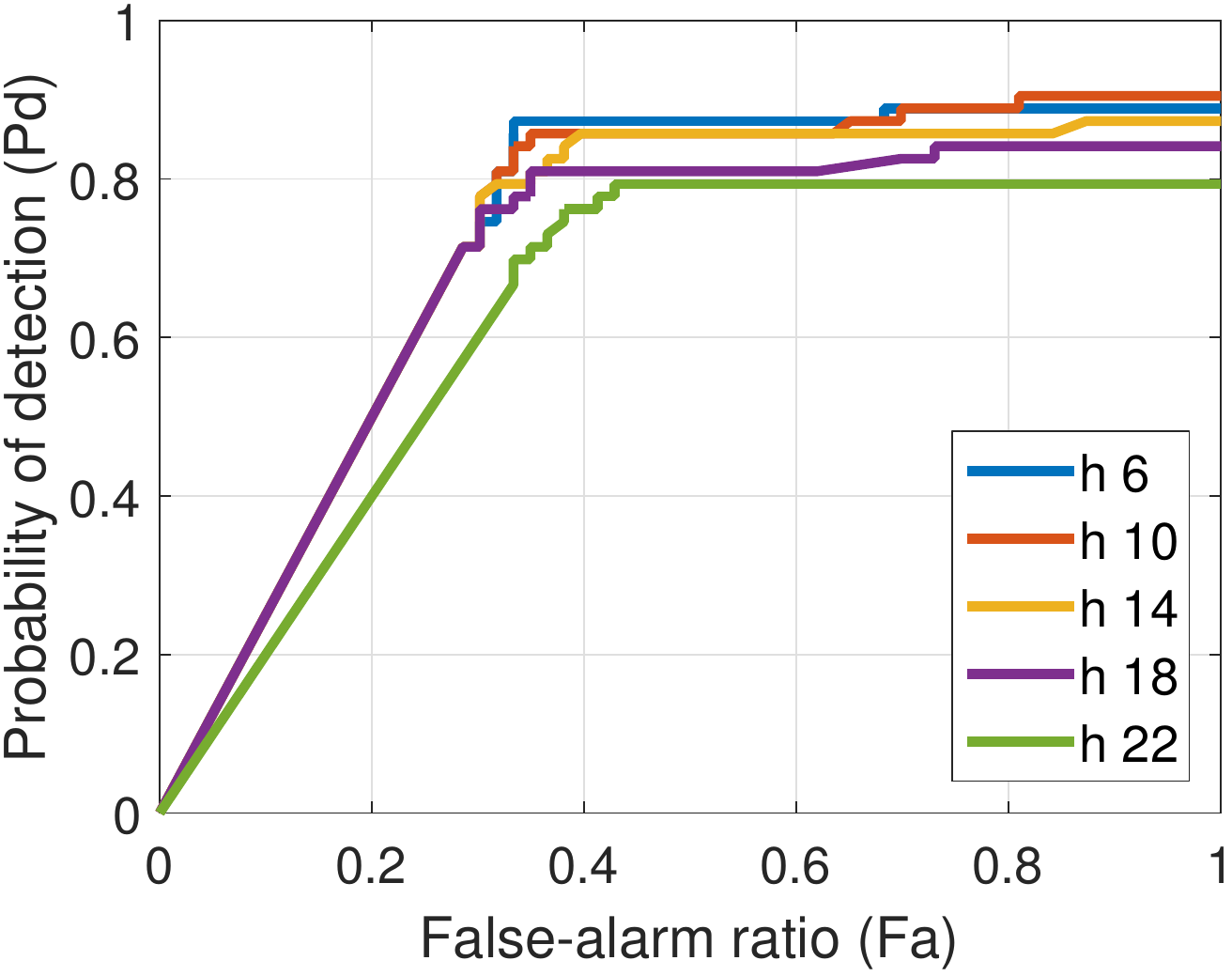}
	}     
																																															                                            
	\subfloat{
		\includegraphics[width=0.215\textwidth]{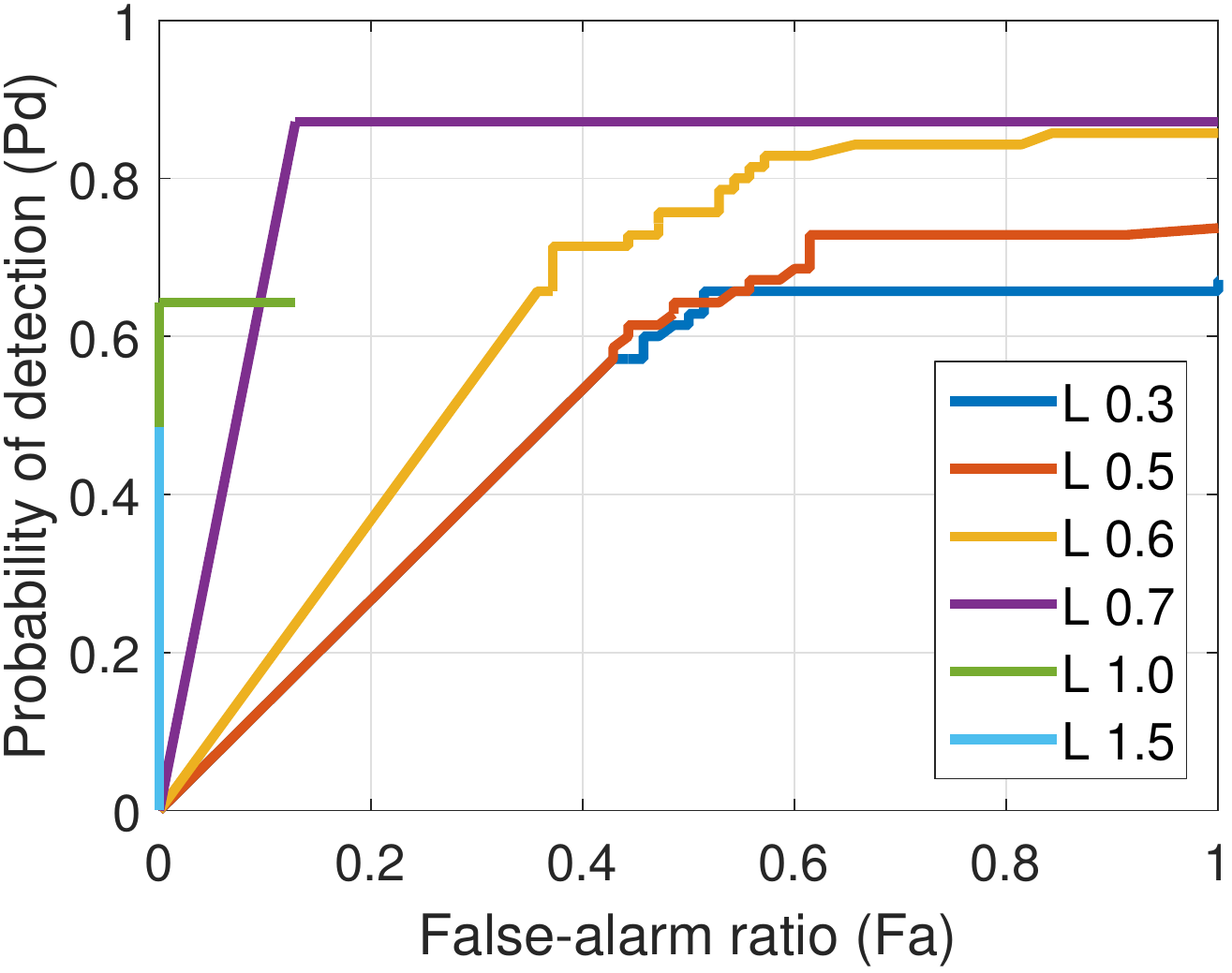}
	}
	\subfloat{
		\includegraphics[width=0.215\textwidth]{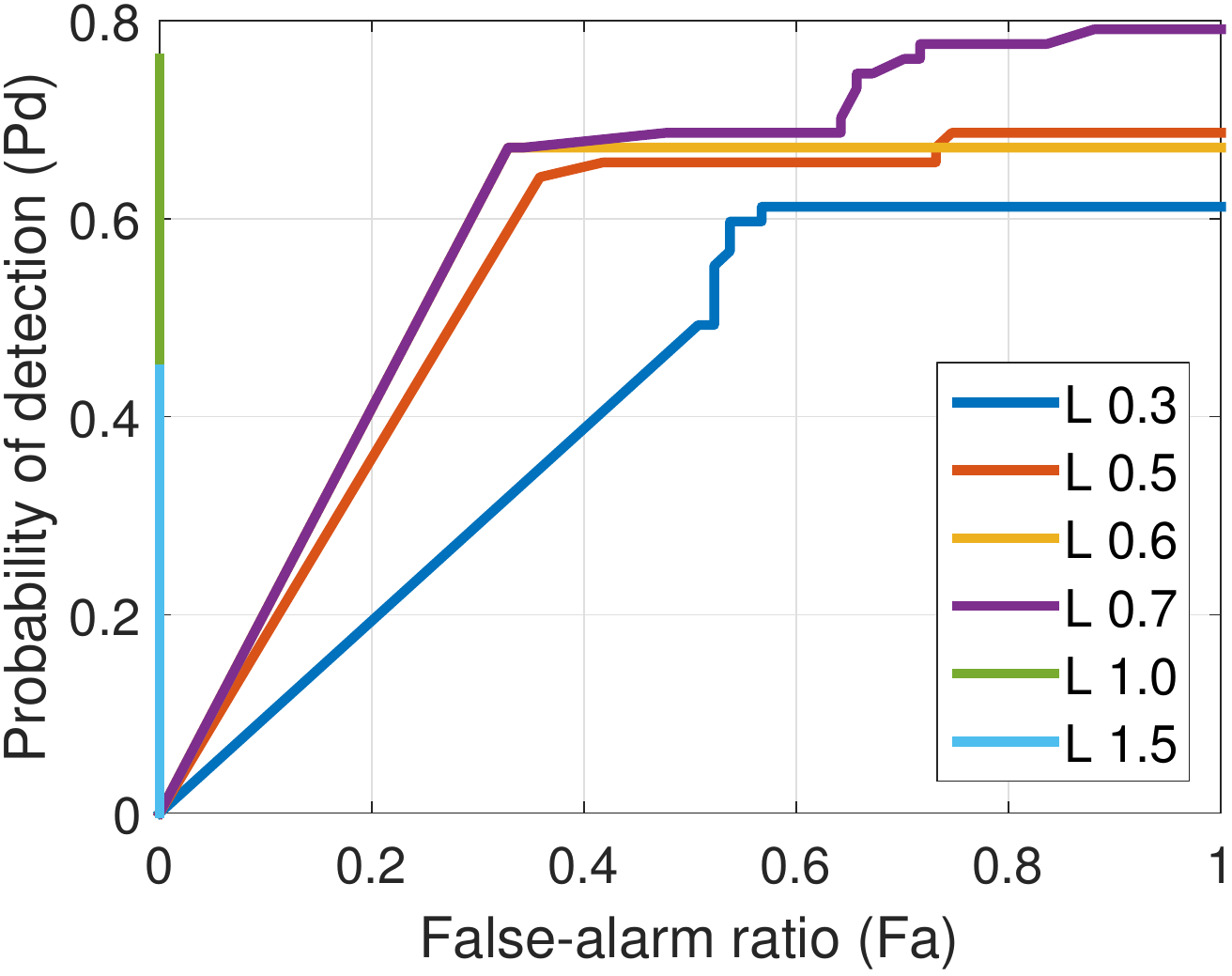}
	}
	\subfloat{
		\includegraphics[width=0.215\textwidth]{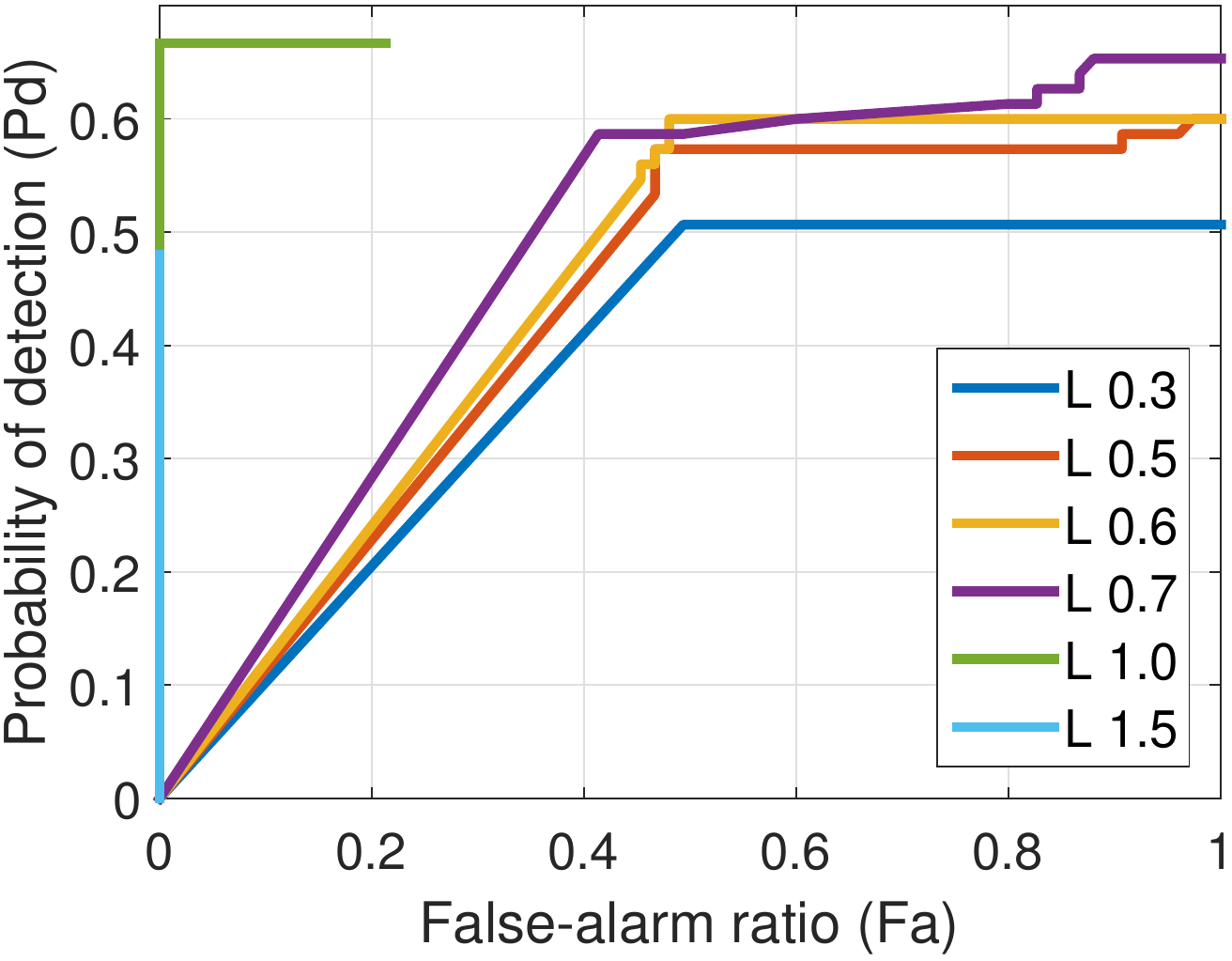}
	} 
	\subfloat{
		\includegraphics[width=0.215\textwidth]{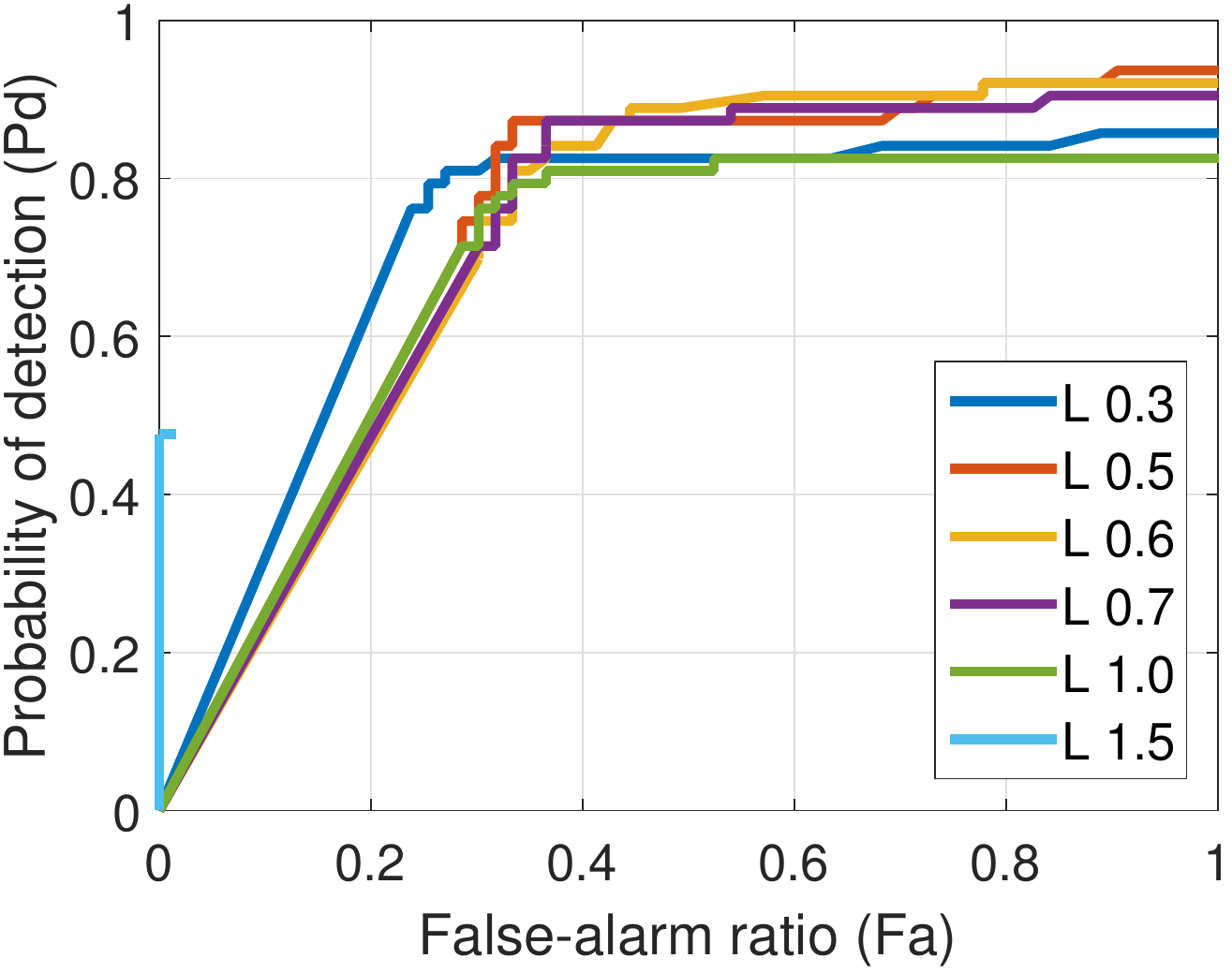}
	} 
																																															                                            
	\subfloat{
		\includegraphics[width=0.215\textwidth]{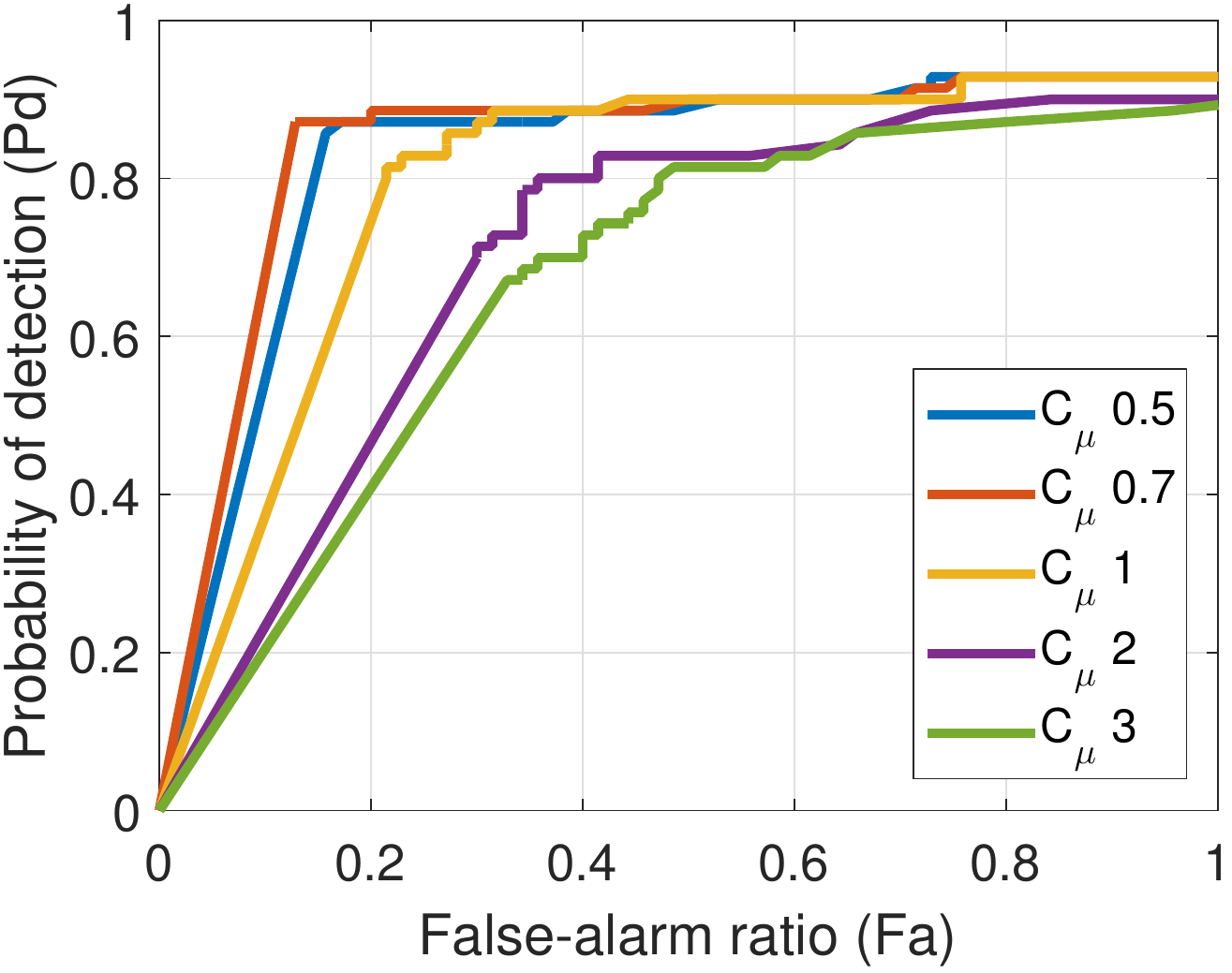}
	}
	\subfloat{
		\includegraphics[width=0.215\textwidth]{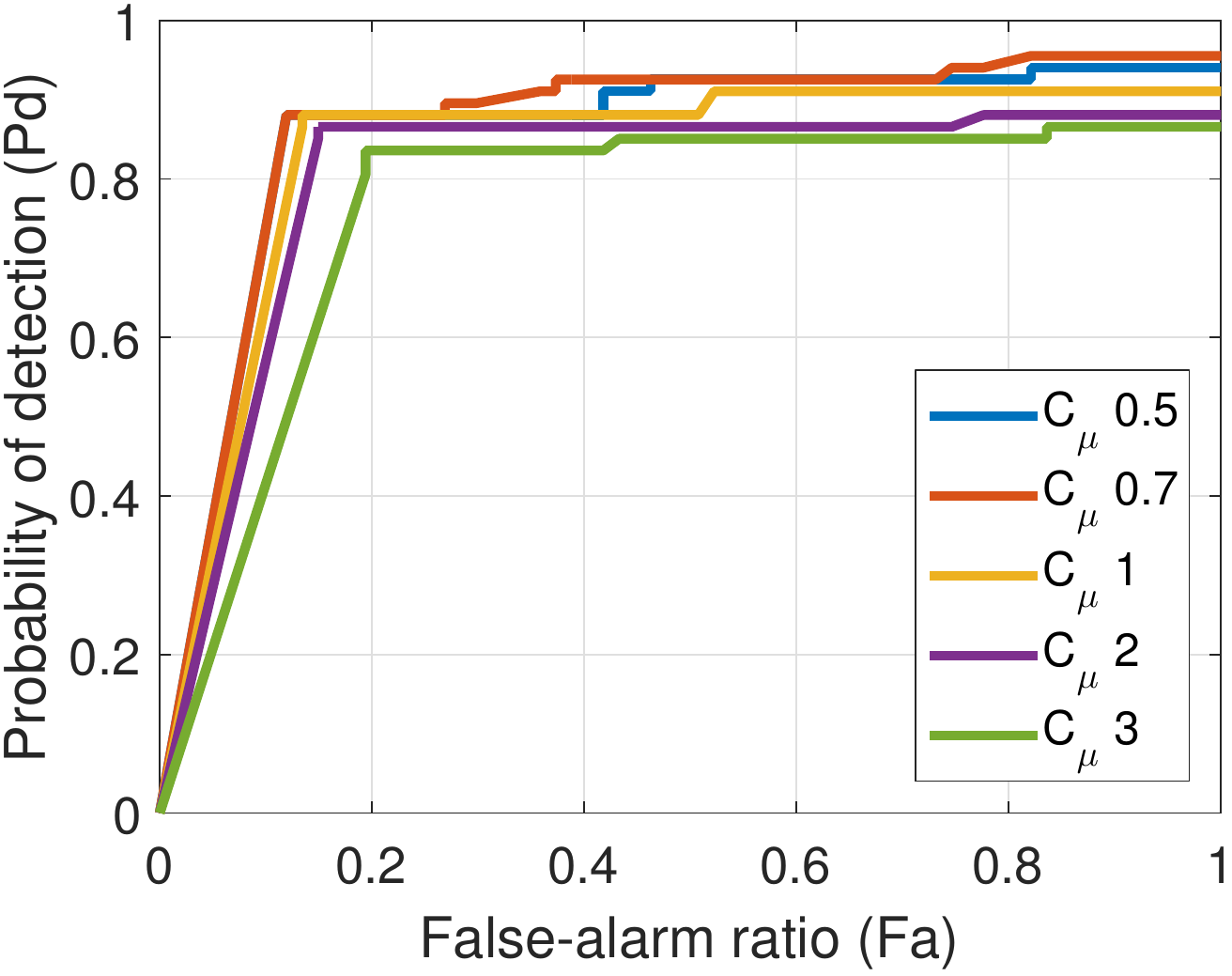}
	}
	\subfloat{
		\includegraphics[width=0.215\textwidth]{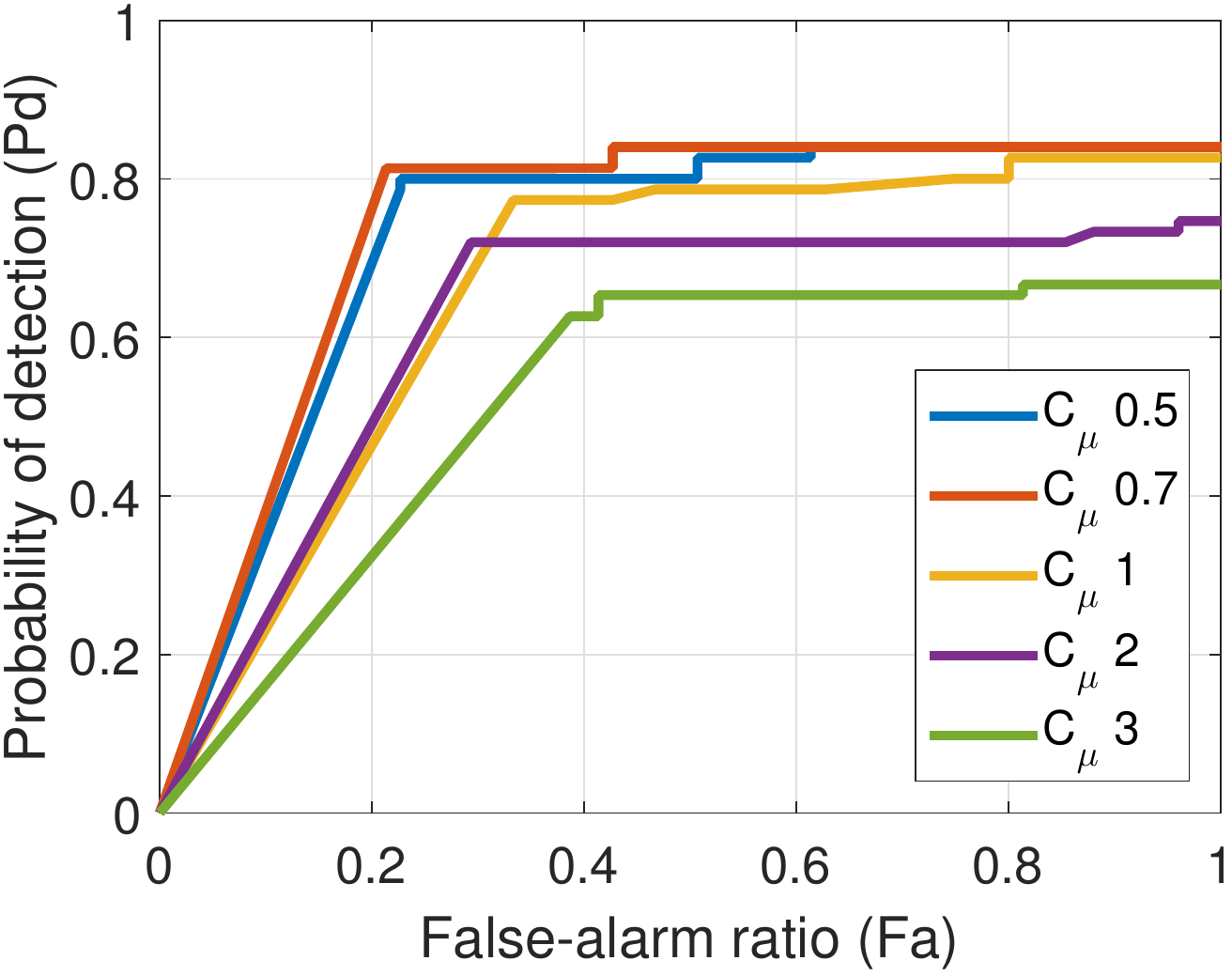}
	} 
	\subfloat{
		\includegraphics[width=0.215\textwidth]{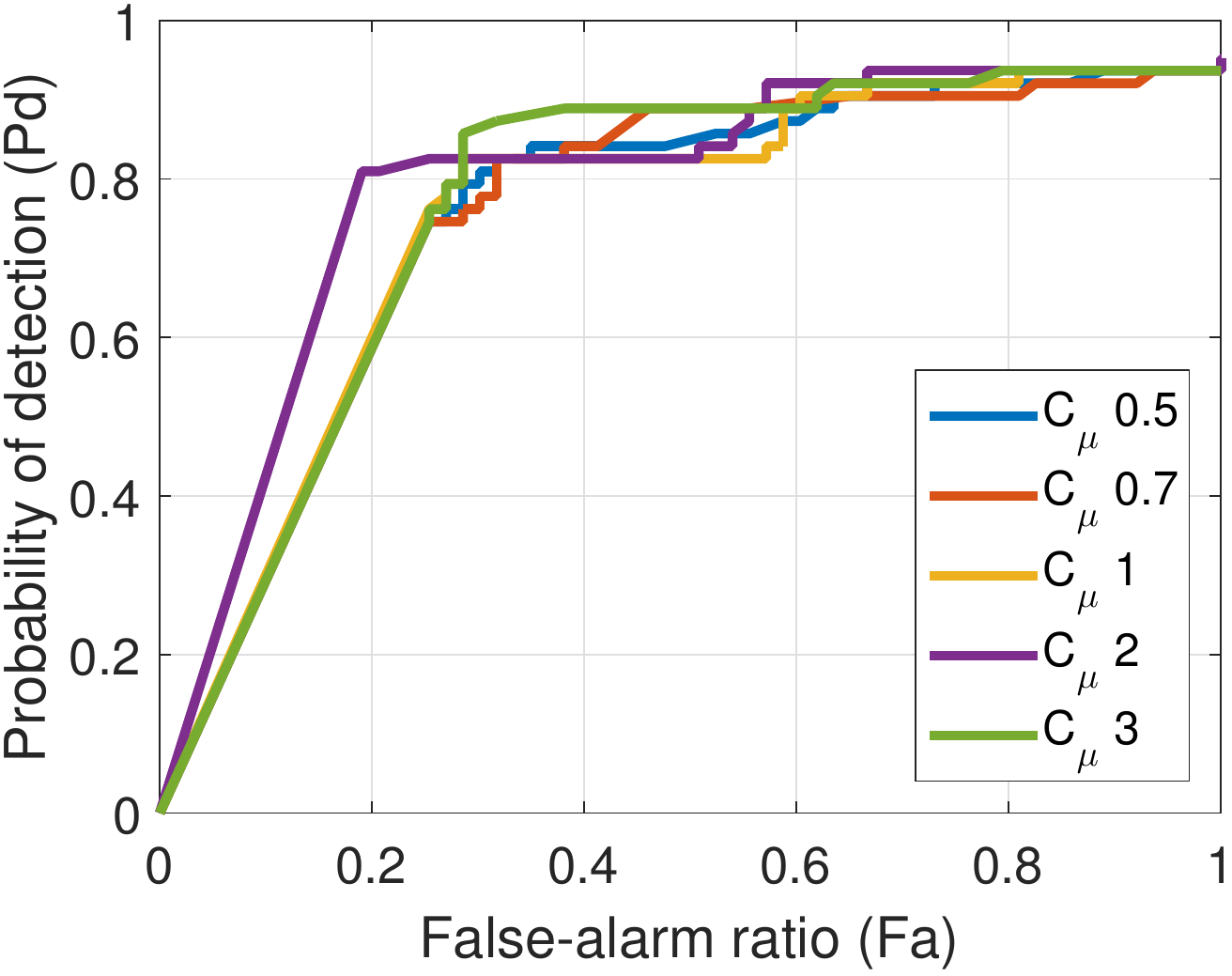}
	} 
	\caption{The ROC curves for Sequence 1 -- 4 with respect to different patch sizes, sliding steps, $h$, weighting parameters and penalty factors. Row 1: Different patch sizes, Row 2: Different sliding steps, Row 3: Different $h$, Row 4: Different weighting parameters, Row 5: Different penalty factors.}
	\label{fig:Parameter_Analysis}
\end{figure*}

It not only has a large impact on the separation, but also influences the computational complexity. 
The matrix size of mode-$1$ and mode-$2$ unfoldings of the patch-tensor is $I \times (J\cdot P)$; the matrix size of mode-$3$ unfolding is $J \times (I\cdot P)$. 
Obviously, a smaller patch size will lead to a smaller computational complexity. 
On one hand, we hope a larger patch size to ensure that the target is sparse enough. 
On the other hand, a larger patch size reduces the correlationships between the non-local patches, which degrades the separation results. 
To seek a balance between a low computational burden, target sparsity, and background correlationship, we vary the patch size $P$ from $10$ to $60$ with ten intervals and provide the ROC curves in the first row of \cref{fig:Parameter_Analysis}. 
By observing the ROC curves, we can have the following conclusions. 
Firstly, a smaller patch size is easier to raise false alarms while a larger patch size may lead to a relatively lower detection probability, which just demonstrates our above analysis about the patch size. 
Secondly, the proposed RIPT method is not very sensitive to the patch size. 
The detection result of the patch size among $20$ -- $60$ is acceptable.
Thirdly, $30$ seems a good choice for Sequence 1 -- 4 since it achieves the best performance in ROC.

\subsubsection{Sliding step}
The sliding step influences the patch-tensor size as well. 
To reduce the computational complexity, we prefer a larger sliding step, which means smaller matrices to perform SVD. 
Nevertheless, a larger sliding step also reduces the redundancy of the original patch-tensor $\bm{\mathcal{T}}$ and undermines the final detection results since our proposed model is based on the non-local repentance of correlated patches. 
To investigate its influence, we vary the sliding step $S$ from $8$ to $16$ with two intervals. 
The results are displayed in the second row of \cref{fig:Parameter_Analysis}. 
It could be observed that the ROC curve of small sliding step like $8$ tends to have a more sharp shape, but its overall detection probability remains relatively low. 
The best value for sliding step is among $12$ to $14$, here we pick $12$. 
In addition, by comparing the first row with the second row of \cref{fig:Parameter_Analysis}, we can conclude that the algorithm is quite robust to the variation of step length.

\subsubsection{Weight stretching parameter $h$}
It controls the local structure weight's suppression degree to the clutter edges. 
We vary $h$ from $6$ to $22$ in the experiment and illustrate the ROC curves in the third row of \cref{fig:Parameter_Analysis}. 
Generally, we would like a larger $h$ which suppresses the undesirable non-target components thoroughly. 
Nevertheless, since the target-clutter distinguishing scheme is not always perfect, an overlarge $h$ would also wipe out some targets. 
A typical example is the different performance of $h = 18$ or $22$ among four sequences. 
For Sequence 2 and 3, $h = 18$ or $22$ achieves the best performance.
But, they perform the worst for Sequence 1 and 4.
It is because the target moves along the cloud edge in many frames of Sequence 1 and 4, and an overlarge $h$ would easily mistake the target as the edge and suppress it completely, resulting in a lower detection probability. 
On the contrary, a smaller $h$ might preserve the small target, but it also retains some non-target components, making the false-alarm ratio relatively high. 
For Sequence 2 and 3, when the detection probability is fixed, the false-alarm ratio of $h = 6$ is the largest. 
In order to seek a balance, we set the optimal $h$ as $10$ in the following experiment.

\subsubsection{Weighting parameter $\lambda$}
% Among so many factors, the weighting parameter is the most important. 
% In \cref{alg:alg1}, we have indicated that $\lambda$ controls the thresholding value of the target patch-tensor. 
% A wise weighting parameter could preserve the target and suppress the edges simultaneously. 
% However, a global constant weighting parameter shrinks all the components in the target patch-tensor equally. 
% With large weighting parameter, the background in the target image would be very clean, remaining few false alarms. 
% However, the over-large weighting parameter would also shrink the dim target, leaving out the true ones. On the contrary, a small weighting parameter not only preserves the true target, but also preserves many non-target components. 
% What's worse, the non-target components might be brighter than the targets, leading to a high false-alarm ratio. 
% This lies the motivation of our proposed model to replace the global constant weighting parameter with an element-wise and wiser one. 
Despite the usage of local structure weight, fine tuning of $\lambda$ is still of great importance. 
We show the effects of $\lambda$ in the fourth row of \cref{fig:Parameter_Analysis}. 
Since $\lambda$ is set as $L / \sqrt{\min(I,J,P)}$ in our model, instead of directly varying $\lambda$, we vary $L$ from $0.3$ to $1.5$. 
From the illustration, it can be observed that a large $\lambda$ does keep the false-alarm ratio being quite low like. 
For example, the ROC curves of $L = 1.0$ and $L = 1.5$ for Sequence 2 are straight line segments. 
But their detection probabilities are also low because many dim targets are suppressed by the overlarge threshold. 
On the contrary, when the detection probability is fixed, the false-alarm ratio of $L = 0.3$ is always higher than the others, suggesting that a too small $L$ is also not a good choice. 
% Empirically, we set $L = 0.7$ for the optimal weighting parameter in the following experiment.

\subsubsection{Penalty factor $\mu$}
It is precisely the shrinking threshold of \cref{eq:T_solver}, which influences the low-rank property of the background patch-tensor. 
With a smaller $\mu$, more details are preserved in the background patch-tensor. 
Thus fewer non-target components are left in the target patch-tensor. 
Nevertheless, the small target might be preserved in the background patch-tensor as well, resulting no target in the target image. 
On the contrary, a larger $\mu$ would lead to more non-target components lying in the target patch-tensor. 
Thus, it is necessary to choose an appropriate value for $\mu$ to keep the balance between detection probability and false-alarm ratio. 
Since we set $\mu = C_{\mu}\text{std}(\text{vec}(\bm{\mathcal{F}}))$, instead of varying $\mu$ directly, we investigate the influence of the penalty factor on Sequence 1 -- 4 by varying $C_{\mu}$ from $0.5$ to $3$. 
The results are shown in the last row of \cref{fig:Parameter_Analysis}, from which we can observe that an overlarge or too small $\mu$ is not an optimal choice and the best value for our four sequences is about $0.7$.

\subsection{Comparison with State-of-the-Arts}

\begin{figure*}[htbp]
	\centering
	\includegraphics[width=0.98\textwidth]{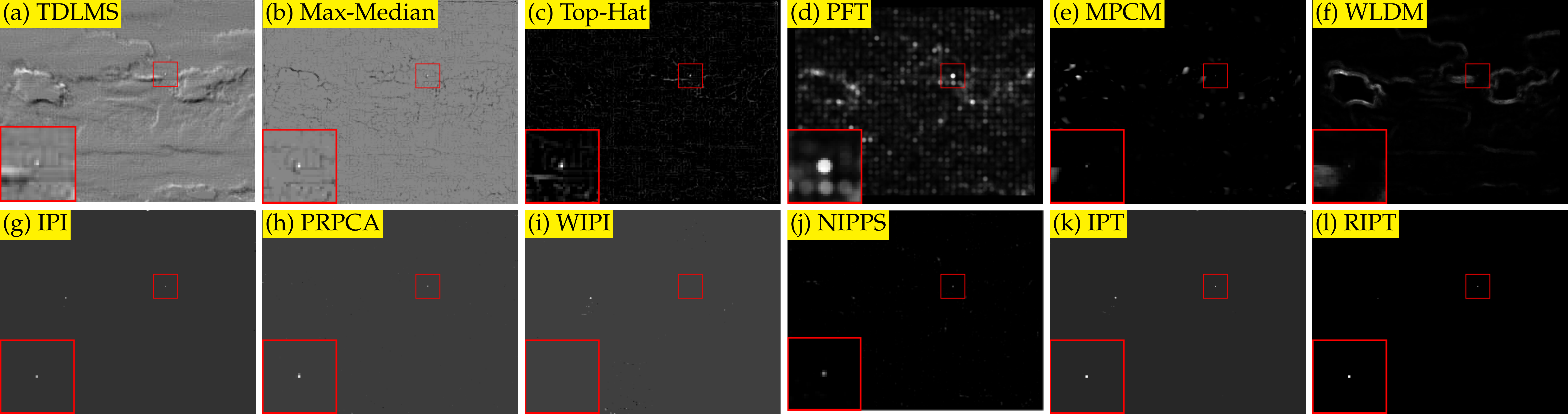}
	\caption{The separated target images of the $65$th frame in Sequence $1$ by twelve tested methods. For better visualization, the demarcated area is enlarged in the left bottom corner. It is better to be seen by zooming on a computer screen.}
	\label{fig:Set_2_Target}
\end{figure*}

\begin{figure*}[htb!]
	\centering
	\includegraphics[width=0.98\textwidth]{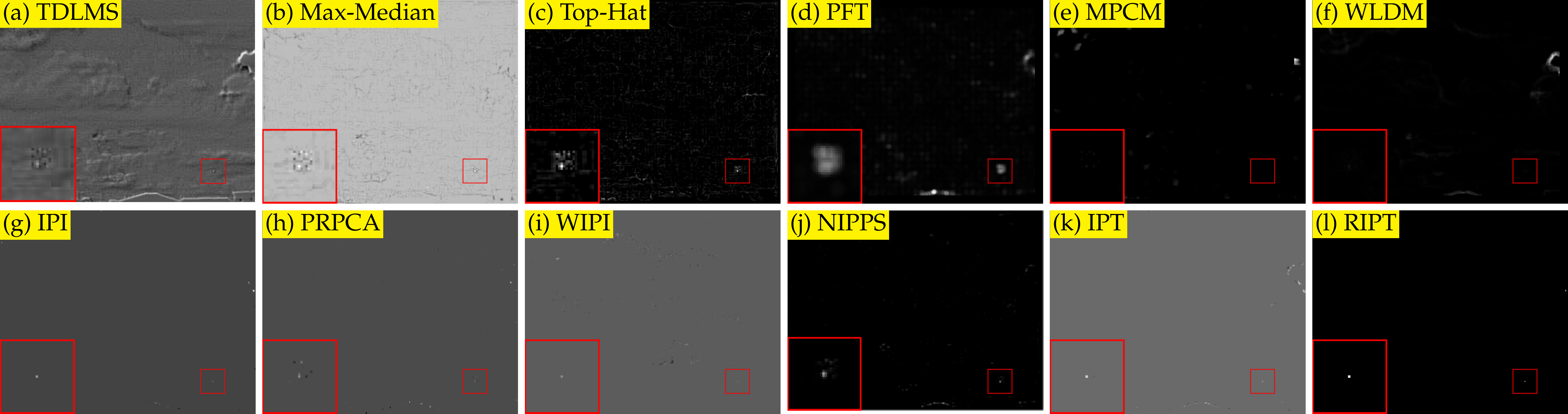}
	\caption{The separated target images of the $52$th frame in Sequence $2$ by twelve tested methods. For better visualization, the demarcated area is enlarged in the left bottom corner. It is better to be seen by zooming on a computer screen.}
	\label{fig:Set_3_Target}
\end{figure*}

\begin{figure*}[htbp]
	\centering
	\includegraphics[width=0.98\textwidth]{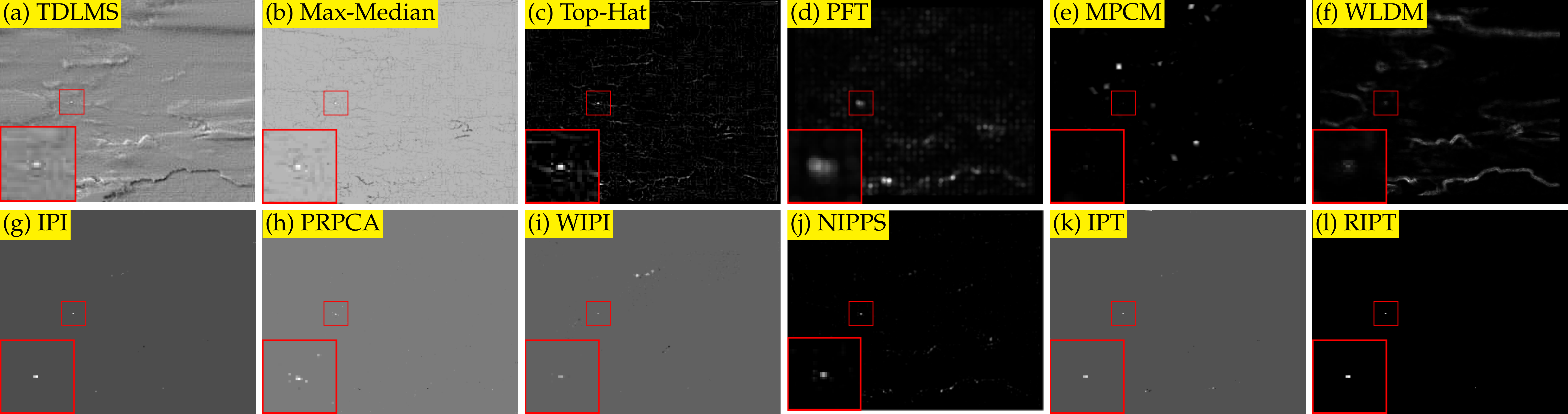}
	\caption{The separated target images of the $53$th frame in Sequence $3$ by twelve tested methods. For better visualization, the demarcated area is enlarged in the left bottom corner. It is better to be seen by zooming on a computer screen.}
	\label{fig:Set_4_Target}
\end{figure*}

\begin{figure*}[htbp]
	\centering
	\includegraphics[width=0.98\textwidth]{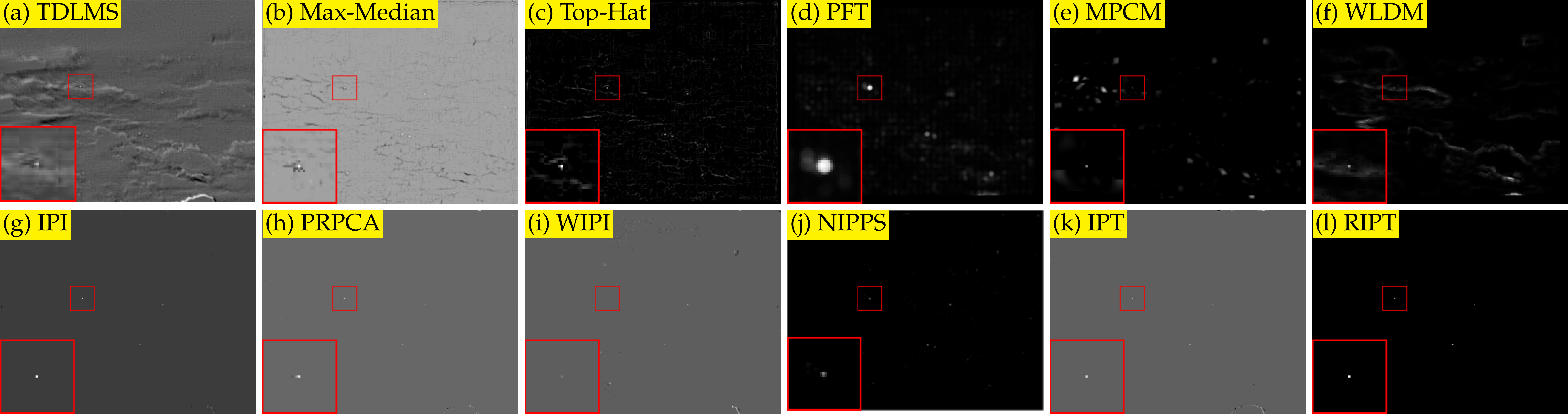}
	\caption{The separated target images of the $56$th frame in Sequence $4$ by twelve tested methods. For better visualization, the demarcated area is enlarged in the left bottom corner. It is better to be seen by zooming on a computer screen.}
	\label{fig:Set_5_Target}
\end{figure*}

In this subsection, we first compare the proposed model with the other state-of-the-art methods on the ability of clutter suppression. 
\cref{fig:Set_2_Target} -- \cref{fig:Set_5_Target} show the separated target images by twelve tested methods for four representative frames of Sequence 1 -- 4 in \cref{fig:Original}. 
It can be seen that the classical Max-Median filter does enhance the tiny targets in \cref{fig:Set_2_Target}(b) - \cref{fig:Set_5_Target}(b). 
Nevertheless, many non-target pixels are also enhanced simultaneously, especially in \cref{fig:Set_3_Target}(b) and \cref{fig:Set_4_Target}(b), which would raise many false alarms. 
In \cref{fig:Set_2_Target}(a) - \cref{fig:Set_5_Target}(a) produced by TDLMS, the phenomenon of enhancing non-target isolated points does not exist, but the cloud edges are highlighted, making them much brighter than the small target. 
Since the given target size matches the real target size exactly, the Top-Hat operator succeeds to enhance the target region. 
If not match, the Top-Hat operator is likely to lose the target. 
No matter whether the given and real target sizes match, Top-Hat cannot well suppress the background clutters. 
Many strong clutters still remain in resulting images, as illustrated in \cref{fig:Set_2_Target}(c) - \cref{fig:Set_5_Target}(c). 
Although PFT can retain the target to a certain extent, the target is not necessarily the brightest and there are also many non-target salient residuals, as shown in \cref{fig:Set_2_Target}(d) - \cref{fig:Set_5_Target}(d).
MPCM and WLDM failed to achieve good results because they suffered from a phenomenon we named rare structure effect which is caused by the inaccuracy of the local dissimilarity measure and often happens when the target is extremely dim.
In next subsection, we will further discuss this phenomenon.

In fact, the common and intrinsic reason behind the unsatisfactory performance of all these six methods lies in their pre-set assumption about the target shape, namely a hot spot brighter than its neighborhood. 
Nevertheless, when the target is too dim to maintain its significant contrast over non-target components, just like \cref{fig:Original}(a) -- (d), they might not perform as well as they usually do.

The last six tested methods are all low-rank based methods.
Comparing with above six methods, their results contain fewer background details.
Relatively speaking, the effects of PRPCA and WIPI are not as good as the other four methods. 
Different from the other low-rank based methods that all build their low-rank assumptions on the data structure composed of patches, PRPCA supposes the individual patch is low-rank. 
Thus in PRPCA, each patch is applied to an individual RPCA process.
Then all the separated target patches are synthesized into a target image. 
By comparing \cref{fig:Set_3_Target}(g) and \cref{fig:Set_5_Target}(g) with \cref{fig:Set_3_Target}(h) and \cref{fig:Set_5_Target}(h) , it can be seen that fewer edges were left by IPI than PRPCA. 
It is because the rare structure in a patch is not necessarily rare in the patch-image due to the redundancy of the whole image. 
Therefore, the results of IPI and IPT are much better than those of PRPCA. 
As to WIPI, considering the targets in Sequence 1 -- 4 is much dimmer than those in Ref.\ \cite{Dai2016IPT77}, it is fair to say that the steering kernel based patch-level weight is still not robust enough to handle all of the complex infrared backgrounds. 
From \cref{fig:Set_2_Target}(l) and \cref{fig:Set_5_Target}(l), we can see that with the help of the local structure weight, the non-target components were suppressed completely via our proposed model. 
For example, the cloud residuals in \cref{fig:Set_2_Target}(g) by IPI is brighter than its target, while in \cref{fig:Set_2_Target}(l), it is wiped out thoroughly. 
Based on above comparisons, it is fair to conclude that the proposed RIPT model achieves the most satisfying performance in infrared background suppression among twelve tested methods.

For infrared small target detection, the biggest difficulty is the interference of complex backgrounds. 
These undesirable background clutters raise the false alarm rates, and might even overwhelm the dim targets. 
Therefore, the ability of successfully suppressing the background clutters is a major concern in evaluating an infrared small target detection method.
Quantitative evaluating indices are also widely used to assess this ability. 
\cref{tab:Quantitative} shows the experimental data of all twelve tested methods for \cref{fig:Original}(a) -- (d). 
The gray-scale of every separated target image is rescaled to the range $0--255$. 
It could be observed that our proposed method gets the highest scores among all indices and all tested images. 
Different from the filtering based methods, for the low-rank based methods, \textbf{Inf}, namely infinity, is quite common, which just means that the target neighborhood completely shrinks to zero. 
In addition, it should be noted that the high scores in these three quantitative indices merely reflect the good suppression performance in a local region, and not necessarily mean a good global suppression ability.
\begin{table*}[htb]
	\centering
	\begin{threeparttable}    
		\setlength{\tabcolsep}{5pt}
		\scriptsize
		\caption{Quantitative comparison of the tested methods for the representative images of Sequence $1$-$4$}
		\begin{tabularx}{0.92\textwidth}{l r r r r r r r r r r r r r r r}
			\toprule
			& \multicolumn{3}{l}{$65$th frame of Sequence $1$} & & \multicolumn{3}{l}{$52$th frame of Sequence $2$} & & \multicolumn{3}{l}{$53$th frame of Sequence $3$} & & \multicolumn{3}{l}{$56$th frame of Sequence $4$} \\ 
			\cmidrule{2-4} \cmidrule{6-8} \cmidrule{10-12} \cmidrule{14-16}    
			Method     & \multicolumn{1}{l}{LSNRG} & \multicolumn{1}{l}{SCRG} & \multicolumn{1}{l}{BSF} &           & \multicolumn{1}{l}{LSNRG} & \multicolumn{1}{l}{SCRG} & \multicolumn{1}{l}{BSF} &           & \multicolumn{1}{l}{LSNRG} & \multicolumn{1}{l}{SCRG} & \multicolumn{1}{l}{BSF} &        & \multicolumn{1}{l}{LSNRG} & \multicolumn{1}{l}{SCRG} & \multicolumn{1}{l}{BSF} \\
			\midrule          
    Max-Median & 5.49  & 10.69  & 12.10  &       & 1.87  & 5.46  & 7.37  &       & 2.96  & 6.21  & 11.27  &       & 7.54  & 9.81  & 16.66  \\
    Top-Hat & 3.47  & 13.47  & 12.34  &       & 2.10  & 12.55  & 8.08  &       & 3.10  & 9.48  & 11.24  &       & 4.04  & 22.13  & 21.06  \\
    PFT   & 4.83  & 53.01  & 7.43  &       & 1.37  & 10.96  & 3.22  &       & 0.68  & 7.48  & 3.38  &       & 9.16  & 113.25  & 18.77  \\
    MPCM  & 1.48  & 7.69  & 3.46  &       & 1.62  & 11.84  & 15.98  &       & 0.38  & 1.91  & 2.15  &       & 1.88  & 14.17  & 4.68  \\
    WLDM  & 0.87  & 2.00  & 1.99  &       & 2.22  & 9.95  & 12.23  &       & 1.94  & 8.19  & 3.95  &       & 3.11  & 12.11  & 3.56  \\
    TDLMS & 1.36  & 3.44  & 3.53  &       & 1.76  & 4.27  & 3.38  &       & 2.61  & 4.39  & 4.49  &       & 1.99  & 4.77  & 4.50  \\
    IPI   & 220.38  & 5215.82  & 19256.20  &       & 10.72  & 104.34  & 172.90  &       & \textbf{Inf} & \textbf{Inf} & \textbf{Inf} &       & \textbf{Inf} & 2788.19  & 4939.03  \\
    PRPCA & 5.17  & 382.58  & 20179.12  &       & 1.30  & 26.88  & 2628.42  &       & 1.68  & 31.85  & 1982.80  &       & 2.83  & 267.31  & 20966.41  \\
    CWRPCA & 2.67  & 36.77  & 602.45  &       & 4.62  & 40.59  & 58.23  &       & 7.69  & 98.57  & 201.89  &       & 52.92  & 441.65  & 2065.42  \\
    NIPPS & 15.95  & 315.08  & 670.65  &       & 3.89  & 66.99  & 81.05  &       & 20.59  & 343.06  & 735.19  &       & 87.92  & 2280.13  & 3103.00  \\
    IPT   & 9.80  & 2096.70  & 87797.82  &       & 2.14  & 332.56  & 17488.27  &       & 3.22  & \textbf{Inf} & \textbf{Inf} &       & 2.86  & \textbf{Inf} & \textbf{Inf} \\
    RIPT  & \textbf{Inf} & \textbf{Inf} & \textbf{Inf} & \textbf{} & \textbf{Inf} & \textbf{Inf} & \textbf{Inf} & \textbf{} & \textbf{Inf} & \textbf{Inf} & \textbf{Inf} &       & \textbf{Inf} & \textbf{Inf} & \textbf{Inf} \\
			\bottomrule
		\end{tabularx}%
    \label{tab:Quantitative}%
		\begin{tablenotes}
			\item[*] Different from the filtering based methods, for the low-rank based methods, \textbf{Inf}, namely infinity, is quite common, which just means that the target neighborhood completely shrinks to zero.
		\end{tablenotes}
	\end{threeparttable}        	
\end{table*}%

\begin{figure*}
	\centering
	\subfloat[]{
		\includegraphics[width=0.232\textwidth]{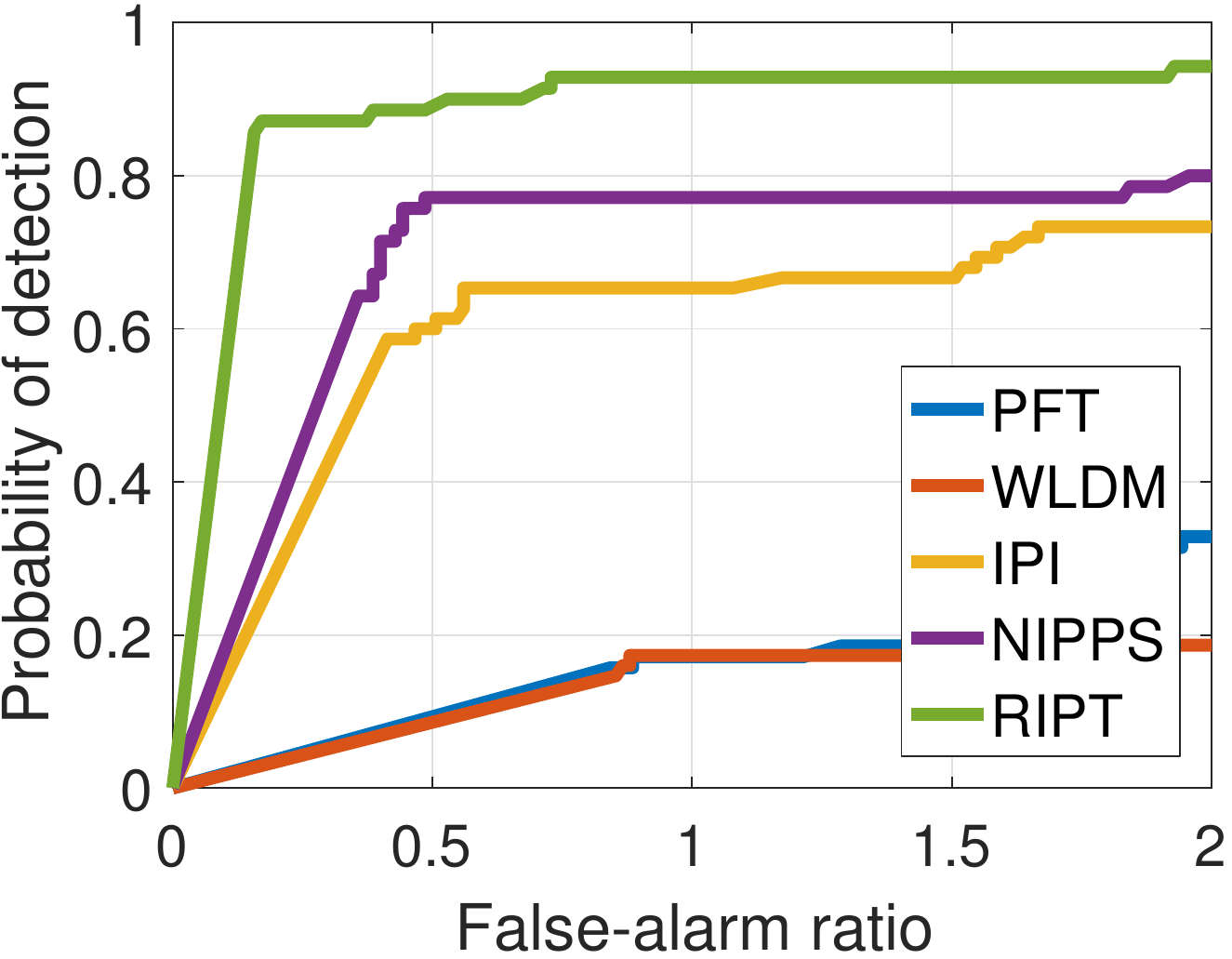}
	}
	\subfloat[]{
		\includegraphics[width=0.232\textwidth]{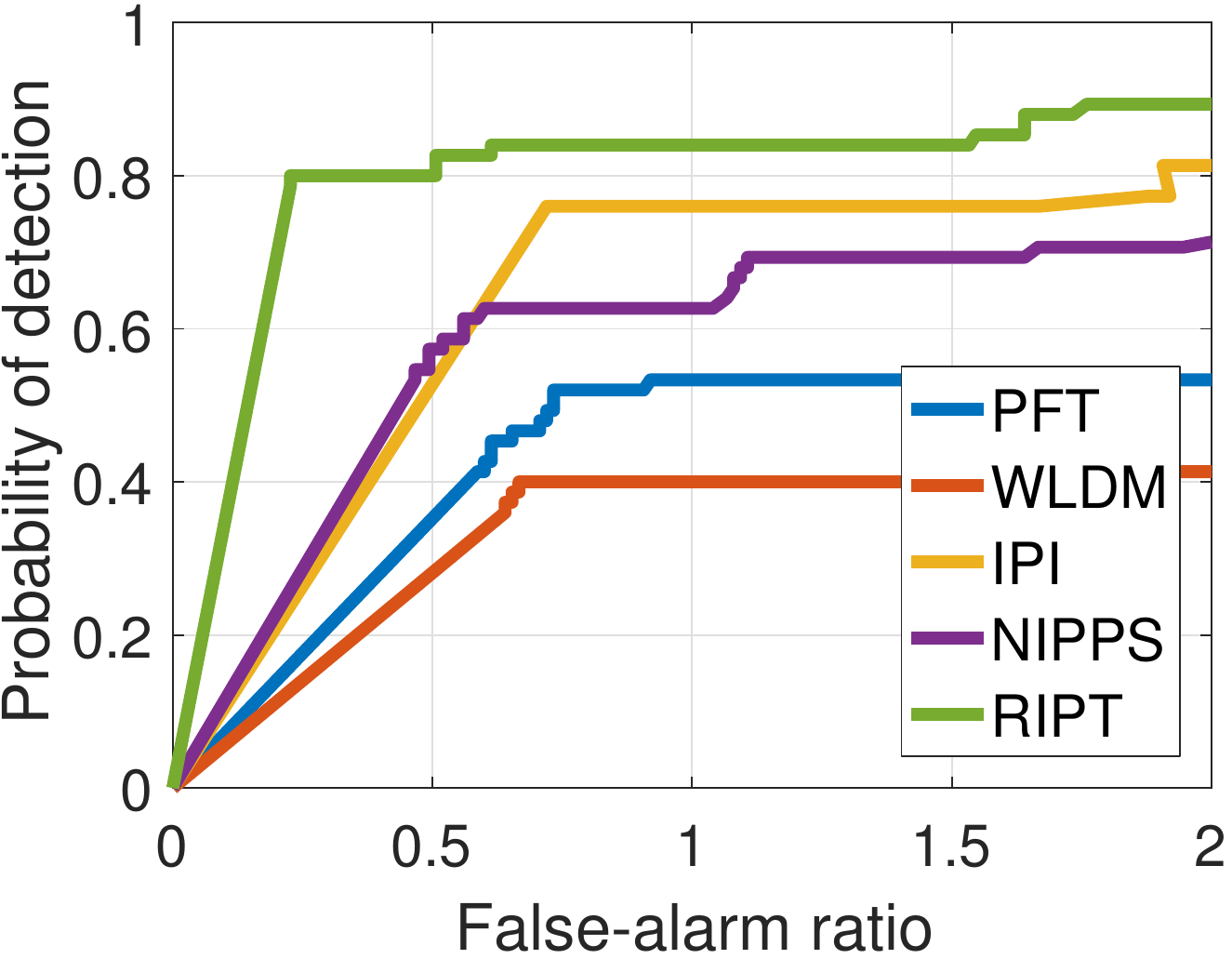}
	}
	\subfloat[]{
		\includegraphics[width=0.232\textwidth]{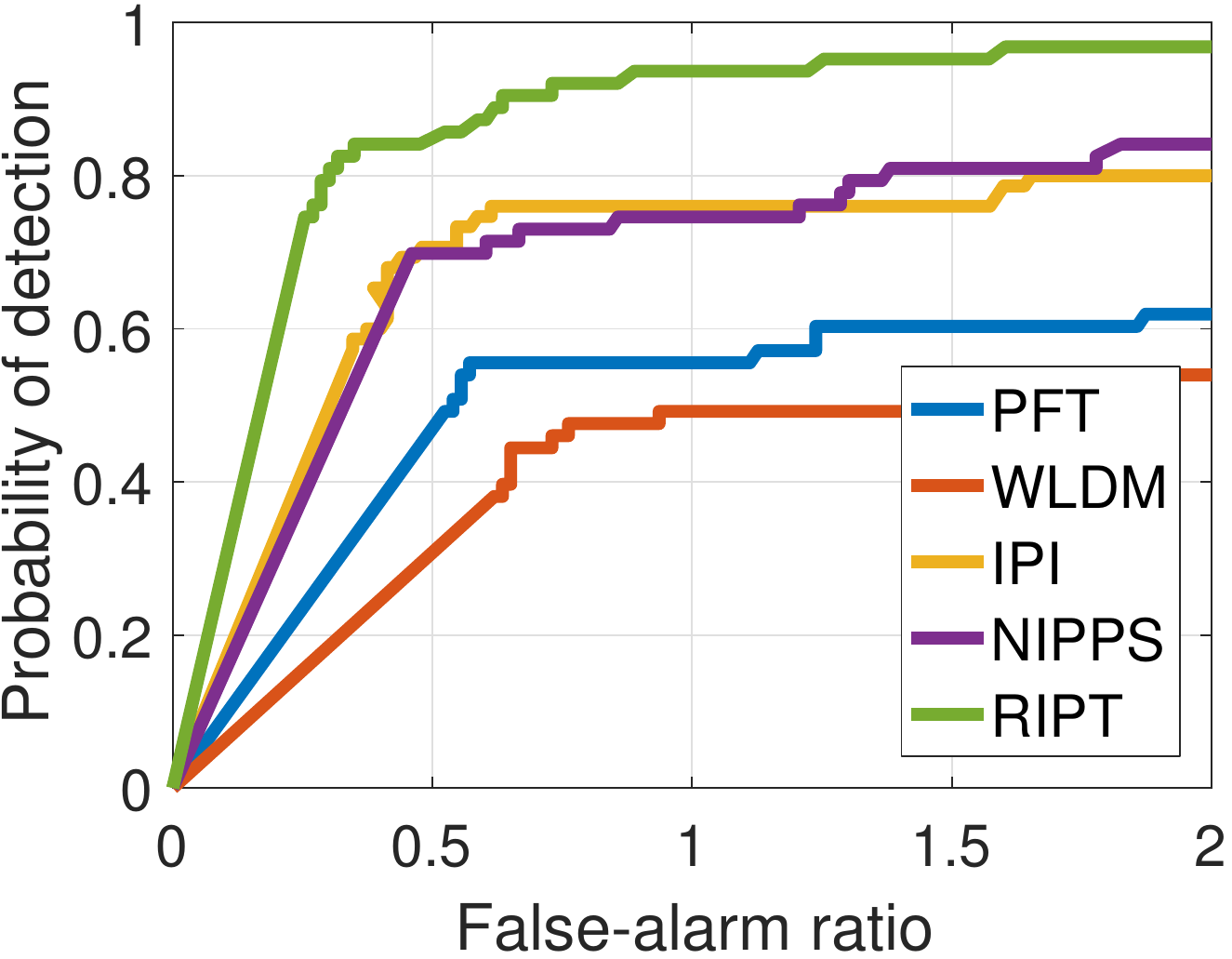}
	} 
	\subfloat[]{
		\includegraphics[width=0.232\textwidth]{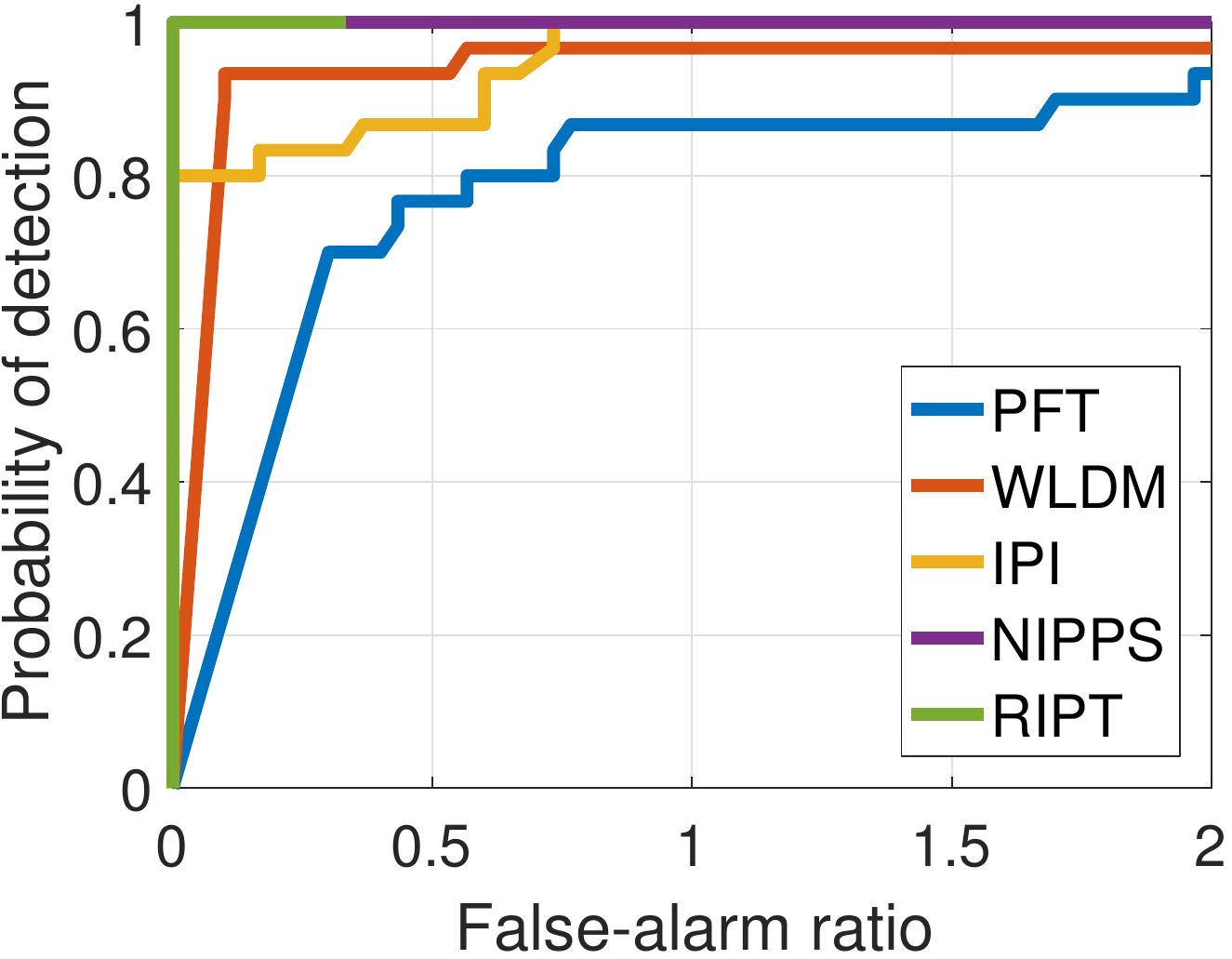}
	} 
	\caption{ROC curves of detection results of four real infrared sequences: (a) Sequence 1, (b) Sequence 3, (c) Sequence 4, and (d) Sequence 5.}
	\label{fig:ROC}
\end{figure*}

To further reveal the advantage of the proposed model, we display the ROC curves of Sequence 1 and Sequence 3 -- 5 for comparison in \cref{fig:ROC}.
The most interesting point is the performances of the state-of-the-art WLDM on Sequence 1, 3, 4 and Sequence 5 are very different. 
For Sequence 5, WLDM performs very well but fails in Sequence 1 -- 4.
We believe the reason lies in the rare structure effect which is a born problem for local contrast method.
NIPPS's performance is slightly better than the IPI model.
Finally, the proposed algorithm achieves the highest detection probability for the same false-alarm ratio, meaning that the proposed RIPT model has a better performance than the other models. 

\section{Conclusion}\label{sec:Conclusion}
To further suppress the strong edges while preserving the spatial correlation, a reweighted infrared patch-tensor model for small target detection is developed in this paper, simultaneously combining non-local redundant prior and local structure prior together. 
A local structure weight is designed based on the structure tensor and served as an edge indicator in the weighted model. 
In addition, a sparsity enhancement scheme is adopted to avoid the target image being contaminated. 
Then the target-background separation task is modeled as a reweighted robust tensor recovery problem, which can be efficiently solved via ADMM. 
Detailed experimental results show that our proposed model is robust to various scenarios and obtains 
the clearest separated target images compared with the state-of-the-art target-background separation methods. 

\section*{Acknowledgments}
The authors would like to thank the editor and anonymous reviewers for their helpful comments and suggestions. 
This work was supported in part by the National Natural Science Foundation of China under Grants no. 61573183, and Open Research Fund of Key Laboratory of Spectral Imaging Technology, Chinese Academy of Sciences under Grant no. LSIT201401.

\bibliography{ref}

\begin{IEEEbiography}[{\includegraphics[width=1in,height=1.25in,clip,keepaspectratio]{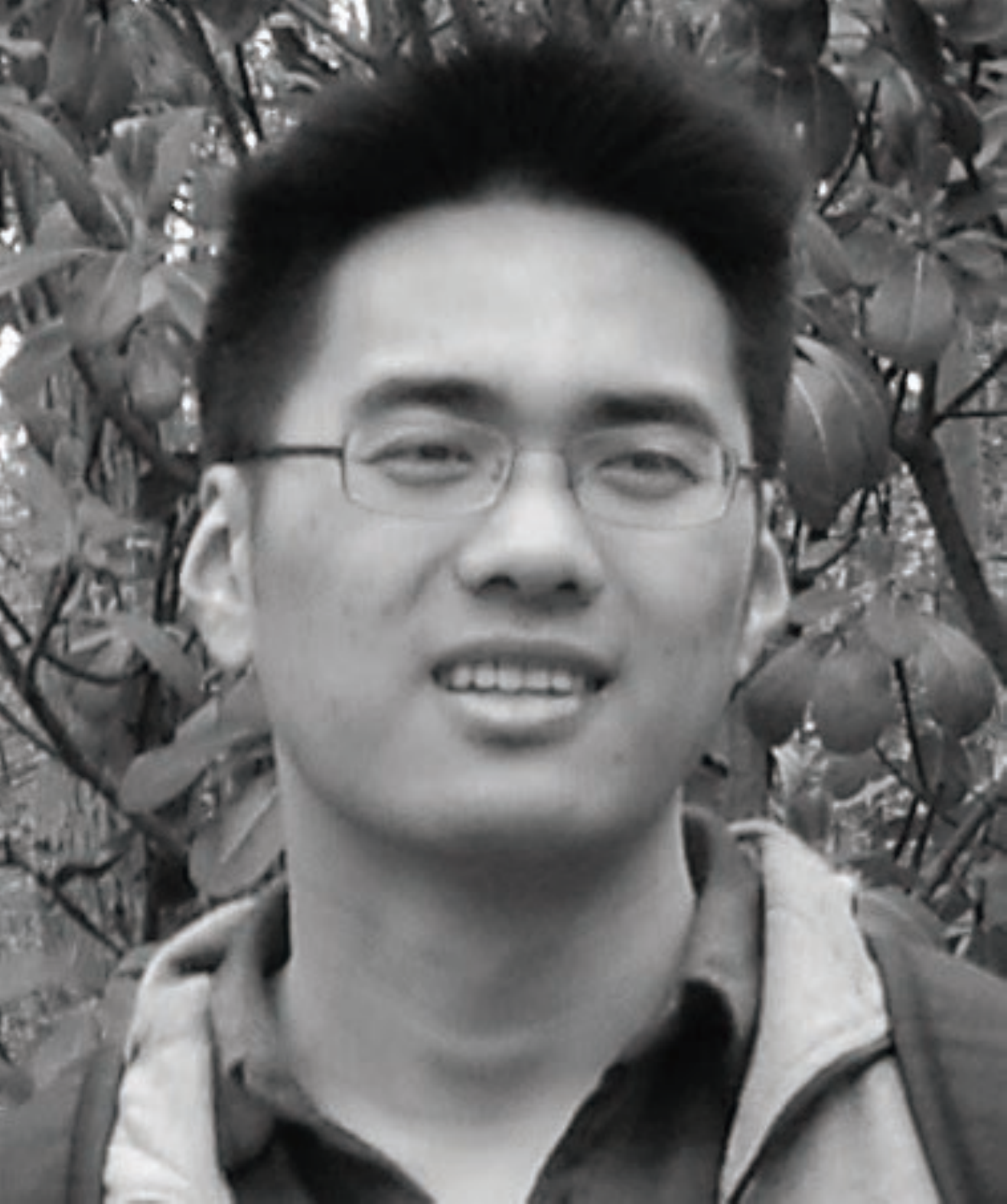}}]{Yimian Dai}
	received the B.S. degree from Nanjing University of Aeronautics and Astronautics (NUAA), Nanjing, China, in 2013, where he is currently pursuing the Ph.D. degree. His current interests include target detection, image restoration, and machine learning.
\end{IEEEbiography}

\begin{IEEEbiography}[{\includegraphics[width=1in,height=1.25in,clip,keepaspectratio]{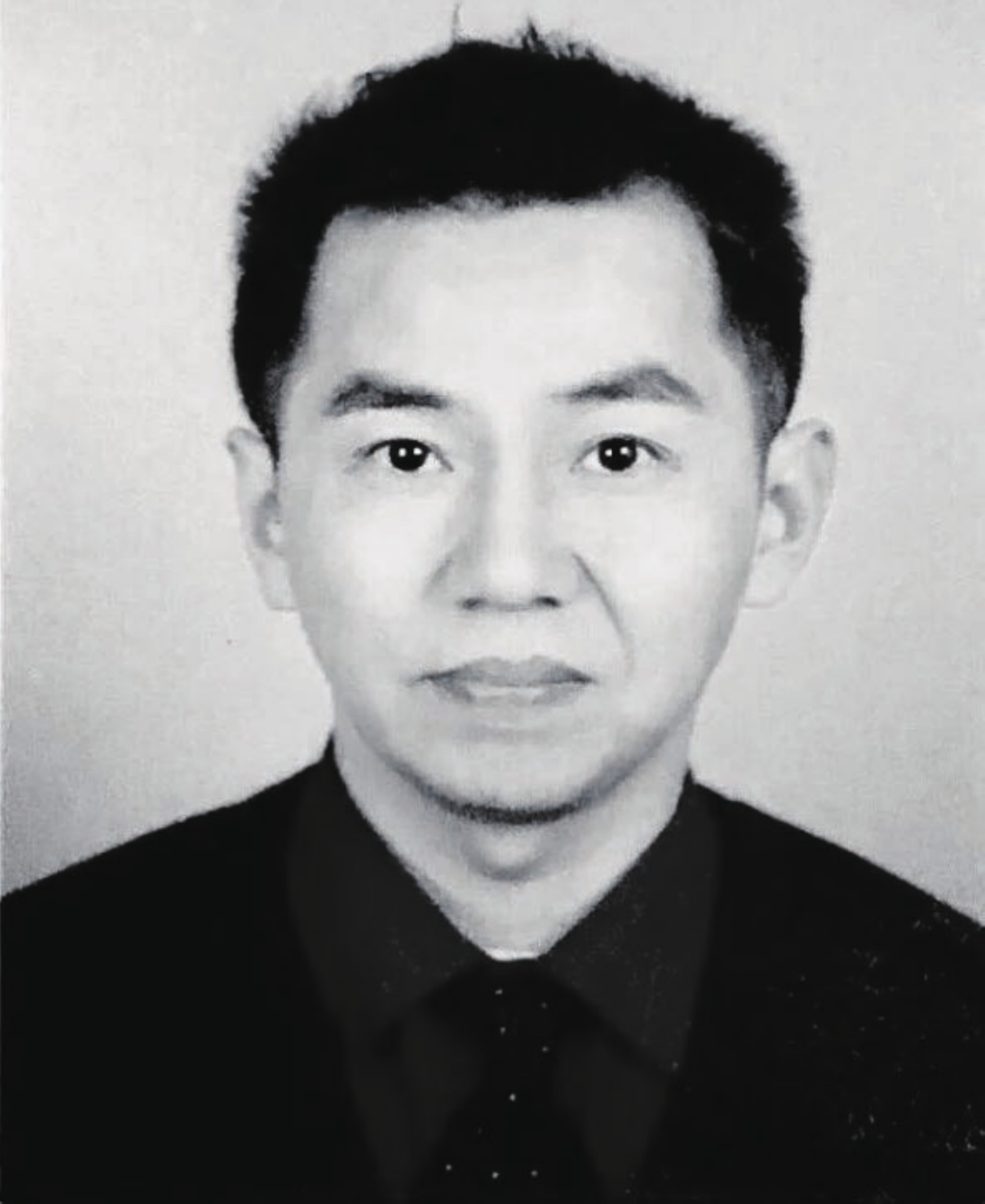}}]{Yiquan Wu}
	received his M.S. and Ph.D. degrees from Nanjing University of Aeronautics and Astronautics in 1987 and 1998, respectively. He is at present a professor and Ph.D. supervisor in the Department of Information and Communication Engineering at the Nanjing University of Aeronautics and Astronautics, where he is involved in teaching and research in the areas of image processing and recognition, target detection and tracking, and intelligent information processing. 
\end{IEEEbiography}

\end{document}